\documentclass[a4paper]{cas-dc}
\usepackage[numbers]{natbib}

\usepackage{tabularx}
\usepackage{hyperref}
\usepackage{comment}
\usepackage{amsmath,amssymb,amsfonts,gensymb}
\usepackage{cleveref}
\usepackage{algorithmic}
\usepackage{graphicx}
\usepackage{subcaption}
\usepackage{caption}
\usepackage{textcomp}
\usepackage{amssymb}
\usepackage{pifont}
\usepackage{supertabular,booktabs}
\usepackage[normalem]{ulem}
\usepackage{multirow}
\usepackage{fancybox}
\usepackage{anyfontsize}	
\usepackage{lipsum}
\usepackage[utf8]{inputenc}	
\usepackage[T1]{fontenc}
\usepackage{textcomp}		
\usepackage{pdflscape}
\usepackage{array}
\usepackage{longtable}
\usepackage{xcolor}
\usepackage{pgfplotstable}
\usepackage{tikz}
\usetikzlibrary{shapes.geometric, arrows}

\tikzstyle{infowh1} = [rectangle, rounded corners, 
minimum width=3cm, 
minimum height=1cm,
text centered, 
font=\bfseries,
draw=white, 
fill=white!0]

\tikzstyle{infowh2} = [rectangle, rounded corners, 
minimum width=3cm, 
minimum height=1cm,
text centered,
draw=white, 
fill=gray!10]

\tikzstyle{infobl} = [rectangle, rounded corners, 
minimum width=3cm, 
minimum height=1cm,
text centered, 
draw=white, 
fill=blue!10]

\tikzstyle{infoyel} = [rectangle, rounded corners, 
minimum width=3cm, 
minimum height=1cm,
text centered, 
draw=white, 
fill=yellow!10]

\tikzstyle{arrow} = [thick,->,>=stealth]

\newcommand{\cs}[1]{\textcolor{red}{[#1]}}

\begin{document}
\let\WriteBookmarks\relax
\def\floatpagepagefraction{1}
\def\textpagefraction{.001}

\shorttitle{A Survey on Deep Learning-Based Monocular Spacecraft Pose Estimation: Current State, Limitations and Prospects}    

\shortauthors{Pauly et al.}  

\title [mode = title]{A Survey on Deep Learning-Based Monocular Spacecraft Pose Estimation: Current State, Limitations and Prospects}

\author[1]{Leo Pauly}
\ead{leo.pauly@uni.lu}

\author[1]{Wassim Rharbaoui}
\ead{wassim.rharbaoui@uni.lu}

\author[1]{Carl Shneider}
\ead{carl.shneider@uni.lu}

\author[1]{Arunkumar Rathinam}
\ead{arunkumar.rathinam@uni.lu}

\author[1]{Vincent Gaudilli\`ere}
\ead{vincent.gaudilliere@uni.lu}

\author[1]{Djamila Aouada}
\ead{djamila.aouada@uni.lu}

\affiliation[1]{organization={Interdisciplinary Centre for Security, Reliability and Trust (SnT)},addressline={University of Luxembourg}, city={Luxembourg}, postcode={L-1855},state={Luxembourg},country={Luxembourg}}

\begin{abstract}
Estimating the pose of an uncooperative spacecraft is an important computer vision problem for enabling the deployment of automatic vision-based systems in orbit, with applications ranging from on-orbit servicing to space debris removal. Following the general trend in computer vision, more and more works have been focusing on leveraging Deep Learning (DL) methods to address this problem. However and despite promising research-stage results, major challenges preventing the use of such methods in real-life missions still stand in the way. In particular, the deployment of such computation-intensive algorithms is still under-investigated, while the performance drop when training on synthetic and testing on real images remains to mitigate. The primary goal of this survey is to describe the current DL-based methods for spacecraft pose estimation in a comprehensive manner. The secondary goal is to help define the limitations towards the effective deployment of DL-based spacecraft pose estimation solutions for reliable autonomous vision-based applications. To this end, the survey first summarises the existing algorithms according to two approaches: hybrid modular pipelines and direct end-to-end regression methods. A comparison of algorithms is presented not only in terms of pose accuracy but also with a focus on network architectures and models' sizes keeping potential deployment in mind. Then, current monocular spacecraft pose estimation datasets used to train and test these methods are discussed. The data generation methods: simulators and testbeds, the domain gap and the performance drop between synthetically generated and lab/space collected images and the potential solutions are also discussed. Finally, the paper presents open research questions and future directions in the field, drawing parallels with other computer vision applications.
\end{abstract}

\maketitle

\section{Introduction}
\label{sec:introduction}

\begin{figure}[!t]
	\centering
	\includegraphics[scale=.45]{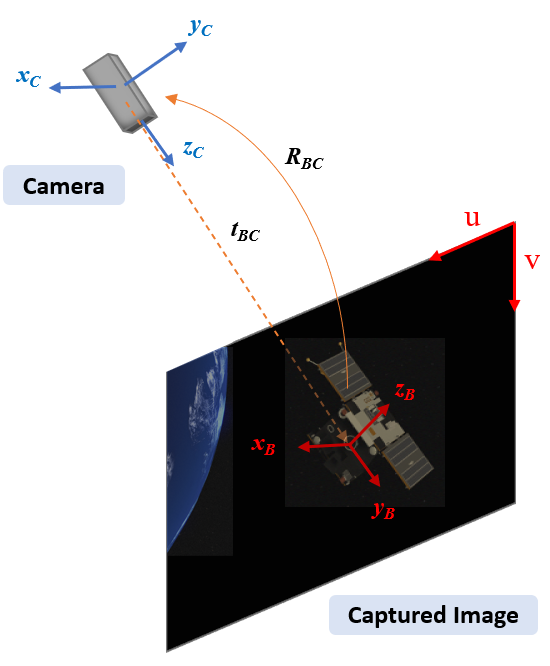}
	\caption{Spacecraft pose estimation is the problem of finding the relative position ($t_{BC}$) and orientation ($R_{BC}$) of the target spacecraft reference frame (B) shown in red, with respect to the camera reference frame (C) shown in blue, mounted on a chaser spacecraft.}
	\label{fig:figure_4}
\end{figure}

In recent years, the number of satellites launched into orbit has increased rapidly, aided by lower launch costs and minimal entry barriers, making space more accessible than ever before \cite{jones2018recent,witze_2023}. Each space mission has a unique set of goals that influences the satellite's size, functions, and intended lifetime. In most mission scenarios, the satellites launched into orbit will last for the entire mission life-cycle, and at the end of life, they are either moved to the graveyard orbit or left to re-enter the Earth's atmosphere. However, a few space missions may encounter anomalies or malfunctions before their full life span. These malfunctioned satellites may become non-cooperative and threaten existing space infrastructure. To tackle such scenarios, the demand for orbital missions targeting On-Orbit Servicing (OOS) and Active Debris Removal (ADR) has steadily increased, as OOS and ADR are considered key spaceflight  capabilities for the next decade.  OOS is defined as the process of inspection, maintenance, and repair of a system as an in-space operation. Commercial OOS missions aim to perform various functions, including providing life extension, maintaining the spacecraft, rescuing and recovering satellites from deployment failures and assisting astronauts with extravehicular activities \cite{kreisel2002orbit,li2019orbit}. ADR is the process of removing obsolete space objects (such as satellites, rocket bodies, or fragments of spacecraft) through an external disposal method, thus minimizing the build-up of unnecessary objects and lowering the probability of on-orbit collisions that can fuel a “collision cascade” \cite{wijayatunga_design_2023, may2021triggers}.  Several technology demonstration missions, including PROBA-3 by the European Space Agency (ESA) \cite{llorente2013proba}, PRISMA by OHB Sweden~\cite{PRISMA_OHB}, and commercial missions such as MEV-1 by Northrop Grumman \cite{redd2020}, had been carried out successfully in recent years. Future missions such as Clearspace-1 by ESA and Clearspace \cite{biesbroek2021clearspace} are already in preparation to demonstrate ADR in 2026.

An important aspect of OOS and ADR missions is that it requires rendezvous and proximity operations near the target before performing mission-specific operations. 
To perform any rendezvous operations, it is essential to know the target spacecraft's position and orientation (i.e. pose), allowing the relative navigation algorithms to generate real-time trajectories onboard the spacecraft. Several sensor options are available to perform inference and observation of the target spacecraft state, including Monocular RGB/Greyscale Cameras, Stereo Cameras, Thermal cameras, Range Detection and Ranging (RADAR), Light Detection and Ranging (LIDAR), etc. Monocular cameras are widely preferred over other active sensors (like LIDARs and RADARs) due to their relative simplicity, small size, weight, power requirements, and ability to be easily integrated into a wide range of spacecraft configurations. 

Recovering the relative pose between a camera and an observed object from a single image is a fundamental computer vision problem \cite{marullo20226d,park2019pix2pose,szeliski2022computer}. Given an image and the corresponding intrinsic camera parameters, the relative pose estimation problem involves estimating the relative transformation, \textit{i.e.} translation and rotation, between the camera and the target object. The location of the object in the camera reference frame is specified by $\mathit{t}\in\mathbb{R}^3$, and its orientation is most often represented by a quaternion $\mathit{q}=(q_0,q_1,q_2,q_3) \in\mathbb{R}^4$. The relative orientation (rotation) can also be represented using standard 3D rotation representations such as rotation matrix or Euler's Angles \cite{huynh2009metrics}. In \Cref{fig:figure_4}, a simple illustration of the spacecraft pose estimation problem is presented, where axes $x_C, y_C, z_C$ represent the camera reference frame mounted on the chaser (C) spacecraft and $x_B, y_B, z_B$ represent the target spacecraft's body (B) reference frame.  Spacecraft pose estimation is the problem of finding the relative position ($t_{BC}$) and orientation ($R_{BC}$) of the reference frame of a target spacecraft with respect to the reference frame of a camera mounted on a chaser spacecraft, using a single image from a monocular camera.  

In the last decade, vision-based spacecraft pose estimation has utilized hand-engineered features described using feature descriptors and detected using feature detectors to detect these features in the 2D images and to finally use their 3D correspondences to find the relative pose \cite{kelsey2006, dpose2014}. Although the use of feature correspondences between the detected features in the 2D image and 3D feature locations, together with perspective transformation, aids in pose solution convergence, the features are not robust to harsh lighting conditions encountered in space. The feature-based approaches perform poorly in variable illumination conditions, low signal-to-noise ratio, and high contrast characteristics encountered in space imagery. This results in a poor estimation of the target state in many scenarios. Spacecraft pose estimation before the evolution of deep learning algorithms has been summarised in \cite{cassinis2019review, opromolla2017review}. With their gain in popularity and exponential growth, Deep Learning (DL)-based approaches have prompted many new developments in recent years. According to the findings of the recent ESA's Spacecraft Pose Estimation Challenges \cite{kisantal2020,park2023satellite}, DL-based methods have been the preferred option for tackling the problem of uncooperative spacecraft pose estimation. However, investigated DL-based approaches still heavily rely on annotated data that are cumbersome to obtain. While synthetic data generation and laboratory data acquisition have been identified as the most tractable way to train and test such algorithms, the performance drops significantly on the test image domain compared to the train image domain, such problem being known as the \textit{domain gap} \cite{wang2022generalizing}. Dedicated strategies have therefore to be investigated to mitigate it. In addition, the laboratory conditions under which test images are acquired still differ from space-borne conditions, adding another level of domain discrepancy that is yet to be addressed.

A recent survey on the DL-based approaches for spacecraft relative navigation \cite{song2022deep} provides a general narrative across different use cases, including spacecraft pose estimation. In this survey, we focus on monocular pose estimation of non-cooperative targets using DL approaches and review the latest developments in the field. In addition, we conduct a comparison between the two main types of approaches and assess the still unmet needs that would enable the deployment of DL-based algorithms in real space missions. Furthermore, we explore the fundamental counterpart of any DL-based algorithm that is the data. We review the existing datasets, generation engines and testbed facilities. We also analyse the current validation procedure that consists in testing on laboratory-acquired images algorithms trained on synthetic data, after discussing the methods proposed to address this domain gap. Finally, we provide the reader with prospects on research directions that could help making the leap to the deployment of reliable DL-based spacecraft pose estimation algorithms for autonomous in-orbit operations. Note that we mainly considered the works published until Dec 2022 for this survey.

The following sections are organized as follows. Section \ref{sec:algorithms} provides a comprehensive survey of the two main types of DL-based algorithms for spacecraft pose estimation, before highlighting their limitations. Section \ref{sec:data} presents the datasets, generation engines and testbed facilities. It also presents the main existing methods to address the domain gap between synthetic and laboratory images, and discuss the underlying validation procedure. Section \ref{sec:future} discusses open research problems and future directions and finally, Section \ref{sec:conclusion} concludes the survey.

\begin{figure*}[!t]
	\centering
\begin{tikzpicture}[node distance=2cm]

\node(SPE)[infowh1]{
\begin{tabular}{c}
    Spacecraft \\
    Pose \\ 
    Estimation
\end{tabular}
};

\node(HMA)[infobl, above of=SPE, xshift=2cm, yshift=0.5cm]{\begin{tabular}{c}
    Hybrid \\ 
    Modular \\
    Approach
    \end{tabular}};

\node(SSD)[infowh2, right of=SPE, xshift=9.8cm, yshift=4.5cm]{
\begin{tabular}{l}
- Multi-stage detectors \cite{chen2019,huan2020pose,rathinam2020} \\
- Single-stage detectors \cite{park2019towards,huo2020,piazza2021deep,black2021real,cosmas2020,spetpu22,kecenli2022,CA-SpaceNet2022}
\end{tabular}
};

\node(RKL)[infowh2, right of=SPE, xshift=9cm, yshift=2.5cm]{\begin{tabular}{l}
- Regression of keypoint locations \cite{huan2020pose,park2019,spetpu22} \\
- Segmentation-driven approach \cite{gerard2019,hu2021,legrandend} \\
- Heatmap prediction \cite{chen2019,rathinam2020,huo2020,piazza2021deep,cosmas2020} \\
- Bounding box prediction \cite{kecenli2022,CA-SpaceNet2022}
\end{tabular}};

\node(PNP)[infowh2, right of=SPE, xshift=10.1cm, yshift=0.5cm]
{
\begin{tabular}{l}
    - PnP solver \cite{chen2019,gerard2019,park2019,huo2020,piazza2021deep,huan2020pose,rathinam2020,black2021real,hu2021,spetpu22,kecenli2022,CA-SpaceNet2022}\\
    - Learning-based method \cite{legrandend} \\ 
\end{tabular} 
};

\node(SL)[infoyel, right of=SPE, xshift=3.5cm, yshift=4.5cm]{\begin{tabular}{c}
    Spacecraft \\ 
    localisation 
    \end{tabular}};
    
\node(KP)[infoyel, right of=SPE, xshift=3.5cm, yshift=2.5cm]{\begin{tabular}{c}
    Keypoint \\ 
    Prediction 
    \end{tabular}};
    
\node(PC)[infoyel, right of=SPE, xshift=3.5cm, yshift=0.5cm]{\begin{tabular}{c}
    Pose \\ 
    Computation 
    \end{tabular}}; 
    
\node(DETEA)[infobl, below of=SPE, xshift=4.5cm, yshift=.1cm]{\begin{tabular}{c}
    Direct End- \\ 
    to-end \\Approach\cite{sharma2018pose,sharma2019pose,park2022robust,proenca2020,posso2022mobileurso,garcia2021lspnet,huang2021non,phisannupawong2020vision}
    \end{tabular}};

\draw[arrow](SPE) |- (HMA);
\draw[arrow](SPE) |- (DETEA);
\draw[arrow](HMA) |- (SL);
\draw[arrow](HMA) -- (KP);
\draw[arrow](HMA) |- (PC);
\draw[arrow](SL) -- (SSD);
\draw[arrow](KP) -- (RKL);
\draw[arrow](PC) -- (PNP);
\tikzstyle{arrow2} = [thick,->,dotted]
\draw[arrow2](KP) -- (PC);
\draw[arrow2](SL) -- (KP);

\end{tikzpicture}
\caption{Tree diagram of spacecraft pose estimation algorithms reviewed in this paper. Blue boxes show the two different categories of approaches: hybrid modular and direct end-to-end. The yellow boxes and the sub-branches (grey boxes) show the separate stages and the different methods used at each stage, respectively, of the hybrid modular approach.}
	\label{fig:algo_overview}
\end{figure*}
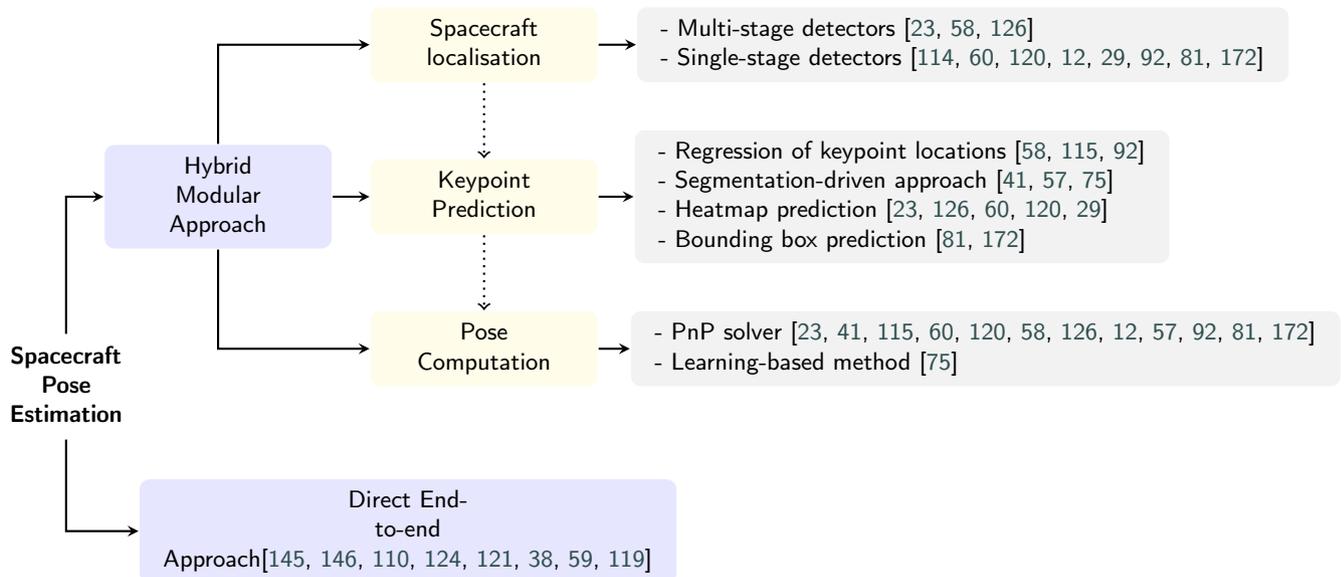

\section{Algorithms}
\label{sec:algorithms}

\begin{figure}[t]
	\centering
	\includegraphics[trim={0 0 0 0},clip,scale=.26]{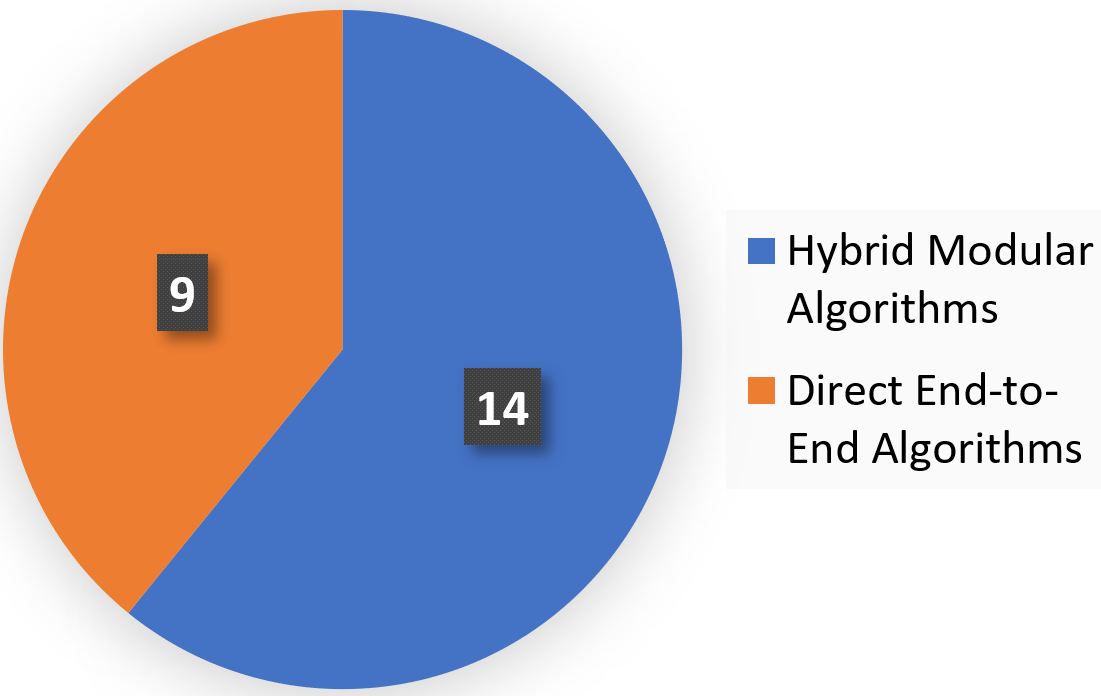}
	\caption{Distribution of algorithms surveyed in this paper}
	\label{fig:algo_distribution}
\end{figure}

\begin{figure*}[t]
	\centering
	\includegraphics[scale=.365]{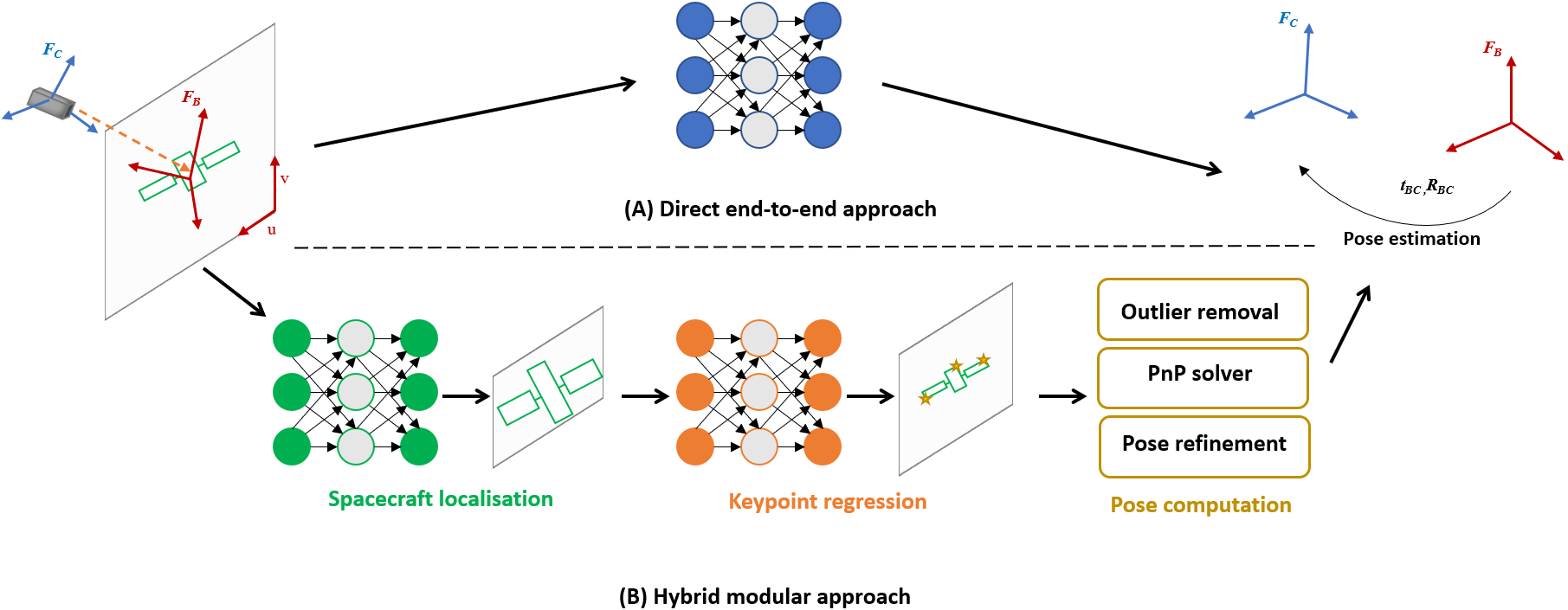}
	\caption{ Illustration of different approaches for spacecraft pose estimation. A) Direct end-to-end approaches which use deep learning. B) Hybrid modular approaches which consist of three steps: object detection/localisation, keypoint regression, and pose computation. The first two steps use deep learning and the third step uses a classical algorithm which performs outlier removal necessary for the PnP solver and finally pose refinement.} 
	\label{fig:figure_1new}
\end{figure*}

The use of DL has had significant implications in developing computer vision algorithms over the last decade ~\cite{voulodimos2018deep}, ~\cite{chai2021deep}, improving their performance and robustness for applications such as image classification \cite{wang2019development}, segmentation \cite{minaee2021image}, and object tracking \cite{ciaparrone2020deep}. Following this trend, the proposals of DL-based spacecraft pose estimation algorithms have outnumbered~\cite{song2022deep}, ~\cite{cassinis2019review} the classical feature-engineering-based methods~\cite{d2014pose,shi2016spacecraft,liu2014relative,sharma2018robust,rondao2018multi,capuano2019robust} in recent years. \Cref{fig:algo_overview} presents an overview tree diagram of the algorithms reviewed in this survey and \Cref{fig:algo_distribution} shows their branching into different approaches. DL-based spacecraft pose estimation algorithms broadly fall under two categories: 1) Hybrid modular approaches, and 2) Direct end-to-end approaches.
%
\\
Hybrid modular approaches (see \Cref{fig:figure_1new}-A) combine multiple DL models and classical computer vision methods for spacecraft pose estimation. On the other hand, direct end-to-end approaches (see \Cref{fig:figure_1new}-B) only use a single DL model for pose estimation, trained end-to-end. Each of these approaches are discussed in detail (\Cref{sec:hybrid_algorithms} and \Cref{sec:direct_algorithms}), with a comparative analysis (\Cref{sec:algo_comparison}) and a discussion on limitations (\Cref{sec:algo_limitations}) below. 

\begin{figure*}[t]
	\centering
	\includegraphics[scale=.61]{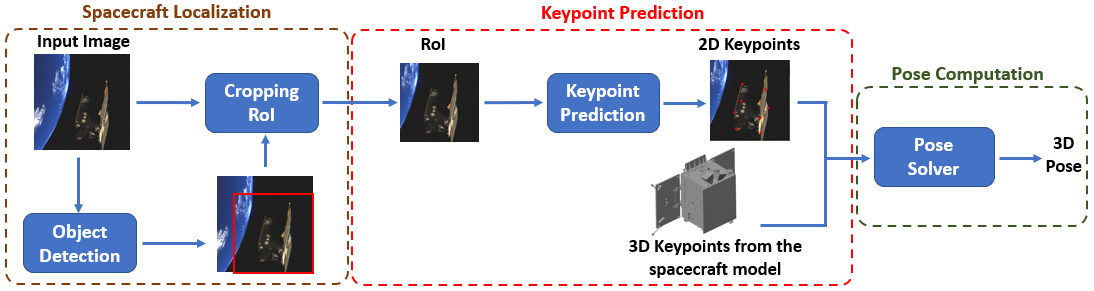}
	\caption{Hybrid modular approach for spacecraft pose estimation. The spacecraft localisation stage is outlined in blue, the keypoint prediction stage is in red and the pose computation stage is shown in green. Spacecraft image from the SPARK2 dataset is used for illustration \cite{rathinam_arunkumar_2022_6599762}.} 
	\label{fig:hybrid_approach_overview}
\end{figure*}

\subsection{Hybrid Modular Approaches}
\label{sec:hybrid_algorithms}

This survey defines hybrid approaches as those using a combination of DL models and classical computer vision methods for spacecraft pose estimation. The hybrid algorithms have three common stages (see \Cref{fig:hybrid_approach_overview}):  (1) \textit{spacecraft localisation} for detecting and cropping the spacecraft region in the image, (2) \textit{keypoint prediction} for predicting 2D keypoints locations of pre-defined 3D keypoints inside cropped regions and (3) \textit{pose computation} for computing the pose from these 2D-3D correspondences. The following subsections describe each of these stages in detail. 

\subsubsection{Spacecraft Localisation} The spacecraft object size in the image varies considerably with changes in the relative distance between the chaser and target spacecraft as illustrated in \Cref{fig:object_size_vary}. This scale variance affects the performance of the pose estimation algorithm~\cite{kisantal2020}. The spacecraft localisation stage uses a DL object detection framework to detect the spacecraft by predicting bounding boxes around the object (spacecraft). These bounding boxes are then used to crop out the region of interest (RoI) in the image containing the spacecraft. The extracted RoI is then processed for pose estimation in the subsequent stages. Based on literature~\cite{jiao2019survey}, DL-based object detectors for spacecraft localisation can be classified into two categories: 

\begin{figure}[!t]
	\centering
	\includegraphics[width=\linewidth]{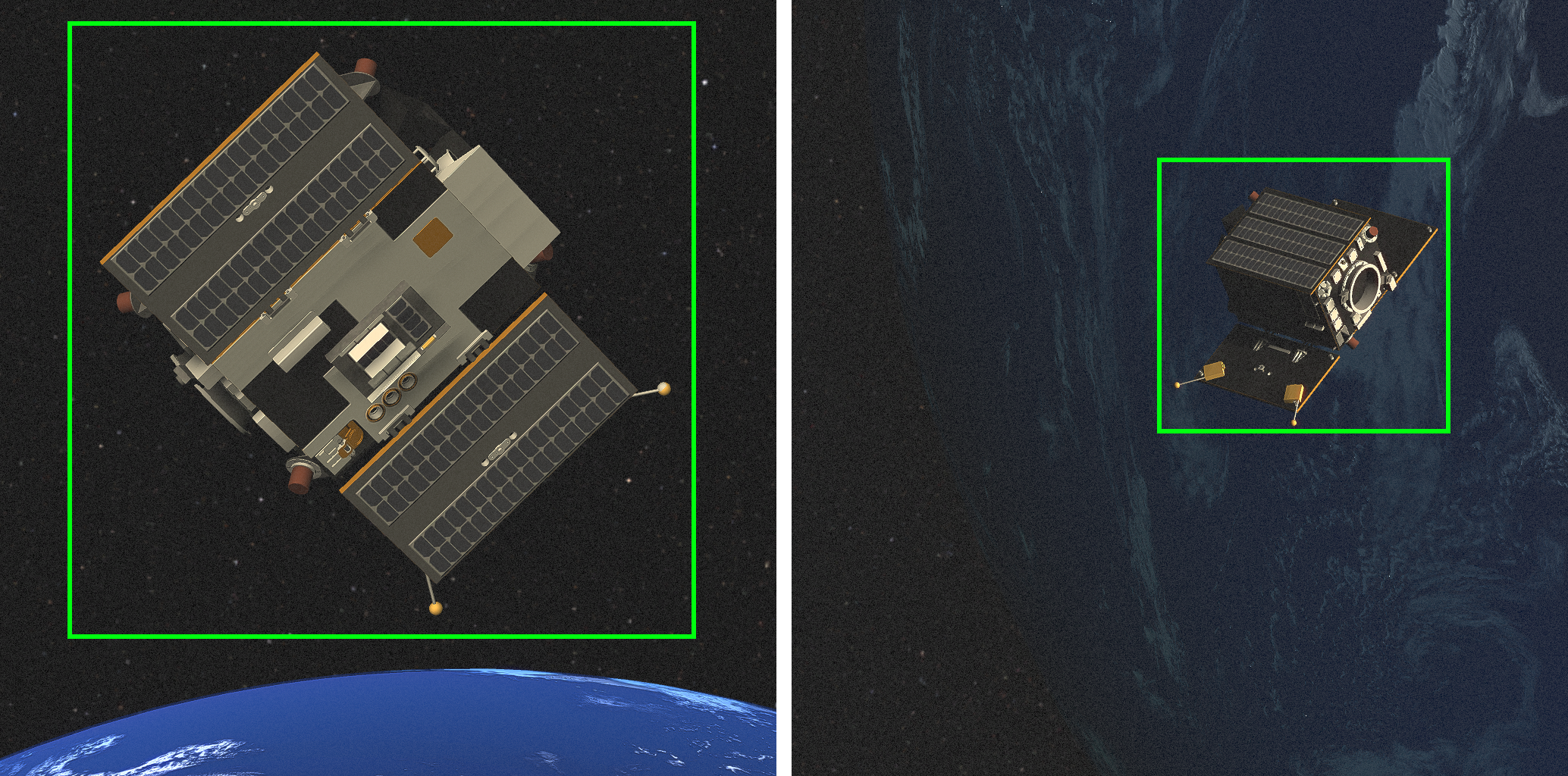}
	\caption{Illustrating variations in spacecraft size in captured images. The bounding boxes predicted by an object detector are shown in green. These images are taken from the SPARK2 \cite{rathinam_arunkumar_2022_6599762} dataset, and show the Proba-2 spacecraft class.}
     \label{fig:object_size_vary}
\end{figure}

\begin{itemize}
    \item Multi-stage object detectors 
    \item Single-stage object detectors
\end{itemize}

\textbf{Multi-stage object detectors:} In these detectors object detection proceeds in multiple stages. The first stage generates region proposals, i.e. image areas with a higher probability of containing objects to be detected. These region proposals are then refined and classified in the second stage. Detectors of this kind generally provide highly accurate detections. However, due to their multi-stage nature, they suffer from longer image processing times (high latency) and higher number of parameters making them resource-intensive. This can be particularly detrimental in resource-constrained scenarios such as those encountered in space. Faster R-CNN~\cite{ren2015faster} and  Mask R-CNN~\cite{he2017mask} are the commonly used multi-stage object detectors for spacecraft localisation. 

\textbf{Single-stage object detectors:} These detectors, on the other hand, are lightweight detectors with a reduced number of parameters and have lower latency for real-time detection. YOLO~\cite{redmon2016you} (and its derivatives), SSD~\cite{liu2016ssd}, and MobileDet~\cite{xiong2021mobiledets} are the single-stage detectors applied in the different spacecraft pose estimation algorithms reviewed this survey. 

Several other object detectors have also been proposed in the wider computer vision literature, which can be applied for spacecraft localisation. Zaidi \textit{et al.}~\cite{zaidi2022survey} and Zou \textit{et al.}~\cite{zou2019object} presented detailed surveys on different classes of object detectors and their characteristics. The modular nature of the hybrid approaches makes it easier to replace object detectors in the pose estimation algorithms based on criteria such as the number of parameters, resource utilisation, latency and real-time inference.

\subsubsection{Keypoint Prediction}

\begin{figure*}[!t]
	\centering
	\includegraphics[scale=.39]{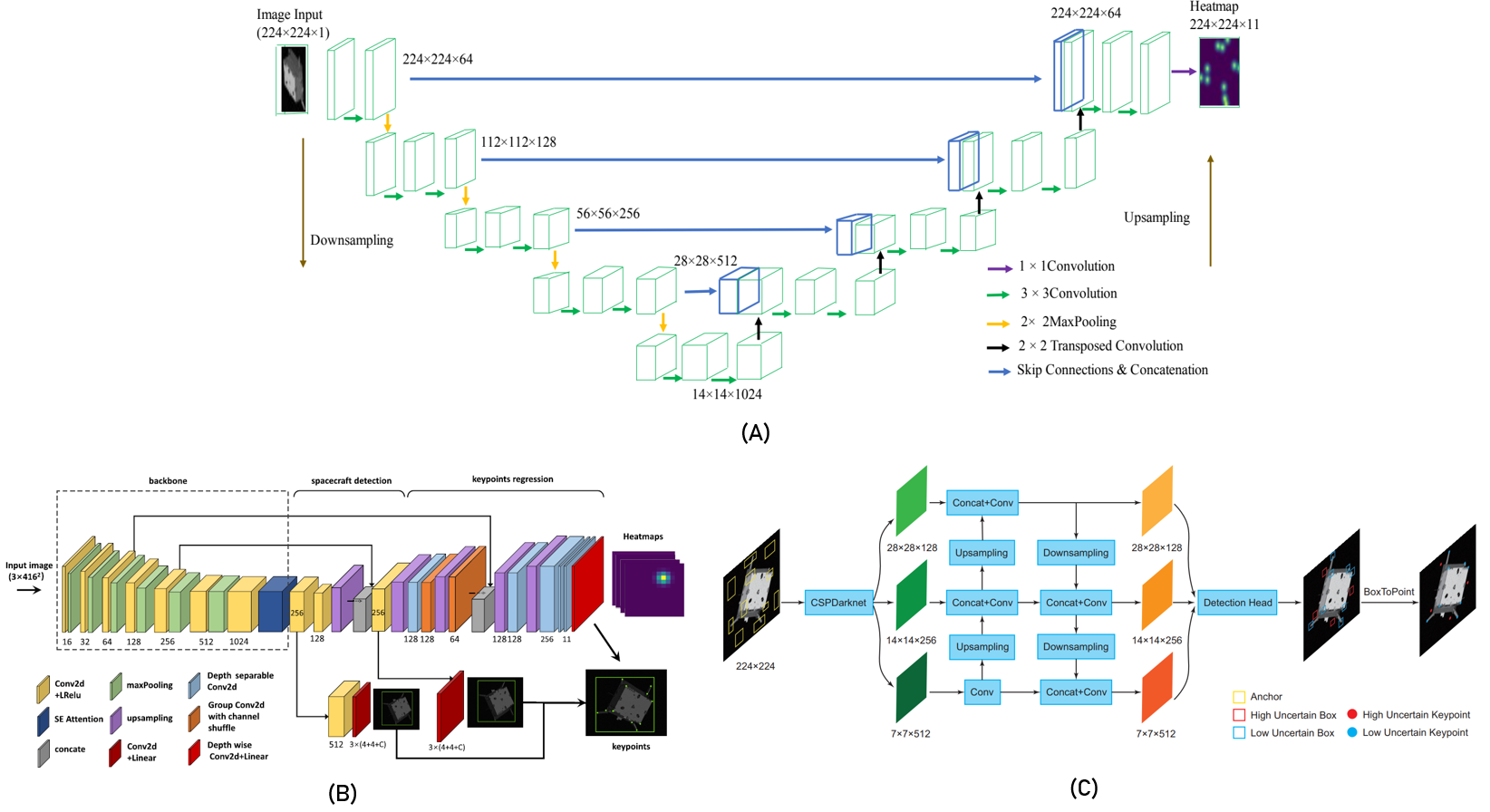}
	\caption{(A) Keypoint heatmap prediction with a ResNet-UNet architecture \cite{cosmas2020} (B) YOLO-like CNN detector with a heatmap regression subnetwork~\cite{huo2020} (C) Keypoint prediction is formulated as a keypoint bounding box detection problem~\cite{kecenli2022}.}
	\label{fig:kp_architectures}
\end{figure*}

In this stage, the 2D projections of a set of predefined 3D keypoints are predicted from the cropped regions containing the spacecraft using a DL model (see \Cref{fig:hybrid_approach_overview}). The 3D keypoints are generally defined by the CAD model of the spacecraft. If the CAD model is not available, multiview triangulation (as in~\cite{old_chen2019}~\cite{price2021monocular}~\cite{huo2020}) or Structure from Motion (SfM) techniques~\cite{hartley2003multiple} can be used for reconstructing a wireframe 3D model of the spacecraft containing the 3D keypoints.

\textbf{Regression of keypoint locations:} 
A common method for predicting keypoints is to directly regress the keypoint locations. Huan et al.~\cite{huan2020pose} uses a CNN regression model with an HRNet\cite{wang2020deep} backbone for directly regressing the 2D keypoint locations as a $1\times1\times2M$ vector, where $M$ is the number of keypoints. Park et al.~\cite{park2019} uses a YOLOv2\cite{redmon2017yolo9000} based architecture with a MobileNetv2\cite{sandler2018mobilenetv2} backbone with only 5.64M parameters for regressing keypoints. The lightweight nature of the model makes it suitable for deployment in space hardware or edge devices. Similarly, Lotti et al.~\cite{spetpu22} also propose a deployable CNN regression model for keypoint regression with EfficientNet-Lite backbone~\cite{efficientnetlite}, which is obtained by removing operations not well supported for mobile applications (deployment) from the original EfficientNets\cite{tan2019efficientnet}.


\textbf{Segmentation-driven approach:}
Algorithms in~\cite{gerard2019},\cite{hu2021} and~\cite{legrandend} follow the segmentation-driven approach from Hu et al.~\cite{hu2019segmentation} for regressing the keypoint locations, with a dual-headed (segmentation and regression) network architecture and a shared backbone. The input image is divided into a grid and the segmentation head separates the foreground grid cells (containing the spacecraft) from the background. The regression head predicts the location of each keypoint as an offset from the centre of each of the grid cells. Only the predictions from foreground (spacecraft) grid cells contribute to the prediction of the keypoint location, making predictions more accurate. Additionally,~\cite{legrandend} also presents different variants of the keypoint prediction model with a lower number of parameters making it suitable for deployment in space hardware. The model with the lowest number of parameters achieving sufficient keypoint prediction accuracy uses a MobileNetv3\cite{howard2019searching} backbone that has only 7.8M parameters. 

\textbf{Heatmap prediction:}
Another method for keypoint prediction is to regress the heatmaps encoding probability of the keypoint locations. The pixel coordinates are then obtained by extracting locations with the highest probability from these heatmaps~\cite{chen2019}~\cite{rathinam2020}~\cite{huo2020}~\cite{piazza2021deep}~\cite{cosmas2020}. The ground truth heatmaps are generated as 2D normal distributions with means equal to the ground truth keypoint locations and unit standard deviations. HRNet~\cite{wang2020deep} network architecture and its derivative, the HigherHRNet~\cite{cheng2020}, is used extensively for heatmap predictions in different algorithms. HRNet architectures maintain high-resolution feature maps throughout the network making it suitable for heatmap prediction tasks. UNet~\cite{ronneberger2015u} architecture is also used for predicting keypoint heatmaps~\cite{cosmas2020} (see \Cref{fig:kp_architectures}-A). Originally developed for image segmentation, UNet architecture consists of a sequence of downsampling layers (contracting path) that captures relevant semantic information. This is followed by symmetrical upsampling layers (expanding path) for precise location predictions. The use of skip connections in the architecture preserves spatial information during downsampling and subsequent upsampling. Huo et al.~\cite{huo2020} presented a lightweight hybrid architecture for keypoint prediction combining a YOLO-like CNN spacecraft detector with a heatmap regression subnetwork (see \Cref{fig:kp_architectures}-B). Sharing the backbone network architecture between the object detection and the keypoint prediction brings down the total number of parameters to $\sim${}.89M, making it suitable to deploy in resource-constrained space systems.

\textbf{Bounding box prediction:} Recently, Li et al.~\cite{kecenli2022} formulated keypoint prediction as a keypoint bounding box detection problem. Instead of predicting the keypoint locations or heatmaps, the enclosing bounding boxes over the keypoints are predicted along with the confidence scores. Authors used CSPDarknet~\cite{bochkovskiy2020yolov4} CNN backbone with a Feature Pyramid Network (FPN)~\cite{lin2017feature} for multi-scale feature extraction, followed by a detection head for the keypoint bounding box detection (see \Cref{fig:kp_architectures}-C). A similar method is also used in~\cite{CA-SpaceNet2022}. Here, a counterfactual analysis~\cite{pearl2018book} framework is used to generate the FPN, which is then fed to the keypoint detector.

\subsubsection{Pose Computation}
The final stage is to compute the spacecraft pose using the 2D keypoints (from the keypoint prediction stage) and the corresponding pre-defined 3D points~\cite{marchand2015pose}. One important step in the pose computation process is to remove the wrongly predicted keypoints, referred to as \textit{outliers}, since the Perspective-$n$-Point (P$n$P) \cite{fischler1981random} solvers are sensitive to the presence of outliers. The RANdom SAmple Consensus (RANSAC)~\cite{strutz2011data} algorithm is commonly used for removing outliers. IterativePnP~\cite{opencv} and EPnP\cite{lepetit2009} are the two solvers extensively used in the different hybrid algorithms. Recently, Legrand et.al~\cite{legrandend} replaced the P$n$P solver with a Multi-Layer Perceptron (MLP) network architecture, the Pose Inference Network (PIN)~\cite{hu2020single}, for regressing the pose from the predicted keypoints. This makes pose computation differentiable and it can be trained with a pose loss function. In the final step, the estimated pose is further refined by optimising a geometrical loss function~\cite{kendall2017geometric} such as the keypoint reprojection error~\cite{chen2019}.

\subsection{Direct End-to-end Approaches}
\label{sec:direct_algorithms}

\begin{figure}[!t]
\centering
\includegraphics[scale=.25]{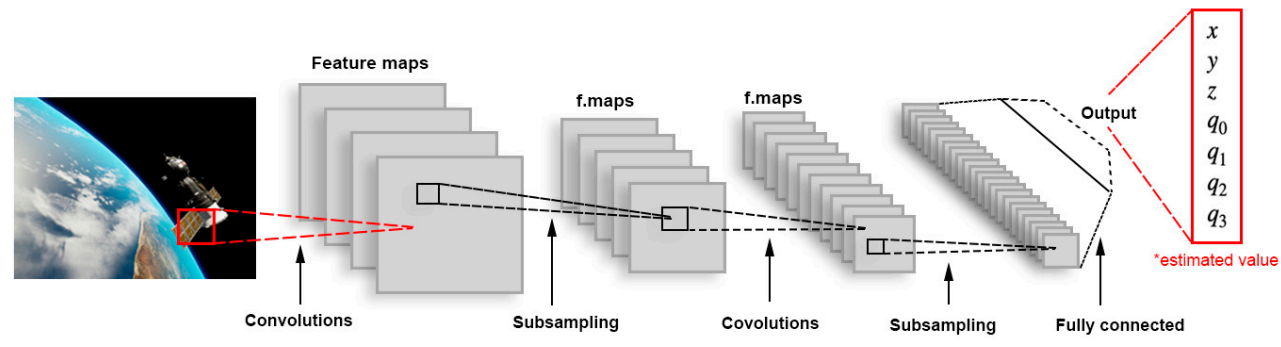}
\caption{Network architecture used in~\cite{phisannupawong2020vision}. A GoogLeNet~\cite{szegedy2015going} based CNN architecture is used to regress the 7D pose vector $[x,y,z,q_0,q_1,q_2,q_3]$.}
\label{fig:direct_position_orientation_regressed}
\end{figure}

In this survey, direct approaches refer to the use of only one DL model in an end-to-end manner for regressing the spacecraft pose directly from the images without relying on intermediate stages. The models are trained using loss functions calculated from the pose error. Unlike hybrid algorithms, the approach does not require any additional information like camera parameters or a 3D model of the spacecraft apart from the ground truth pose labels.  The camera parameters are intrinsically learned by the models during the training process. 

Phisannupawong \textit{et al.}~\cite{phisannupawong2020vision} proposed a GoogLeNet-based~\cite{szegedy2015going} CNN architecture for regressing the 7D pose vector representing position and orientation quaternion (see \Cref{fig:direct_position_orientation_regressed}). The network was trained using different loss functions, an exponential loss function and a weighted Euclidean-based loss function. The experimental results show that the network offers better performance when trained with the latter.  However, directly regressing the orientation using a norm-based loss of unit quaternions fails to achieve higher accuracies and results in a larger error margin~\cite{proenca2020}. This is mainly due to the loss function's inability to represent the actual angular distance of any orientation representation. 

\begin{figure}[!t]
	\centering
	\includegraphics[scale=.9]{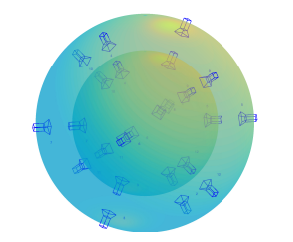}
	\caption{Illustration of pose space discretisation along four degrees of freedom used in~\cite{sharma2018pose}. Two degrees of freedom controlling the position of the camera on the enclosing sphere, one degree of freedom from the rotation of the camera along the bore-sight direction and one degree of freedom from the distance of the camera to the spacecraft.}
	\label{fig:pose_sphere}
\end{figure}

Sharma \textit{et al.}~\cite{sharma2018pose} proposed discretising the pose space itself into pose classification labels by quantising along four degrees of freedom as illustrated in \Cref{fig:pose_sphere}. Two degrees of freedom controlling the position of the camera (w.r.t. to the spacecraft) along the surface of the enclosing sphere, one degree of freedom denoting the rotation of the camera along the bore-sight angle and one degree of freedom determined by the distance of the camera from the spacecraft. An AlexNet-based~\cite{Krizhevsky2012ImageNetCW} CNN network is used for classifying the spacecraft images into these discretised pose label classes, trained with a Softmax loss function~\cite{wang2022comprehensive}. However, this is constrained by the total number of pose class labels to be learned. A larger number of pose labels will need an equivalent number of neurons in the final softmax layer, increasing model size considerably. Also, the method provides an initial guess and requires further refinement to produce more accurate pose estimations. 

\begin{figure}[!t]
	\centering
	\includegraphics[scale=.355]{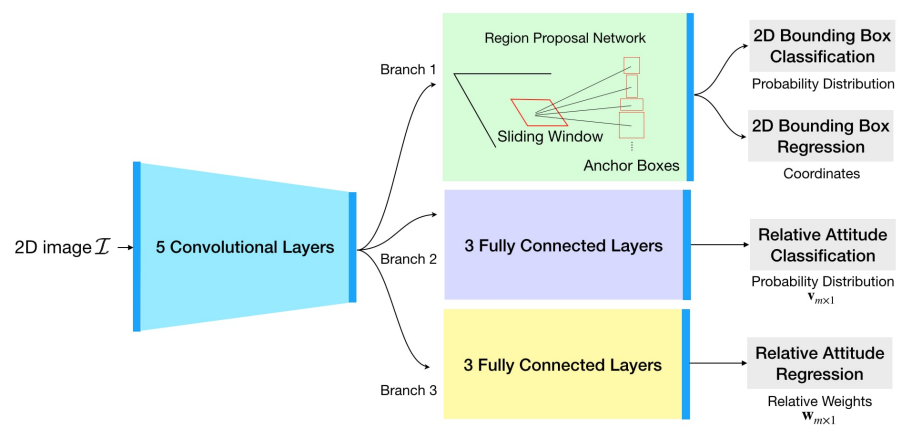}
	\caption{Network architecture used for spacecraft pose estimation in~\cite{sharma2019pose}. Branch 1 localises the spacecraft outputting the bounding box, branch 2 predicts the probability distribution for orientation classification and branch 3 regresses the weights for each orientation class.}
	\label{fig:SPN}
\end{figure}

To overcome these limitations, Sharma et al.~\cite{sharma2019pose} later presented Spacecraft Pose Network (SPN), a model with a five-layer CNN backbone followed by three different sub-branches (see \Cref{fig:SPN}). The first branch localises the spacecraft in the input image and returns the bounding box. The second branch classifies the target orientation in terms of a probability distribution of discrete classes. It minimises a standard cross entropy loss for a set of closest orientation labels. Finally, the third branch takes the candidate orientation class labels obtained from the previous branch and minimises another cross-entropy loss to yield the relative weighting of each orientation class. The final refined attitude is obtained via quaternion averaging with respect to the computed weights, which represents a soft classification approach. The position is then estimated from the constraints imposed by the detected bounding box and the estimated orientation, using the Gauss–Newton optimisation algorithm~\cite{mittelhammer2000econometric}.

Similar network architecture is also used in~\cite{huang2021non}. A ResNet50 model~\cite{he2016deep} with a Squeeze-and-Excitation (SE) module~\cite{hu2018squeeze}  is used as the base CNN network for feature extraction. The first sub-network, the attitude-prediction-subnetwork, estimates the orientation by soft classification and error quaternion regression. The second pose regression sub-network, predicts the position of the spacecraft by direct regression. Finally, the object detection sub-network detects the spacecraft by predicting the enclosing bounding box. The bounding box is used to validate the position and orientation prediction.

\begin{figure}[!t]
	\centering
	\includegraphics[scale=.32]{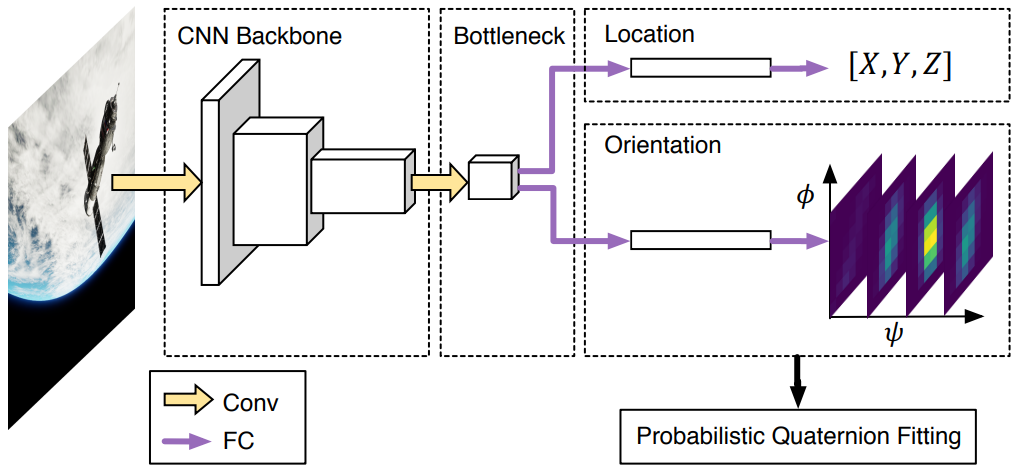}
	\caption[Direct end-to-end approach for spacecraft pose estimation. ]{Direct end-to-end approach for spacecraft pose estimation. The position is regressed directly and the orientation is obtained with soft classification~\cite{proenca2020}}.
	\label{fig:urso}
\end{figure}

Proença et al.~\cite{proenca2020} propose URSONet, a ResNet-based backbone architecture followed by two separate branches for the estimation of the position and orientation (see \Cref{fig:urso}). The position estimation was carried out through a simple regression branch with two fully connected layers while minimising the relative error in the loss function. A continuous orientation estimation via classification with soft-assignment coding was proposed for orientation estimation. Each ground truth label is encoded as a Gaussian random variable in the orientation discrete output space. The network was then trained to output the probability mass function corresponding to the actual orientation. Poss et al.~\cite{posso2022mobileurso} presented Mobile-URSONet, a mobile-friendly deployable lightweight version of the URSONet. The ResNet backbone was replaced with a MobileNetv2~\cite{sandler2018mobilenetv2} model, and the number of fully connected layers in the sub-branches was reduced to one (from two). It reduced the number of parameters to a range of 2.2M to 7.4M, 13 times smaller than the URSONet. Moreover, this was achieved without a considerable degradation in performance.

Recently, Park et al.~\cite{park2022robust} presented SPNv2, improving on the original SPN~\cite{sharma2019pose} for addressing the domain gap problem. SPNv2 has a multi-scale multi-task network architecture with a shared feature extractor following the EfficientPose~\cite{bukschat2020efficientpose} network, which is based on the EfficientDet\cite{tan2020efficientdet} feature encoder comprised of an EfficientNet\cite{tan2019efficientnet} backbone and a Bi-directional FPN (BiFPN)~\cite{tan2020efficientdet} for multi-scale feature fusion. This is followed by multiple prediction heads for each of the tasks learned: binary classification of spacecraft presence, bounding box prediction, target position and orientation estimation, keypoint heatmap regression and pixel-wise binary segmentation of the spacecraft foreground. The results show that joint multi-task learning helps in domain generalisation by preventing the shared feature extractor from learning task-specific features. The authors also propose an online domain refinement (ODR) using target domain images (without labels) to be performed on board spacecraft. The ODR fine-tunes SPNv2 on the target images by minimising the Shannon entropy~\cite{shannon1948mathematical} on the segmentation task prediction head. The paper also presents different variants of the algorithm by changing the number of parameters in the EfficientNet backbone. The smallest variant with 3.8M parameters has comparable performance to the best-performing variant with 52.5M parameters on the SPEED+ synthetic dataset.  

\begin{figure}[!t]
	\centering
	\includegraphics[scale=.45]{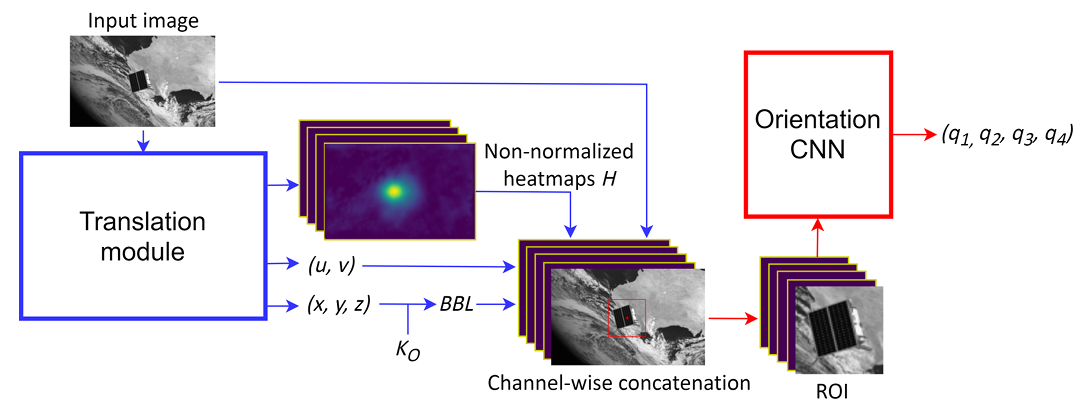}
	\caption{LSPnet architecture for spacecraft pose estimation~\cite{garcia2021lspnet}}.
	\label{fig:LSPNet}
\end{figure}

Garcia et al.~\cite{garcia2021lspnet} presented a network architecture with two CNN modules: the translation and orientation modules, for pose estimation (see \Cref{fig:LSPNet}). The translation module has a UNet architecture~\cite{ronneberger2015u} for predicting the 3D position $[x,y,z]$ of the target (from the intermediate feature embedding layer) and the 2D spacecraft location in the image $[u,v]$ (from the final heatmap output). This is then used to generate the enclosing bounding box for the spacecraft and the RoI is cropped out. The orientation module with a CNN regression network predicts the spacecraft orientation $[q_0,q_1,q_2,q_3]$ from the cropped RoI.

Finally, Musallam et al. evaluated their state-of-the-art absolute pose regression network E-PoseNet~\cite{Musallam_2022_CVPR} on the SPEED dataset. The model is based on the PoseNet architecture~\cite{Kendall_2015_ICCV}, where the backbone is replaced by a SE(2)-equivariant ResNet18 backbone~\cite{NEURIPS2019_45d6637b}. The equivariant features encode more geometric information about the input image. Moreover, equivariance to planar transformations
constrains the network in a way that can aid generalization, especially due to the weights sharing. Finally, the rotation-equivariant ResNet shows a significant reduction in model size compared to the regular ResNet architecture, to obtain the same feature size.

\subsection{Algorithm Comparison}
\label{sec:algo_comparison}
\begin{figure*}[!t]
    \centering
    \begin{subfigure}[b]{0.47\textwidth}
        \includegraphics[width=\textwidth]{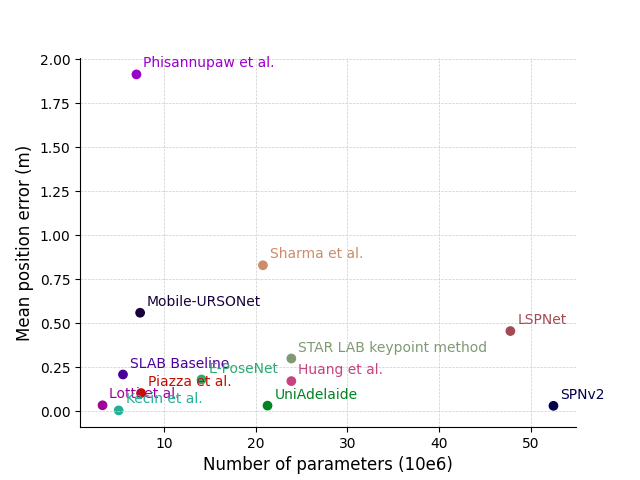}
        \caption{}
        \label{}
    \end{subfigure}
    \begin{subfigure}[b]{0.47\textwidth}
        \includegraphics[width=\textwidth]{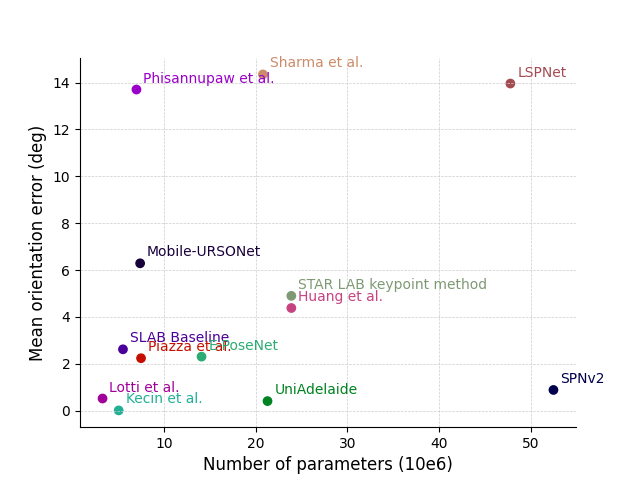}
        \caption{}
        \label{fig:errorvsparams}
    \end{subfigure}
    \caption{Comparison of pose estimation algorithms in terms of number of parameters versus (a) position error and (b) orientation error.}
    \label{fig:errorvsparams}
\end{figure*}

In this section, different spacecraft pose estimation algorithms are compared.
\Cref{table:hybrid_approach_review} and \Cref{table:direct_approach_review} summarise different hybrid and direct algorithms, respectively, with a comparison of DL models used, the total number of parameters, and the pose accuracy. 
The performance of the pose estimation algorithm is expressed in terms of the mean position and orientation errors. The position error is calculated as:
\begin{equation}
\label{eq:mean_position_error}
E_{t} = ||t_{\text{predicted}} - t_{\text{groundtruth}}||_2
\end{equation}
\noindent and the orientation error is calculated as: 
\begin{equation}
\label{eq:mean_orientation_error}
E_{R} = 2* \mathrm{arccos}\left(|<q_{\text{predicted}}, q_{\text{groundtruth}}>|\right)
\end{equation}

\noindent where, $t_{\text{predicted}}$, $t_{\text{groundtruth}}$ are the predicted and the ground truth translation vectors and $q_{\text{predicted}}$, $q_{\text{groundtruth}}$ are the predicted and the ground truth rotation quaternions respectively. $|<,>|$ indicates the absolute value of the vector dot product and $\|_2$ is the Euclidean norm. The mean position and orientation error values on the SPEED~\cite{kisantal2020} synthetic test set are reported where available~\cite{kisantal2020}. In other cases, the error values on the corresponding synthetic published dataset are reported. Similarly, in many instances, authors do not report the total number of parameters in their algorithms. In such cases, an approximate number of parameters is estimated based on the known backbone models and frameworks used. This survey is the first attempt to compare different DL-based spacecraft pose estimation algorithms in terms of performance reported on different datasets and the number of model parameters with available information in the literature. 

A key aspect of the spacecraft pose estimation algorithms is the deployment on edge devices for their use in space. Unlike the commonly used resource-abundant workstations, computing resources are scarce in space systems. Hence, deploying large DL models that have a very large number of parameters is difficult. On the other hand, using smaller DL models with a lower number of parameters leads to a drop in performance. Thus, a trade-off is needed between the use of large, high-performing models and smaller, deployable models. Based on \Cref{table:hybrid_approach_review} and \Cref{table:direct_approach_review},  \Cref{fig:errorvsparams} shows this trade-off by plotting the algorithm performance against the total number of model parameters. The results show that the algorithms \cite{lotti2022deep,piazza2021deep,kecenli2022} and the SLAB Baseline \cite{park2019} provide a good trade-off in terms of the performance and the number of parameters. 


Another factor of comparison for algorithms is the modular nature of the approaches themselves. The hybrid algorithms are built by integrating three components: spacecraft localisation, keypoint regression and pose computation. This helps to work and improve each stage of the algorithms in isolation. For example, changes in the camera model can be incorporated into the pose computation stage without retraining the localisation and keypoint regression models. This provides more flexibility in building the algorithms for different pose estimation applications. By contrast, the direct algorithms comprise only a single DL model trained end-to-end. The entire model has to be retrained to incorporate changes such as changes in camera parameters. 

In terms of performance comparison between the approaches, analysis of the top-10 methods from the first edition of ESA Kelvin Satellite Pose Estimation Challenge (KSPEC'19)~\cite{kisantal2020} show that the hybrid approaches perform comparatively better than the direct approaches. The hybrid and direct algorithms have mean position errors of $0.0083 \pm 0.0269$~m and $0.0328 \pm 0.0430$~m and mean orientation errors of $1.31 \pm 2.24\degree$ and $9.76 \pm 18.51\degree$, respectively. Analysis of the recently concluded second edition of the same challenge (KSPEC'21)~\cite{park2023satellite} also gives similar indications. Winning algorithms on both streams of the challenge used the hybrid approach.  

\subsection{Limitations}
\label{sec:algo_limitations}

Recently, several promising algorithms have been developed for DL-based spacecraft pose estimation using both the hybrid and the direct approaches. However, these algorithms still have several limitations that need to be considered and have room for further improvement. This section highlights these limitations with discussions on each topic. 

\subsubsection{Deployability} 
Deployability is a key aspect of any space algorithm. Despite the recent progress in spacecraft pose estimation algorithm development, the deployment remains an important open research question. The limitations of current algorithms in terms of deployability refer to the challenges of implementing these algorithms in real-world space missions. 

Among the current research works, only a small fraction of the developed algorithms are tested and evaluated on edge systems for space deployment \cite{lotti2022deep,cosmas2020,CA-SpaceNet2022}. Also, authors rarely report factors effecting algorithm deployability such as latency, inference time, memory requirements, power consumption and  computational cost. These missing details are important to understand the deployability of a model \cite{hadidi2019characterizing,baller2021deepedgebench}, on resource-constrained environment such in a space system with limited computational capabilities. 

Another limitation is the extensive use of off-the-shelf DL models and frameworks (refer to \Cref{table:hybrid_approach_review} and \Cref{table:direct_approach_review}). While these off-the-shelf models work well on a workstation, they may not be suitable for space deployment due to several reasons. Primarily,  these models are designed to work on systems with abundant resources and are computationally expensive, requiring significant processing power and memory. Secondly, these models (or certain DL layers) may not be supported \cite{Xilinxfpga} by the AI accelerators used in current space systems like FPGA-based~\cite{furano2020towards,leon2022towards} accelerators. Hence it is required to build algorithms with architectures specifically customised for space applications and hardware. 

\subsubsection{Explainability} Explainability refers to the ability to understand how an algorithm arrives at its predictions, and it is an essential factor in building trust and ensuring safety in critical applications such as space missions. This makes error analysis and troubleshooting easier. A key limitation of the current DL-based spacecraft pose estimation algorithms is their lack of explainability.   In the direct approach, the black-box nature of DL models in general~\cite{azodi2020opening} makes interpreting the errors and failures very difficult. Comparably, the hybrid approach tackles the spacecraft pose estimation problem in stages, providing better interpretability. However, these algorithms still lack capabilities such as reasoning\cite{li2018deep} or modelling the uncertainty between the input data and the predictions made \cite{wang2020survey} . 

\subsubsection{Robustness to Illumination Conditions}

\begin{figure}[!t]
	\centering
	\includegraphics[scale=.4]{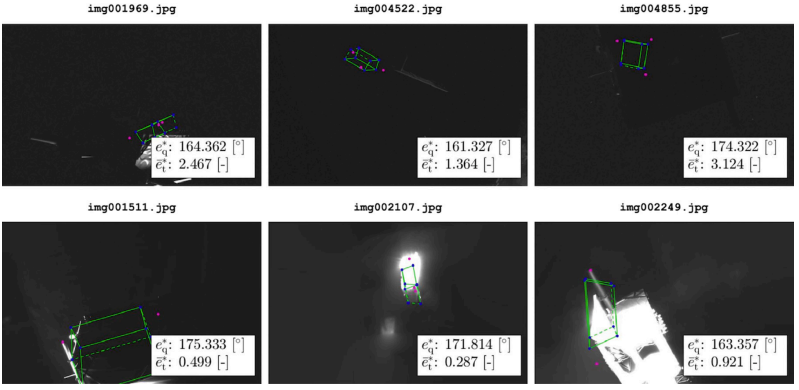}
	\caption{. Visualization of the worst 3 predictions made by stream-1 winning method of the KSPEC'21 challenge on \textit{lightbox} (top-row) and \textit{sunlamp} (bottom-row) images \cite{park2023satellite}. These results show considerable drop in accuracy of estimated poses (shown in green) under extreme lighting conditions, highlighting a important limitation of vision-based spacecraft pose estimation algorithms.}
	\label{fig:illumination_robustness}
\end{figure}

Monocular vision-based algorithms are in general sensitive to changes in lighting conditions. This can affect the accuracy and robustness of the pose estimation, especially in the dynamic illumination conditions in space. For example, shadows, reflections and sun glare can all create visual noise and make it difficult to identify and track features on the spacecraft. Analysis of the results (see \Cref{fig:illumination_robustness}) from the latest edition of KSPEC (KSPEC'21)~\cite{park2023satellite} shows that even the best vision-based spacecraft pose estimation algorithms performs poorly on images with extreme lighting conditions. 

Overcoming these limitations will require continued research and development in areas including algorithm design, evaluation protocols on edge devices, sensor technology and modelling of environmental factors. \Cref{sec:future} outlines future directions of research in spacecraft pose estimation algorithm development to address these challenges. Finally, any DL-based algorithm development cannot be separated from the question of the datasets, both for training and validating the algorithms. 
The next section (\Cref{sec:data}) presents a detailed discussion of spacecraft pose estimation datasets (\Cref{sec:datasets}) with a focus on the domain gap problem (\Cref{ssec:domain_gap}) and a discussion on their limitations (\Cref{sec:dataset_limitations}). 

\onecolumn
\begin{landscape}
\sffamily
\begin{longtable}
{|p{2.4cm}|p{3.1cm}|p{1.7cm}|p{3.1cm}|p{1.6cm}|p{1.7cm}|p{3.6cm}|p{1.5cm}|p{1.5cm}|}
\captionsetup{font=sf}
\caption{\sffamily Summary of the hybrid algorithms for spacecraft pose estimation. Details of the object detector and keypoint prediction models (including the estimated number of parameters) and the pose computation methods used are provided. The mean position and orientation error values on the SPEED synthetic test set are reported where available. In cases where the number of parameters is not reported by the authors, estimated values based on the known backbone models and frameworks are given. Additionally, the links to the publicly available algorithms are included in \Cref{ssec:algo_links}.} \label{table:hybrid_approach_review} \\
\hline
\rowcolor[HTML]{EFEFEF} \textbf{Ref}  & \textbf{Object Detector}  & \textbf{Parameters} \textbf{(millions)}    & \textbf{Keypoint Prediction}        & \textbf{Parameters} \textbf{(millions)}    & \textbf{Total Parameters} \textbf{(millions)} & \textbf{Pose Computation}            & \textbf{Mean position error (E\textsubscript{t}) \hspace{1cm} (m)} & \textbf{Mean \hspace{0.5cm} orientation error (E\textsubscript{R}) \hspace{1cm} (deg)}  \\ 
\hline
 UniAdelaide~\cite{chen2019}    & Faster-RCNN~\cite{ren2015faster} with HRNet-W18-C~\cite{wang2020deep} as the backbone  & $\sim${}21.3*~\cite{sun2019}               & Pose-HRNet-W32~\cite{sun2019deep} & $\sim${}28.5~\cite{sun2019deep} & $\sim${}49.8 
(176.2~\cite{spetpu22})       & PnP + RANSAC refined with a geometric loss optimized using SA-LMPE optimiser       & 0.0320\textsuperscript{+}           & 0.4100\textsuperscript{+}    \\ 
\hline

\rowcolor[HTML]{EFEFEF} EPFL\_cvlab~\cite{gerard2019}&  Not applied & -NA-  & Yolov3~\cite{redmon2018yolov3} with DarkNet-53~\cite{redmon2018yolov3}  as the backbone followed by a segmentation and regression decoder branchers & $\sim${}59.1\cite{long2020} & $\sim${}59.1 (89.2~\cite{spetpu22})       & EPnP~\cite{lepetit2009} +~RANSAC  & 0.0730\textsuperscript{+}            & 0.9100\textsuperscript{+}    \\ 
\hline

SLAB Baseline~\cite{park2019}               & YOLOv3~\cite{redmon2018yolov3} with MobileNetV2~\cite{sandler2018mobilenetv2} as the backbone       & 5.53             & YOLOv2~\cite{redmon2017yolo9000} with MobileNetV2~\cite{sandler2018mobilenetv2} as the backbone     & 5.64          & 11.2 & EPnP           &  0.2090\textsuperscript{+}   & 2.6200\textsuperscript{+}   \\ 
\hline

\rowcolor[HTML]{EFEFEF} Huo et al.~\cite{huo2020}                & Tiny-YOLOv3~\cite{redmon2018yolov3} architecture** with a detection subnetwork     & -NA-       & Tiny-YOLOv3~\cite{redmon2018yolov3} architecture** with a regression subnetwork           & -NA- & $\sim${}0.89  & PnP+RANSAC refined with a Log-cosh geometric loss optimized by Levenberg-Marquardt solver~\cite{more1978levenberg} & 0.0320  &    0.6812   \\ 
\hline

Piazza et al.~\cite{piazza2021deep}  & YOLOv5 & 7.5  & HRNet32~\cite{sun2019deep} & $\sim${}28.6*~\cite{cheng2020} & $\sim${}36.1 & EPnP refined with a geometric loss optimised by Levenberg-Marquardt solver & ~0.1036 & 2.2400 \\ 
\hline

\rowcolor[HTML]{EFEFEF} Huan et al.~\cite{huan2020pose} & Cascade Mask R-CNN~\cite{cai2018cascade} with HRNet as backbone & -NA-  & HRNet~\cite{sun2019deep} & $\sim${}28.5 to $\sim${}63.6~\cite{sun2019deep} & -NA- & EPnP refined with a Huber style geometric loss optimised as non-linear least-squares problem  & 0.1823 &  2.8723 \\ \hline

 STAR LAB keypoint method~\cite{rathinam2020}          & Faster-RCNN~\cite{ren2015faster} with RestNet50~\cite{he2016deep} backbone           & $\sim${}23.9*~\cite{leong2020semi}          & HigherHRNet~\cite{cheng2020} with HRNet-W32~\cite{sun2019deep} as the backbone   & $\sim${}28.6*~\cite{cheng2020} & $\sim${}54.2       & PnP + RANSAC   & 0.3000 (URSO-OrViS dataset)            & 4.9000 (URSO-OrViS dataset)     \\ 
\hline

\rowcolor[HTML]{EFEFEF} Black et al.~\cite{black2021real}  & SSD~\cite{liu2016ssd} MobileNetV2~\cite{sandler2018mobilenetv2}  & -NA-  & MobilePose~\cite{hou2020mobilepose} architecture with MobileNetV2~\cite{sandler2018mobilenetv2} as backbone & -NA- &  6.9 & EPnP + RANSAC & 1.0800 (Cygnus dataset) & 6.4500 (Cygnus dataset)  \\ 
\hline

Wide-Depth-Range~\cite{hu2021}      & Not applied & -NA- & FPN~\cite{lin2017feature} architecture with DarkNet-53~\cite{redmon2018yolov3} as the backbone        & 51.5          & 51.5  & PnP + RANAC with and without a pose refinement strategy      & -NA-  & -NA-      \\ 
\hline

\rowcolor[HTML]{EFEFEF} Cosmas et al.~\cite{cosmas2020}\dag           & YOLOv3~\cite{redmon2018yolov3}    & $\sim${}59.1*~\cite{long2020pp}        & ResNet34-UNet~\cite{he2016deep, ronneberger2015u} architecture     & $\sim${}21.5*~\cite{leong2020} & $\sim${}80.6       & -NA- & -NA-           & -NA-      \\
\hline

Lotti et al.~\cite{spetpu22}\dag & MobileDet~\cite{xiong2021mobiledets} & 3.3 & Regression head with an EfficientNet-Lite~\cite{tan2019efficientnet} backbone & -NA- & 15.4 & EPnP + RANSAC optimised by Levenberg-Marquardt solver & 0.0340 & 0.5200 \\
\hline

\rowcolor[HTML]{EFEFEF} Kecen et al.~\cite{kecenli2022}\dag & YOLOX-Tiny~\cite{ge2021yolox} & $\sim${}5.06~\cite{ge2021yolox} & FPN~\cite{lin2017feature} architecture with CSPDarknet53~\cite{bochkovskiy2020yolov4} as the backbone & $\sim${}27.6~\cite{bochkovskiy2020yolov4} & $\sim${}32.66 & EPnP  & 0.0049 & 0.0129 \\
\hline

CA-SpaceNet~\cite{CA-SpaceNet2022} & Not used & -NA-  & Keypoint prediction head having three FPNs~\cite{lin2017feature} with two DarkNet-53~\cite{redmon2018yolov3} networks as the backbones  & -NA- & 51.29 M \dag & PnP & -NA- & -NA- \\
\hline

\rowcolor[HTML]{EFEFEF} Legrand et al.~\cite{legrandend}\dag & An ideal object detector assumed & -NA- & DarkNet-53~\cite{redmon2018yolov3} pre-trained on Linemod~\cite{Hinterstoier2012ModelBT} with two decoding heads - a segmentation head and a regression head & 71.2 & -NA- & PIN architecture~\cite{hu2020single} consists of an MLP that aggregates local features per keypoint into a single representation  & 0.201 & 4.687 \\
\hline
\end{longtable}
\raggedright 
\sffamily
\textsuperscript{+} Results from {KSPEC} first edition~\cite{kisantal2020} \\ 
**Backbone shared between the object detector and the keypoint prediction model \\
\dag Best performing variant considered \\
\end{landscape}

\begin{landscape}
\begin{table}
\sffamily
\caption{\sffamily Summary of direct end-to-end algorithms for spacecraft pose estimation. Details of the network architectures used, along with an estimated number of parameters, are presented. The error values on the SPEED synthetic test set are reported where available. In cases, the number of parameters is not reported by the authors, an estimated number of parameters based on the backbone models used are given. Additionally, the links to the publicly available algorithms are included in \Cref{ssec:algo_links}.}
\label{table:direct_approach_review} 
\begin{tabular}{|p{2.5cm}|p{14.2cm}|p{1.7cm}|p{1.7cm}|p{1.7cm}|}
\hline
\rowcolor[HTML]{EFEFEF} \textbf{Reference}  & \multicolumn{1}{c|}{\textbf{\textbf{Model architecture}}} & \textbf{Parameters \hspace{0.5cm}(millions)} &  \textbf{Mean position error (E\textsubscript{t}) \hspace{0.5cm} (m)}  & \textbf{Mean rotation error (E\textsubscript{R}) \hspace{0.5cm} (deg)}  
\\ 
\hline 
Sharma et al.~\cite{sharma2018pose}\dag & AlexNet~\cite{Krizhevsky2012ImageNetCW} with half as many kernels per layer as the original AlexNet architecture, with the last fully connected layer containing as many neurons as the number of pose labels & $\sim${}20.8 & 0.83 (Imitation-25 dataset) & 14.35 (Imitation-25 dataset) \\
\hline 
 \rowcolor[HTML]{EFEFEF} SPN~\cite{sharma2019pose}  & A 5-layer CNN with 3 sub-branches for bounding box classification and regression, relative orientation classification and relative orientation weights regression. & -NA- & 0.7832 & 8.4254  \\ 
\hline
SPNv2\cite{park2022robust}\dag & Bi-directional
Feature Pyramid Network (BiFPN)~\cite{tan2020efficientdet} with EfficientNet~\cite{tan2019efficientnet} backbone and with multi-task head networks shared by the features at all scales. & 52.5 & 0.031 (SPEED+) & 0.885 (SPEED+) \\
\hline

\rowcolor[HTML]{EFEFEF} URSONet~\cite{proenca2020} & ResNet18, ResNet34, ResNet50, ResNet101~\cite{he2016deep} base networks with 2 sub-branch networks for position regression and  probabilistic orientation estimation via soft classification.  & $\sim${}11.4 to $\sim${}42.8~\cite{leong2020semi} ($\sim${}500**) & 0.1450\textsuperscript{+}  & 2.4900\textsuperscript{+}   \\ 
\hline

Mobile-URSONet~\cite{posso2022mobileurso}\dag & MobileNet-v2~\cite{sandler2018mobilenetv2} based network, pre-trained on ImageNet~\cite{deng2009imagenet}, with 2 sub-branches for position regression and probabilistic orientation estimation via soft classification. & 7.4 & 0.5600 & 6.2900 \\
\hline

\rowcolor[HTML]{EFEFEF} LSPnet~\cite{garcia2021lspnet}  &  ResNet50~\cite{he2016deep} base architecture  for position regression followed by an up-sampling CNN for object localisation and a second ResNet50 for orientation regression.   &                  $\sim${}47.8~\cite{leong2020semi}   &  0.4560   &  13.9600 \\ 
\hline

Huang et al.~\cite{huang2021non}   & ResNet50~\cite{he2016deep} base network with 3 sub-branch networks for object detection, position regression and orientating soft classification. &  $\sim${}23.9~\cite{leong2020semi}  & 0.1715 (URSO-OrViS datast)  &    4.3820 (URSO-OrViS dataset)  \\ 
\hline

\rowcolor[HTML]{EFEFEF} Phisannupawong et al.~\cite{phisannupawong2020vision} &  A modified version of GoogLeNet~\cite{szegedy2015going} that forms a general pose estimation model as implemented in PoseNet~\cite{PoseNet2016}. The softmax classifiers in the original GoogLeNet were replaced with affine regressors and each fully connected layer was modified to output a 7D pose vector. & $\sim${}7.0 & 1.1915\textsuperscript{\#} (URSO-OrViS dataset) & 13.7043\textsuperscript{\#}  (URSO-OrViS dataset) \\
\hline

E-PoseNet~\cite{Musallam_2022_CVPR} & PoseNet architecture~\cite{Kendall_2015_ICCV} with SE(2)-equivariant ResNet18 backbone~\cite{NEURIPS2019_45d6637b}.    &                  14.1   &  0.1806   &  2.3073 \\ 
\hline
\end{tabular}
\end{table}
\raggedright 
\vspace{.2cm} 
\sffamily \dag Details of the best-performing variant reported  \\
\sffamily \textsuperscript{+}Results from {KSPEC} first edition~\cite{kisantal2020}. \\  
\sffamily **Number of parameters in the best performing ensemble of models reported by the authors \\ 
\textsuperscript{\#}Median values reported  \\ 
\end{landscape}
\twocolumn

\begin{table*}[ht]
    \centering
\caption{Review of recent spacecraft pose estimation datasets, sorted by year. The \textit{Syn/Lab/Space} column is, the number of synthetic, lab and space-borne images in the dataset, respectively. The \textit{Spacecraft} column specifies the spacecraft used in the dataset. The \textit{resolution} column corresponds to the width x height of the images, in pixels. The \textit{I} column indicates if the images are RGB (\textbf{C}) or grey-scale (\textbf{G}). The \textit{Range} column indicates the distance between the camera and the spacecraft. The \textit{Tools} column is a list of the rendering software used to generate the synthetic data. 
Additionally, the links to the publicly available datasets are included in \Cref{ssec:datasets_links}.}
\label{table:dataset}    
\begin{tabular}{*{11}{|c}}
        \hline
        \rowcolor[HTML]{EFEFEF} \textbf{Dataset} 						& \textbf{Year} & \textbf{Syn/Lab/Space} & \textbf{Spacecraft} & \textbf{Resolution} & \textbf{I} & \textbf{Range} & \textbf{Tools}  \\ \hline
        SHIRT~\cite{park2022cnnukf} 	& 2022 & 5k/5k/- & Tango & 1920x1200 & \textbf{G} & $\leq$8m & OpenGL  \\ \hline
        SPARK2-Stream2~\cite{rathinam_arunkumar_2022_6599762} 	& 2022 & 30k / 900 / - & Proba-2 & 1440 x 1080  & \textbf{C} & [1.5m,10m] & Blender \\ \hline
        COSMO~\cite{lotti2022deep} 		& 2022 & 15k / - / - & COSMO-SkyMed & 1920 x 1200  & \textbf{C} & [36m,70m] & Blender  \\ \hline
        SwissCube~\cite{hu2021} 		& 2021 & 50k / - / - & SwissCube & 1024 x 1024  & \textbf{C} & [0.1m, 1m] & Mitsuba 2 \\ \hline
        SPEED+~\cite{park2021} 			& 2021 & 60k / 10k / - & Tango & 1920 × 1200  & \textbf{G} & $\leq$ 10m & OpenGL  \\ \hline
        Cygnus~\cite{black2021real}	 	& 2021 & 20k / - / 540 & Cygnus & 1024 × 1024  & \textbf{C} & [35m,75m] & Blender   \\ \hline
        SPEED~\cite{kisantal2020}		& 2020 & 15k / 305 /- & Tango & 1920 × 1200  & \textbf{G} & [3m,40.5m] & OpenGL  \\ \hline
        URSO~\cite{proenca2020} & 2019 & 15k / - / - & Dragon, Soyuz & 1080 x 960  & \textbf{C} & [10m,40m] & Unreal Engine 4 \\ \hline
        PRISMA12K~\cite{park2019}		 & 2019 & 12k / - / - & Tango & 752 x 580 & \textbf{G} & - & OpenGL  \\ \hline
        PRISMA12K-TR~\cite{park2019}		 & 2019 & 12k / - / - & Tango & 752 x 580 & \textbf{G} & - & OpenGL \\ \hline
        Sharma et. al.~\cite{sharma2018pose} & 2018 & 500k / - / - & Tango & 227 x 227 & \textbf{C} & [3m,12m] & OpenGL  \\ \hline
    \end{tabular}
\end{table*}

\section{Datasets}
\label{sec:data}

The use of DL models in spacecraft pose estimation necessitates proper training to achieve the robust performance demanded by space applications. The quality of the datasets is likely equally influential in DL model performance compared to designing an effective DL algorithm to reach the intended performance. Large datasets \cite{DBLP:conf/eccv/LinMBHPRDZ14,DBLP:journals/ijcv/RussakovskyDSKS15} with a wide range of application scenarios are usually considered to train DL models, which helps them generalise well for unseen scenarios. Though DL algorithms are evolving towards few-shot \cite{song2022comprehensive} and zero-shot \cite{cao2020research} learning, solving 6 DoF pose prediction problems with high accuracy still depends on large datasets with images spanning a wide range of scenarios \cite{DBLP:journals/ral/RennieSBS16,DBLP:conf/rss/XiangSNF18}. 

Currently, there is a lack of publicly available space-borne image datasets. This limits the application of DL models and their validation to specific targets where actual space-borne images are available and to a limited range of operation scenarios. To overcome this limitation, image rendering tools are the preferred way to generate realistic space-borne images and testbeds are considered for on-ground validation. The rendering tools help generate thousands of images for a wide range of targets with annotations for any user-defined applications such as object detection, semantic segmentation and 6 DoF pose estimation. These generation tools also provide a lot of flexibility to adapt parameters such as camera models, orbital lighting conditions, etc., depending on the final use-case application. 

Spacecraft pose estimation algorithms are usually part of vision-based navigation systems and are validated in a dedicated testbed facility that can simulate the orbital relative motion using robotic arms \cite{pauly2022lessons,park2021robotic} or air-bearing \cite{SABATINI2015184} platforms under realistic space lighting conditions. The target mock-up used in such facilities will be scaled or original depending on various factors, including the size of the facility, mock-up size, application scenario,  etc. While synthetic imagery can be mass-produced to address any requirements, the images produced from testbed scenarios are limited to a certain extent. It includes the Earth in the background, the accurate position of the sun, earth's albedo; such characteristics differentiate the lab/testbed imagery from the actual space imagery. 

From the above discussion, it is evident that the spacecraft pose estimation deals with images from three domains (i.e., synthetic, lab and actual space imagery) during the development, testing/validation and deployment phases. It is the nature of the DL models to overfit the model to the features specific to the training domain, and this challenge is well-known in the literature as \textit{domain gap} \cite{fang2022source,wang2018deep} problem. So, the algorithms need to consider the aspect of domain generalisation from the data viewpoint to improve the algorithm's performance.

\begin{figure*}[!t]
    \centering
    \begin{subfigure}{.4\textwidth}
		\centering
		\includegraphics[width=\textwidth]{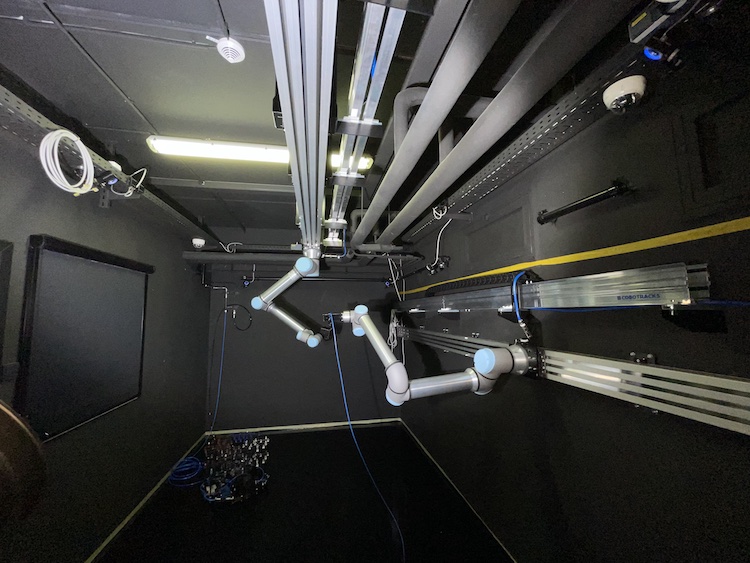}
		\caption{}
		\label{fig:lab_infrastructure-1}
	\end{subfigure}
	\begin{subfigure}{.4\textwidth}
		\centering
		\includegraphics[scale=.162, trim={10cm 0 10cm 0},clip]{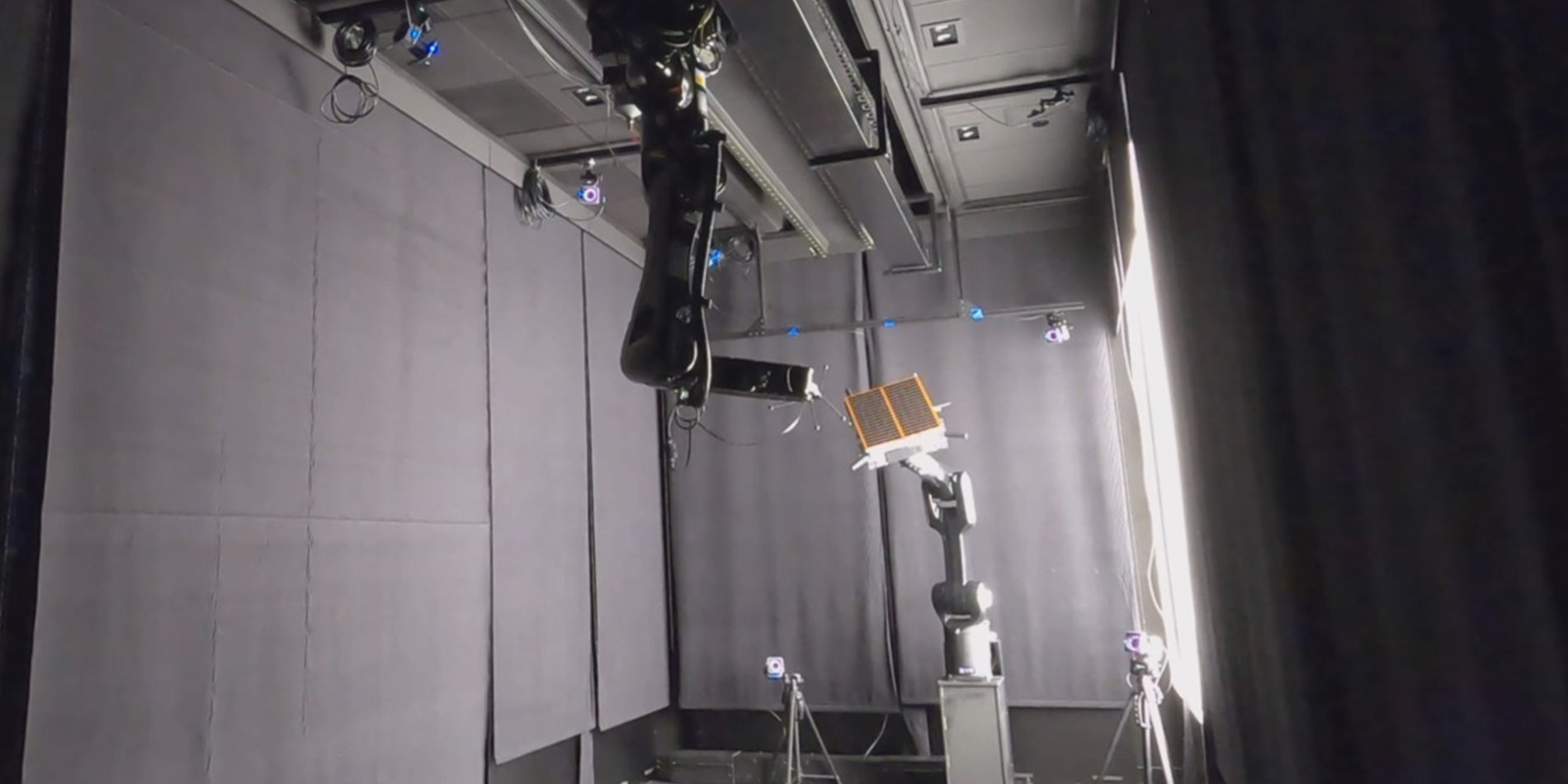}
		\caption{}
		\label{fig:lab_infrastructure-2}
	\end{subfigure}
    \caption{(a) SnT Zero-G Lab at the University of Luxembourg~\cite{pauly2022lessons} (b) TRON facility at Stanford University~\cite{park2021robotic}}
    \label{fig:lab_infrastructure}
\end{figure*}


\subsection{Summary of Datasets, Simulators \& Testbeds} \label{sec:datasets}

This section provides a summary of the spacecraft pose estimation datasets, simulators and rendering tools for synthetic image generation, and testbeds for validation.

\textbf{Datasets:}
\Cref{table:dataset} summarises the properties of the major spacecraft pose estimation datasets. The properties of the datasets include the number of images, the target spacecraft model, image resolution, annotations and the rendering tools used for the synthetic image generation. The number of images in the currently available spacecraft pose estimation datasets is between $10^4$ and $10^5$. This is relatively low compared to some typical datasets used for other machine learning tasks such as image classification and object detection. The COCO \cite{lin2014microsoft} dataset, one of the standard datasets used for  object detection, contains \textasciitilde{}300k images. ImageNet dataset \cite{deng2009imagenet} primarily used for classification contains \textasciitilde{}14M images. Similarly, YCB dataset \cite{DBLP:conf/rss/XiangSNF18}, a recent generic dataset for 6 DoF pose estimation, has \textasciitilde{}133k images. 

The target spacecraft model used in the datasets also plays a vital role in determining the dataset characteristics. For example, a smaller target size will lead to a smaller operation range and vice-versa. The TANGO satellite \cite{tango} model used in the multiple datasets \cite{park2021,park2022cnnukf,kisantal2020,park2019,park2019,sharma2018pose} has a coarse dimension of 80×75×32cm will lead to the operation range of \textasciitilde{}10m. However, for Soyuz or Cygnus models in other datasets increases the operating range to 40 \textasciitilde{} 80 m. Similar constraints will apply to testbed data as well. A 1:1 mockup scale of TANGO spacecraft in SPEED+ \cite{park2021} leads to a lower range in the lab-generated images due to the size constraint of the facility. Usually, a scaled mock-up is considered a solution to increase the validation range in the testbed scenarios.




The level of annotations may vary for different datasets; for spacecraft pose estimation applications, each image in the dataset must be appropriately annotated with corresponding relative 6DoF pose labels. All the datasets mentioned in \Cref{table:dataset} are adequately annotated with 6DoF pose labels. However, the hybrid algorithm approach discussed in \Cref{sec:hybrid_algorithms} demands secondary annotations such as the bounding boxes and the keypoints. To recover the secondary annotations from pose labels, it is necessary to have 3D information on the edges or vertices of the target. Even for the standard datasets such as SPEED \cite{kisantal2020} and SPEED+ \cite{park2021}, the only way to use a hybrid approach is to recover the 3D locations of interested keypoints is via the 3D reconstruction methods \cite{chen2019}. These recovered keypoints will be used to construct secondary annotations such as bounding boxes, keypoints, segmentation masks and even ellipse heatmap annotations \cite{gaudilliere20233d}. The lack of secondary annotations can be an issue for multi-task learning approaches where the annotations (such as segmentation masks) could be used to define auxiliary tasks intended to prevent learning domain-specific features to improve generalisation \cite{park2022robust}. Several learning-based approaches are evolving to generate secondary annotation to address the label scarcity, such as depth estimation using a single image depth estimator~\cite{mertan2022single} and an image segmentation technique~\cite{yuheng2017image}. Some self-supervised approaches are evolving as an alternate way to get bounding box annotations for a single target in the image \cite{wang2022self}. Though these approaches aid annotations, they cannot replace the properly calibrated annotations recorded during synthetic data generation.


\textbf{Simulators and Rendering Tools:} Computer graphics allow us to create realistic images of objects based on high-quality textures using ray tracing. Ray Tracing techniques mimic how light interacts with the real world and rely on evaluating and simulating the path of view lines from the observer camera to objects in the field of view. This simulation enables the calculation of the light intensity of associated pixels. Several efforts were made towards creating simulators for space applications. Realistic image simulation tools were used in previous missions to aid vision-based navigation in space/planetary environments (such as the Lunar environment, Asteroid surface), and it includes the PANGU (Planet and Asteroid Natural scene Generation Utility) \cite{martin2019planetary} and the SurRender \cite{brochard2018scientific} by Airbus.  The University of Dundee has developed the PANGU simulation tool, which generates realistic, high-quality, synthetic surface images of planets and asteroids. PANGU uses a custom GPU-based renderer to render the scene. Airbus’s Surrender can be used in two modes of image rendering, ray tracing and OpenGL\cite{shreiner2009opengl}. It can produce physically accurate images providing the known irradiance (each pixel contains an irradiance value expressed in W/m2). Other general rendering tools such as Blender~\cite{lotti2022deep}, Unreal Engine~\cite{proenca2020}, and Mitsuba~\cite{hu2021} were also used to generate synthetic images. The main issue with these tools is that they are designed for general usage and are not customised for space imagery. A brief comparison of rendering tools for synthetic imagery was provided in \cite{arunkumar2021}. Recently, efforts \cite{kisantal2020}, \cite{bechini2022spacecraft} have been made toward developing simulation tools specific for the purpose of synthetic image generation for spacecraft pose estimation. SPEED and SPEED+ images are obtained using the Optical Simulator \cite{beierle2019variable}, based on an OpenGL rendering pipeline. The images from SPEED and SPEED+ are validated against the real images of TANGO spacecraft from the PRISMA mission using histogram comparison \cite{kisantal2020}. However, to our knowledge, no tool can be considered a de facto standard to generate space imagery for spacecraft pose estimation.

\textbf{Testbeds:}
In spacecraft pose estimation, collecting images from space for training and evaluating algorithms is extremely difficult and expensive. Laboratory testbeds (see \Cref{fig:lab_infrastructure}) are considered as an alternative to replicate relative motion and orbital lighting conditions. \Cref{tab:testbeds}, summarises different laboratory testbed facilities based on their size, manipulation capabilities, tracking systems, 
perception sensors and orbital motion simulations. Some of the SoTA testbed includes The Robotic Testbed for Rendezvous and Optical Navigation (TRON) at Stanford's Space Rendezvous Laboratory (SLAB)~\cite{park2021robotic}, STAR Lab at the University of Surrey~\cite{arunkumar2021}, SnT Zero-G Lab at the University of Luxembourg~\cite{pauly2022lessons}, GMV Platform-Art~\cite{COLMENAREJO2019206, GMVplatformart}, German Aerospace Center European Proximity Operations Simulator 2.0 (DLR EPOS)~\cite{Benninghoff2017EuropeanPO} and European Space Agency's GNC Rendezvous, Approach and Landing Simulator (GRALS)~\cite{cassinis2021ground}. These testbeds generally have robotic manipulators to carry the payloads. The payloads can be different target spacecraft mock-up models or mounted cameras mimicking a chaser. Different lighting equipment has been used for simulating space conditions. For instance, in SPEED+~\cite{park2021}, the images collected in a sunlamp setup replicate the sun's bright light, and those collected with a lightbox setup emulate the diffused light of the earth's albedo, respectively. Motion capture systems are extensively used to collect pose labels based on the reflective markers attached to the target and cameras. However, these motion capture systems should be carefully calibrated \cite{park2021robotic} to generate accurate ground truth data, which can be tedious and time-consuming. 

The next section discusses the major issue with the current spacecraft pose estimation datasets: the domain gap between synthetic data used for training and the real laboratory/ space-borne images used for testing/ validating and deploying the DL-based algorithms.

\begin{table*}[]
\caption{Summary of different laboratory testbed facilities for evaluating spacecraft pose estimation algorithms}
\label{tab:testbeds}
\begin{tabular}{|p{1.5cm}|p{2.2cm}|p{2.2cm}|p{2.3cm}|p{2.45cm}|p{2.4cm}|p{1.7cm}|} \hline

\rowcolor[HTML]{EFEFEF} \textbf{Facility} & \textbf{STAR Lab~\cite{arunkumar2021}} & \textbf{ TRON~\cite{park2021robotic}} & \textbf{SnT Zero-G Lab~\cite{pauly2022lessons}} & \textbf{GMV Platform-Art\textsuperscript{\copyright}~\cite{COLMENAREJO2019206}\cite{ GMVplatformart}} & \textbf{DLR EPOS 2.0~\cite{Benninghoff2017EuropeanPO}} & \textbf{GRALS~\cite{cassinis2021ground}} \\ \hline

\textbf{Illumination} & 
\begin{tabular}[c]{@{}l@{}}\textbullet~Forza 500 \\ LED spotlight \end{tabular}
& 
\begin{tabular}[c]{@{}l@{}}\textbullet~
LED panels \\ (for  diffused light) \\
\textbullet~Metal-halide \\arc lamp \\(for sunlight) 
\end{tabular}
& 
\begin{tabular}[c]{@{}l@{}}\textbullet~
Godox SL-60 \\ LED Video Light\\
\textbullet~Aputure LS \\ 600d Pro 
\end{tabular}
&
\begin{tabular}[c]{@{}l@{}}\textbullet~Numerically \\ controlled \\ Sun emulator 
\end{tabular}
& 
\begin{tabular}[c]{@{}l@{}}\textbullet~Osram ARRI \\ Max 12/18 \\ (with a 12 kW \\ hydrargyrum \\  medium-arc \\ iodide  lamp)
\end{tabular}
& 
\begin{tabular}[c]{@{}l@{}}\textbullet~
Dimmable, \\ uniform and \\ collimated \\ light source 
\end{tabular}
\\ \hline

\textbf{Perception Sensor} & 
\begin{tabular}[c]{@{}l@{}}\textbullet~FLIR Blackfly \\ (monocular \\ camera)\\ \textbullet~2D/3D LIDAR    \\ \textbullet~Intel RealSense\\ D435i\\ (RGBD camera)\end{tabular} 
& \begin{tabular}[c]{@{}l@{}} \textbullet~Point Grey \\ Grasshopper 3 \\ (monocular  \\ camera) \end{tabular} &
\begin{tabular}[c]{@{}l@{}}
\textbullet~FLIR Blackfly \\(monocular \\ camera) \\\textbullet~Prophesee EKv4 \\ (event camera) \\ \textbullet~Intel RealSense \\ D435i \\ (RGBD camera)
\end{tabular}
&
\begin{tabular}[c]{@{}l@{}}
\textbullet~Optical \\ navigation camera \\
\textbullet~Industrial laser \\ sensor \\
\textbullet~A set of GPS-like \\ pseudolites \\
\end{tabular}
& 
\begin{tabular}[c]{@{}l@{}}
\textbullet~Prosilica \\ GC-655M  \\ (CCD camera) \\
\textbullet~PMDtec \\ Camcube 3.0 \\ (PMD camera) \\
\textbullet~Bluetechnix \\ Argos3D-IRS1020 \\ DLR Prototype \\ (PMD LiDAR)
\end{tabular}
&\begin{tabular}[c]{@{}l@{}} \textbullet~Prosilica \\ GC2450 \\ (monocular \\ camera) 
\end{tabular}\\ \hline

\textbf{Manipulator (Robotic Arms)} & \textbullet~UR5 & \textbullet~KUKA & \textbullet~UR10e & 
\begin{tabular}[c]{@{}l@{}} \textbullet~Mitsubishi \\ PA10-6CE\\\textbullet~KUKA KR150-2\end{tabular} &
\begin{tabular}[c]{@{}l@{}}  \textbullet~KUKA \\ KR100HA\\\textbullet~KUKA KR240-2\end{tabular} &
\begin{tabular}[c]{@{}l@{}}  \textbullet~KUKA\\\textbullet~UR5\end{tabular}  \\ \hline

\textbf{Tracking System} & Qualisys & Vicon & OptiTrack & Model-based tracking algorithm based on virtual visual servoing \& Kanade-Lucas-
Tomasi (KLT)  feature tracker algorithm & VIsion BAsed NAvigation Sensor System (VIBANASS) & VICON \\ \hline

\textbf{Background Material} & Black background curtains & Light-absorbing black commando curtains & Blind made of non-reflective black textile from inside and outside & Black curtains fully covering the walls and ceiling & Black curtain and a black
wrapping of one of the robots made of Molton material & Black background curtains \\  \hline

\textbf{Simulated Operations} &
\begin{tabular}[c]{@{}l@{}}
\textbullet~Proximity \end{tabular}
& \begin{tabular}[c]{@{}l@{}}
\textbullet~Rendezvous \\ \textbullet~Proximity \end{tabular}
& \begin{tabular}[c]{@{}l@{}}
\textbullet~Proximity \\
\textbullet~Rendezvous \\
\textbullet~Orbital \\ maintenance \\ operations \end{tabular}
& \begin{tabular}[c]{@{}l@{}}
\textbullet~Rendezvous  \\\textbullet~Proximity \end{tabular}
& \begin{tabular}[c]{@{}l@{}}
\textbullet~Rendezvous \\ \textbullet~Docking/berthing 
\\ \textbullet~Proximity \end{tabular}
&  \begin{tabular}[c]{@{}l@{}} \textbullet~Rendezvous \\ \textbullet~Proximity  \end{tabular}
\\ \hline

\textbf{Dimensions (WxLxH)} & 3mx2mx2.5m & 8m×3m×3m (simulation room) and 6m (track) & 5m×3m×2.3m  & 15m & 25m (track) & 4m \\ \hline

\textbf{ROS~\cite{quigley2009ros} Supported} & Yes & -NA- & Yes & No & No & No \\ \hline

\end{tabular}
\end{table*}


\subsection{Bridging the Domain Gap}
\label{ssec:domain_gap}

Any DL-based algorithm trained on a synthetic dataset is likely to suffer from a performance drop when tested on real images (whether acquired within a ground-based laboratory or in space), which is referred to as the \textit{domain gap}~\cite{ben2006analysis} problem. Following the related computer vision terminology, the training dataset arises from a \textit{source} domain while the test dataset belongs to the \textit{target} domain. More subtly, a domain gap persists even when the real source and target datasets are acquired under different (laboratory and space) environmental conditions ~\cite{toft2020long}. To ensure the reliable performance of DL-based spacecraft pose estimation algorithms in real-world space missions, it is, therefore, crucial to bridge the domain gap. Several methods have been used in spacecraft pose estimation literature for this purpose. These methods are classified into two categories:
\begin{itemize}
	\item Data level methods: Expanding or adding diversity to the training data by applying different techniques to alter the images, such as \textit{1) data augmentation}~\cite{park2022robust} or \textit{2) domain randomization}~\cite{park2019}
	\item Algorithm level methods: Adapting the learning procedure of the model by using different techniques, such as \textit{3) multi-task learning}~\cite{PARK2023} or \textit{4) adversarial learning}~\cite{park2021}, to make the features extracted from images as less domain-dependent as possible
\end{itemize}



\subsubsection{Data Augmentation}
\label{ssec:data_augmentation}

\begin{figure*}[!t]
	\centering
	\includegraphics[width=\linewidth]{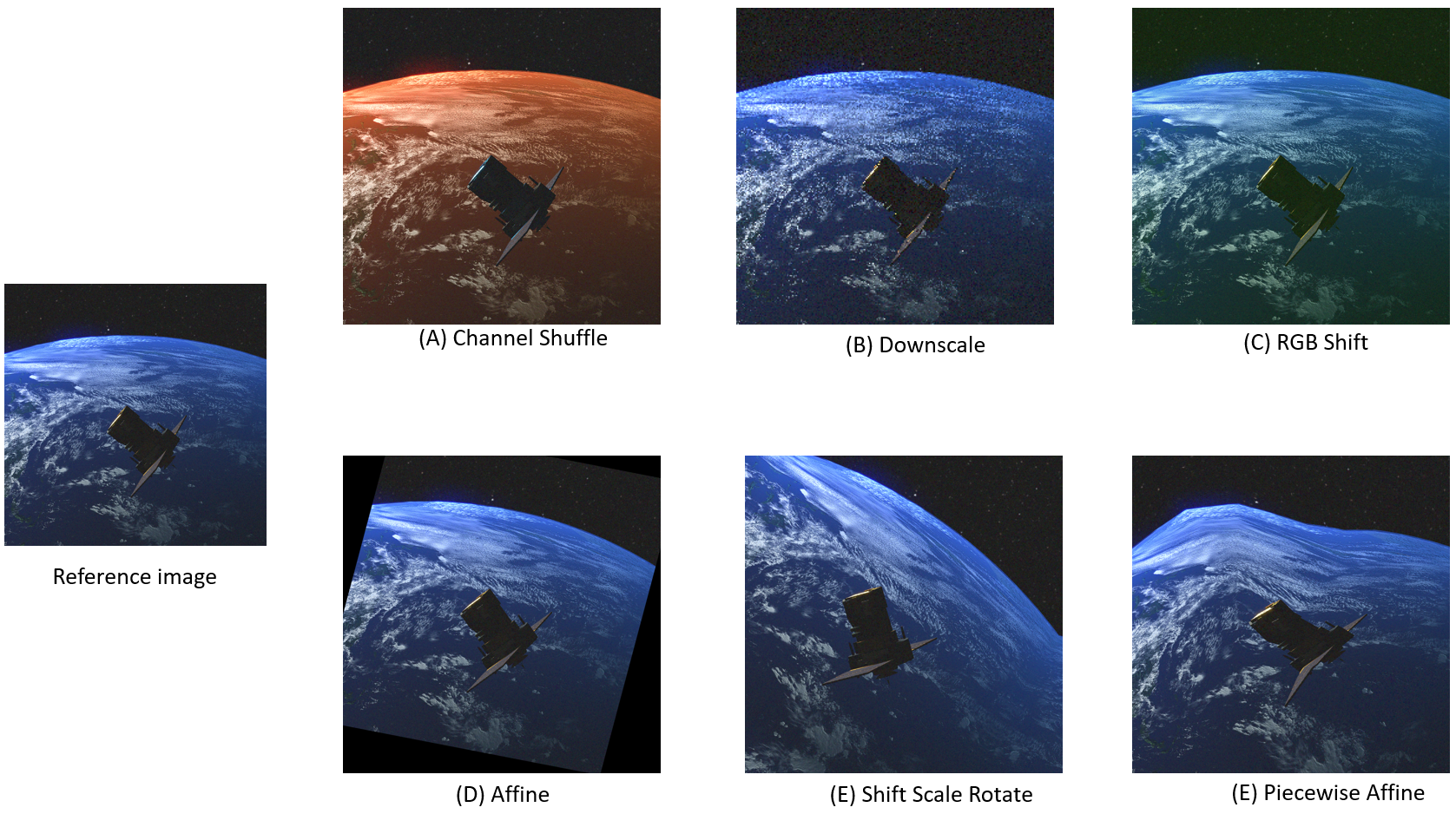}
	\caption{Illustration of different data augmentation methods used on the same reference image taken from SPARK2~\cite{rathinam_arunkumar_2022_6599762} dataset. Images A, B and C show examples of pixel-wise augmentation methods and images D, E and F show the application of spatial augmentation methods. The captions refer to the corresponding functions used by the Albumentations Python library.}
	\label{fig:data_augmentation_techniques}
\end{figure*}
 
This involves artificially creating additional training data by applying various transformations to the existing data \cite{MUMUNI2022100258}. This is done to increase the size and variations of the training set and to make the model more robust to unseen variations in the input data, i.e. to improve the generalisation to unseen domains.
Data augmentation techniques used in spacecraft pose estimation algorithms can be further categorised into:
\begin{itemize}
	\item Pixel-wise data augmentations such as blurring, noising or changing the image contrast
	\item Spatial-level  data augmentations such as rotation or scaling
\end{itemize}

The main difference between the two categories is their effect on the pose labels. The pixel-level augmentations only affect the input image, whereas the spatial-level augmentations require modifying both the input image as well as the pose label, which can be difficult. Figure \ref{fig:data_augmentation_techniques} illustrates different data augmentation techniques (pixel and spatial-level) applied to a reference image of PROBA-2 spacecraft from the SPARK2 \cite{rathinam_arunkumar_2022_6599762} dataset.
Finally, even though data augmentation generally helps with the domain gap problem,  there can be instances when applying data augmentation can be counter-productive. For example, the Random Erase augmentation used by Park et.al ~\cite{park2022robust} is shown to cause a drop in the pose estimation performance. 
Consequently, finding the best set of augmentations for a given context is a hard task in itself~\cite{peng2018jointly}. 
\Cref{table:table_training_info} provides a summary of data augmentation methods used in spacecraft pose estimation algorithms surveyed in this paper.

\begin{table*}[]
\caption{Datasets used and data augmentations applied with different pose estimation algorithms}
\label{table:table_training_info} 
\begin{tabular}{|p{2.5cm}|p{1.9cm}|p{11.5cm}|}
\hline
\rowcolor[HTML]{EFEFEF} \textbf{Algorithm} & \textbf{Datasets Used} & \textbf{Data Augmentations Applied}  \\ \hline


EPFL\_cvlab~\cite{gerard2019}  & SPEED & Rotation, addition of random noise, zooming and cropping  \\ \hline

SLAB Baseline~\cite{park2019} & SPEED, PRISMA12K, PRISMA25 &    
Random variations in brightness and contrast, random flipping, rotation at 90 degree intervals and addition of random Gaussian noise. Also, RoI enlargement and RoI shifting are applied specifically for object detector training. \\ \hline


STAR LAB keypoint method~\cite{rathinam2020} & SPEED, URSO-OrViS &   
Rotation, translation, coarse dropout, addition of Gaussian noise, random brightness and contrast variations applied for training keypoint prediction network \\ \hline

Black et al.~\cite{black2021real} & SPEED, Cygnus & Randomised flips, 90 degree rotations and crops applied for object detector training. Random translation and expansions, random flips, 90 degree rotations, brightness, contrast, and saturation augmentations applied for keypoint prediction training. \\ \hline

Wide-Depth-Range~\cite{hu2021} & SPEED, SwissCube & Random shift, scale and rotation   \\  \hline


LSPnet~\cite{garcia2021lspnet} &  SPEED  &   Centre data augmentation \\  \hline


URSONet~\cite{proenca2020}  & SPEED, URSO-OrViS  & Change in image exposure and contrast, addition of Additive White Gaussian (AWG) noise, blurring and drop out of patches, random camera orientation perturbations and random plane rotations (only for SPEED dataset)    \\ \hline

Mobile-URSONet~\cite{posso2022mobileurso} & SPEED & Random rotation of the camera across the roll axis with a maximum magnitude of 25 degrees, Gaussian blur, random changes to brightness, contrast, saturation, and hue \\ \hline

Huang et al.~\cite{huang2021non} & SPEED, URSO & Change in image exposure and contrast, addition of AWG noise, blurring and drop out patches, random camera orientation perturbations and random plane rotations (only for SPEED) \\ \hline 

Lotti et al.~\cite{spetpu22}& SPEED, CPD & Random image rotations, bounding box enlargement and shifts, random brightness, and contrast adjustments \\ \hline

Kecen et al.~\cite{kecenli2022} & SPEED & Same as SLAB Baseline \\ \hline

SPNv2~\cite{park2022robust} & SPEED+ & Style augmentation via neural style transfer, brightness and contrast, random erase, sun flare, blur (motion blur, median blur, glass blur), noise (Gaussian noise, ISO noise)  \\ \hline

Sharma et al.~\cite{sharma2018pose} & PRISMA (Imitation-25) & Horizontal reflection, addition of zero mean white Gaussian noise\\ \hline


CA-SpaceNet~\cite{CA-SpaceNet2022} & SPEED, SwissCube & Random shift, scale, and rotation 
\\ \hline

Legrand et al.~\cite{legrandend} & SPEED & Random variations in brightness and contrast, 
Gaussian noise augmentations, 
random rotations, 
and random background data augmentation 
\\ \hline

\end{tabular}
\end{table*}

\subsubsection{Domain Randomization}
\label{ssec:domain_randomization}

The goal is to help the model generalise by training it on a set of sufficiently randomized source data so that the target domain appears as just another randomization to the model \cite{Tobin2017DomainRF}. Hence, the expectation is that the model will be less prone to the domain gap~\cite{tobin2017domain}. An example of domain randomization in the context of spacecraft pose estimation is provided in~\cite{park2019}, where the spacecraft texture is randomized using the Neural Style Transfer (NST) technique presented in~\cite{DBLP:journals/corr/abs-1809-05375}. Domain randomization can be seen as a particular case of data augmentation: one does not search for a set of augmentations relevant to a context, but for a sufficiently varied set of augmentations that will make the actual scene appear as just another variation. 

\subsubsection{Multi-Task Learning}
\label{ssec:multi_task_learning} 
\begin{figure}[ht]
    \centering
    \includegraphics[width=\linewidth]{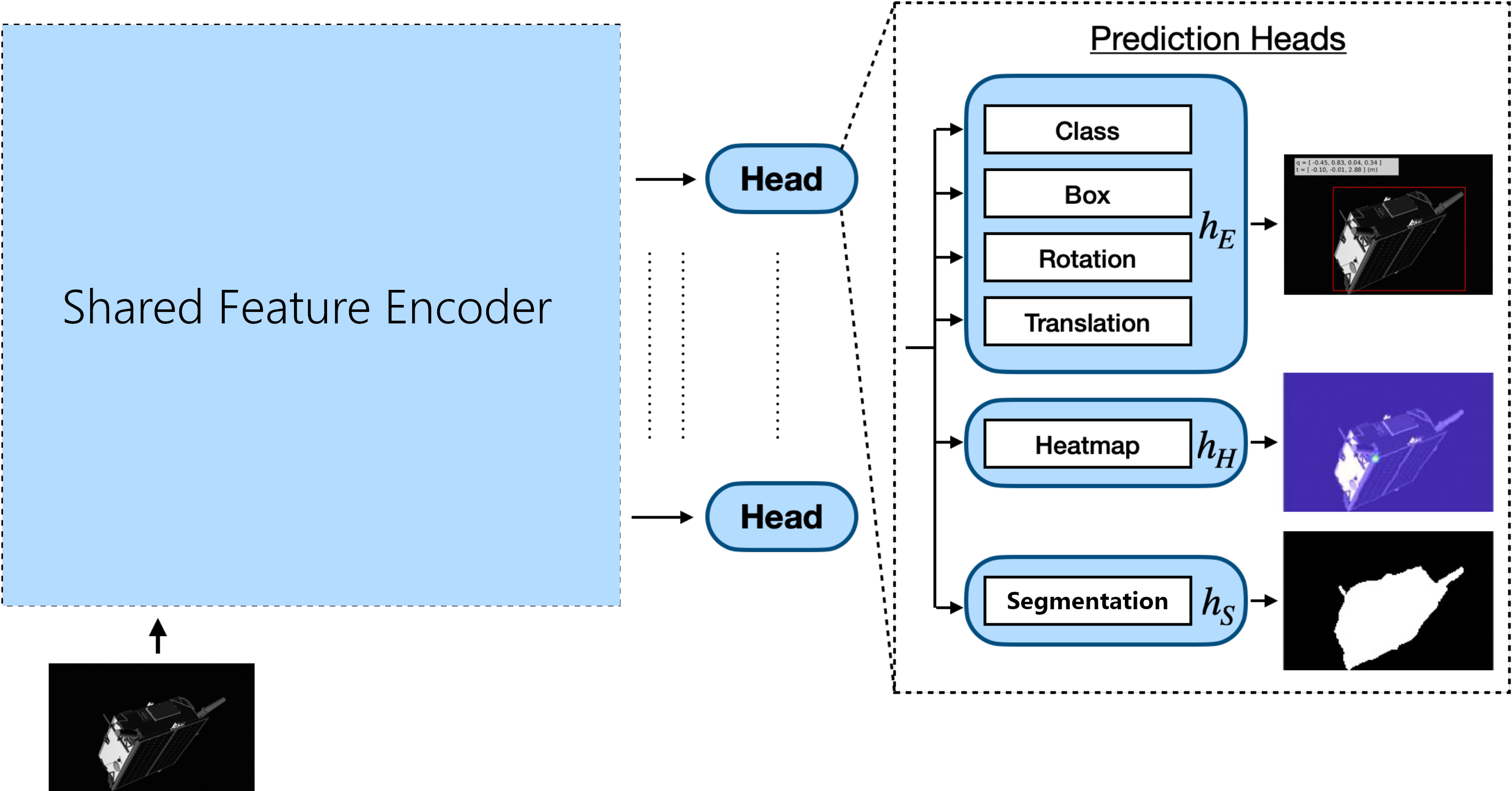}
    \caption{A model architecture used for multi-task learning, where some layers are shared between all tasks, and some layers are dedicated to specific tasks. Adopted from~\cite{park2022robust}.}
    \label{fig:mtl_shared}
\end{figure}

In this approach,  a single DL model is trained to perform multiple related tasks (a primary task and several secondary/ auxiliary tasks) simultaneously. The assumption here is that the model will generalise better on the primary task (spacecraft pose estimation in this context) by being less prone to the noise induced by the primary task~\cite{ruder2017overview}. The most common way of implementing multi-task learning is to have a shared backbone architecture extracting features and feeding these features to the task-specific layers \cite{park2022robust} (see \Cref{fig:mtl_shared}). Here EfficientPose~\cite{bukschat2020efficientpose} network architecture is modified with the addition of two heads: one for the segmentation of the spacecraft and one to compute the 2D heatmaps associated with pre-designated keypoints on the spacecraft. The results show that when the model is trained with different head configurations, the best performance is reached when all the task heads are enabled, thereby showing the effectiveness of multi-task learning. However, the authors show that all the heads do not contribute to the same extent; the segmentation head only improves the performance slightly. This highlights one of the key challenges in multi-task learning: identifying the correct set of secondary tasks that is relevant for a particular primary task \cite{shui2019principled}. 


\subsubsection{Domain-Adversarial Learning}
\label{ssec:adversarial}

This technique \cite{ganin2016domain} is applied to spacecraft pose estimation \cite{park2021} to bridge the domain gap. The goal here is to help the model learn features that are domain-invariant but discriminative with respect to the pose estimation task. A domain classifier, whose purpose is to discriminate between the source and the target domain, is attached to the model and its loss function maximised over the learning phase. The underlying idea of this method is that the less this classifier can distinguish between the source and the target domain, the more domain-invariant the model becomes.

\subsection{Limitations}
\label{sec:dataset_limitations}

Current datasets and evaluation procedures are still insufficient to enable the deployment of DL-based spacecraft pose estimation algorithms in space missions. We identify key limitations below.

\subsubsection{Realism of Synthetically Generated Datasets} 
One factor increasing the domain gap is the realism of the synthetic images used to train the models. The question of rendering realistic images is a hard topic in the context of space because it involves simulating the behaviour of light and its interaction with various materials and surfaces in space. The lack of reference points and the absence of an atmosphere make it difficult to create realistic lighting and shading effects. To achieve a realistic depiction of space, computer graphics techniques need to be tailored specifically to the unique properties of space environments. Therefore, the question of how to render more realistic synthetic space images is a challenging and open research topic.

\subsubsection{Algorithm Evaluation} While several attempts have been made to mitigate the domain gap between synthetic and laboratory images, there persists a one-order-of-magnitude difference between the best pose scores in the 2019 (synthetic test images) and 2021 (laboratory test images) editions of ESA's Satellite Pose Estimation Challenge~\cite{kisantal2020,park2021}. Moreover, ensuring that an algorithm trained on synthetic images (source domain) performs well on laboratory images (target domain) does not guarantee that the performance level will be maintained for space images, mainly as a result of the domain gap between the two environments.


\section{Future Research Directions}
\label{sec:future}
    
Despite the recent progress in spacecraft pose estimation, there is room for improvement in algorithm development and data generation (or collection). This section summarises open research questions and possible future directions for the field.

\subsection{Deployability of Algorithms}
The end goal of developing spacecraft pose estimation algorithms is to deploy them in space-borne hardware with limited resources. However, most existing algorithms are tested on workstations and large server clusters and very limited evaluations have been conducted on edge systems with FPGA~\cite{furano2020towards,leon2022towards} or GPU~\cite{kosmidis2020gpu4s,powell2018commercial,bruhn2020enabling}-based AI accelerators for space applications. In this context, this survey has made an effort to perform a trade-off comparison between the number of parameters (which can be a measure of resource consumption in the deployed hardware) in the DL models used and the algorithm performance. However, the lack of relevant information reported makes this difficult. In future works, it would be valuable for authors to report additional metrics such as size, number of FLOPS and latency, which are suitable measures for estimating the deployability of algorithms. 

Another future direction is to develop novel DL models specifically suited for edge AI accelerators. Unlike commonplace Nvidia GPUs, AI accelerators for space systems support only a limited number of network layers and operations~\cite{vitisaiuserguide}. DL models with unsupported layers will have difficulty to work on such devices. Techniques like Neural Architecture Search (NAS)~\cite{wistuba2019survey} can be used for developing efficient DL models which are deployable in space systems.   

\subsection{Explainability of Algorithms}
In real-world applications, the explainability of algorithms is a key factor in determining their reliability and trustworthiness. Especially in safety-critical applications like in space,  it is important to know why and how a decision / prediction was made. However, the black-box nature of DL models makes them weak for interpreting their inference processes and final results. This makes explainability difficult in DL-based spacecraft pose estimation, especially for direct end-to-end algorithms. Recently, eXplainable-AI (XAI) \cite{gunning2019xai} has become a hot research topic, with new methods developed \cite{bai2021explainable,xu2019explainable}. Several of these proposed methods, like Bayesian deep learning \cite{kendall2017uncertainties} or conformal inference methods ~\cite{shafer2008tutorial, angelopoulos2021gentle, NEURIPS2019_8fb21ee7} can be applied to spacecraft pose estimation improving their explainability, which are interesting directions for future research.

\subsection{Multi-Modal Spacecraft Pose Estimation} Most existing methods focus on visible-range images only. However, visible cameras are likely to suffer from difficult acquisition conditions in space (e.g., low light, overexposure). Therefore, other sensor such as thermal and time-of-flight of event cameras need to be considered in order to extend the operational range of classical computer vision methods. Till now, only a few works have investigated multi-modality for spacecraft pose estimation~\cite{seenic, rondao2021, DBLP:journals/corr/abs-2105-13789,rondao2022chinet} which is a direction of interest for the future of vision-based navigation in space.

\subsection{Generation of More Realistic Synthetic Data} As mentioned in section \ref{sec:data}, the main issue with the application of machine learning to space applications is the lack of data. Moreover, the ubiquitous resort to synthetic data is the source of the current domain gap problem faced in the literature. Addressing this problem could be done through a deep analysis of the rendering engines' images compared to actual space imagery. The results of this analysis could serve as the starting point for developing a rendering engine dedicated to generating realistic data for model training. To the best of our knowledge, PANGU~\cite{martin2019planetary} is the only initiative on this track to date. Another approach for simulation-to-real, is to introduce a physics informed layer into a deep learning system, as for example in~\cite{lengyel2021zero}. This may induce invariance to lighting conditions in images of satellites that result from complex lighting and shadowing conditions for satellites orbiting the Earth, such as from reflections from the satellite itself, from the Earth’s surface, and from the moon’s surface.  

\subsection{Domain Adaptation} One of the main obstacles to the deployment of DL-based pose estimation methods in space is the performance gap when the models are trained on synthetic images and tested on real ones. The second edition of the ESA Pose Estimation Challenge \cite{park2021} was specifically designed to address this challenge, with one synthetic training and two lab test datasets. Winning methods \cite{park2023satellite} have taken advantage of dedicated learning approaches, such as self-supervised, multi-task or adversarial learning. Together with the generation of more realistic synthetic datasets for training, domain adaptation is likely to receive much interest in the coming years to overcome the domain gap problem.

\subsection{Beyond Target-Specific Spacecraft Pose Estimation} Current algorithms estimate the pose of a single type of spacecraft at a time. For every additional spacecraft, a new dataset has to be generated and the algorithm needs to be retrained. However, with the increasing number of spacecraft launched yearly, a natural way forward is to develop more generic algorithms that are not restricted to a particular spacecraft model. Especially in applications such as debris removal, the original spacecraft structure can disintegrate into geometrical shapes not seen by the algorithm during training. Generic 6D pose estimation methods for unseen objects~\cite{gou2022unseen}~\cite{park2020latentfusion} can be exploited to develop spacecraft-agnostic pose estimation algorithms. 

\subsection{Multi-Frame Spacecraft Pose Estimation}
Multi-frame spacecraft pose estimation refers to determining the spacecraft pose using consecutive images, thereby leveraging temporal information. Current spacecraft pose estimation algorithms consider each image frame in isolation and the pose is estimated from information extracted from this single image frame. However, in space, pose estimation algorithms are commonly used in applications such as autonomous navigation, where a sequence of consecutive images (trajectories) is available. Hence using temporal information is key to higher pose estimation accuracy and generating temporally consistent poses. Datasets like SPARK2 \cite{rathinam_arunkumar_2022_6599762} already provide pose estimation data as trajectories. In this direction, recently proposed ChiNet \cite{rondao2022chinet} have used Long Short-Term Memory (LSTM) \cite{10.1162/neco.1997.9.8.1735} units in modelling sequences of data for estimating the spacecraft pose. However, there is a rich history of video-based 6 DoF pose estimation methods leveraging temporal information in general computer vision \cite{beedu2021videopose,beedu2022video, clark2017vidloc}. In future, these methods can be in-cooperated into spacecraft pose estimation, especially for applications such as spacecraft related navigation.  
\section{Conclusions}
\label{sec:conclusion}


Monocular vision-based spacecraft pose estimation has seen considerable progress with the use of DL in recent years. However, there are still fundamental concerns that need to be addressed before these algorithms are deployed in actual space scenarios. This survey highlights these limitations, both in terms of algorithms design and datasets used for training and validation/testing. With this aim, the survey first summarised
the existing algorithms according to two common approaches: hybrid modular and direct end-to-end regression approaches. Algorithms were compared in terms of performance as well as the size of the network architectures used to help understand their deployability. Then the spacecraft pose estimation datasets available for training and validating/testing these methods were discussed. The data generation methods, simulators and testbeds, and strategies used to bridge the domain gap problem  were also discussed in detail. Finally, the survey provides future research directions to address these limitations and to develop spacecraft pose estimation algorithms deployable in real space missions.

\appendix
\section{Additional Information}
\subsection{Publicly available algorithm implementations}
\label{ssec:algo_links}
\begin{itemize}
    \item \url{https://github.com/BoChenYS/satellite-pose-estimation}~\cite{chen2019}
    \item \url{https://indico.esa.int/event/319/attachments/3561/4754/pose\_gerard\_segmentation.pdf}~\cite{gerard2019}
    \item \url{https://github.com/cvlab-epfl/wide-depth-range-pose}~\cite{hu2021}
    \item \url{https://github.com/tpark94/speedplusbaseline}~\cite{park2019}
    \item \url{https://github.com/pedropro/UrsoNet}~\cite{proenca2020}
    \item \url{https://github.com/possoj/Mobile-URSONet}~\cite{posso2022mobileurso}
    \item \url{https://github.com/tpark94/spnv2}~\cite{park2022robust}
    \item \url{https://github.com/Shunli-Wang/CA-SpaceNet}~\cite{CA-SpaceNet2022}
\end{itemize}

\subsection{Links to the publicly available datasets}
\label{ssec:datasets_links}
\begin{itemize}
    \item SHIRT: \url{https://taehajeffpark.com/shirt/}
	\item SPARK2022: \url{https://cvi2.uni.lu/spark2022/registration/}
	\item SwissCube: \url{https://github.com/cvlab-epfl/wide-depth-range-pose}
	\item SPEED+: \url{https://zenodo.org/record/5588480}
	\item SPEED: \url{https://zenodo.org/record/6327547}
	\item URSO-OrViS: \url{https://zenodo.org/record/3279632}
\end{itemize}


\clearpage
\bibliographystyle{cas-model2-names}
\bibliography{main}

\begin{thebibliography}{186}
\expandafter\ifx\csname natexlab\endcsname\relax\def\natexlab#1{#1}\fi
\providecommand{\url}[1]{\texttt{#1}}
\providecommand{\href}[2]{#2}
\providecommand{\path}[1]{#1}
\providecommand{\DOIprefix}{doi:}
\providecommand{\ArXivprefix}{arXiv:}
\providecommand{\URLprefix}{URL: }
\providecommand{\Pubmedprefix}{pmid:}
\providecommand{\doi}[1]{\href{http://dx.doi.org/#1}{\path{#1}}}
\providecommand{\Pubmed}[1]{\href{pmid:#1}{\path{#1}}}
\providecommand{\bibinfo}[2]{#2}
\ifx\xfnm\relax \def\xfnm[#1]{\unskip,\space#1}\fi
\bibitem[{Angelopoulos and Bates(2021)}]{angelopoulos2021gentle}
\bibinfo{author}{Angelopoulos, A.N.}, \bibinfo{author}{Bates, S.},
  \bibinfo{year}{2021}.
\newblock \bibinfo{title}{A gentle introduction to conformal prediction and
  distribution-free uncertainty quantification}.
\newblock \bibinfo{journal}{arXiv preprint arXiv:2107.07511} .
\bibitem[{Azodi et~al.(2020)Azodi, Tang and Shiu}]{azodi2020opening}
\bibinfo{author}{Azodi, C.B.}, \bibinfo{author}{Tang, J.},
  \bibinfo{author}{Shiu, S.H.}, \bibinfo{year}{2020}.
\newblock \bibinfo{title}{Opening the black box: interpretable machine learning
  for geneticists}.
\newblock \bibinfo{journal}{Trends in genetics} \bibinfo{volume}{36},
  \bibinfo{pages}{442--455}.
\bibitem[{Bai et~al.(2021)Bai, Wang, Liu, Liu, Song, Sebe and
  Kim}]{bai2021explainable}
\bibinfo{author}{Bai, X.}, \bibinfo{author}{Wang, X.}, \bibinfo{author}{Liu,
  X.}, \bibinfo{author}{Liu, Q.}, \bibinfo{author}{Song, J.},
  \bibinfo{author}{Sebe, N.}, \bibinfo{author}{Kim, B.}, \bibinfo{year}{2021}.
\newblock \bibinfo{title}{Explainable deep learning for efficient and robust
  pattern recognition: A survey of recent developments}.
\newblock \bibinfo{journal}{Pattern Recognition} \bibinfo{volume}{120},
  \bibinfo{pages}{108102}.
\bibitem[{Baller et~al.(2021)Baller, Jindal, Chadha and
  Gerndt}]{baller2021deepedgebench}
\bibinfo{author}{Baller, S.P.}, \bibinfo{author}{Jindal, A.},
  \bibinfo{author}{Chadha, M.}, \bibinfo{author}{Gerndt, M.},
  \bibinfo{year}{2021}.
\newblock \bibinfo{title}{Deepedgebench: Benchmarking deep neural networks on
  edge devices}, in: \bibinfo{booktitle}{2021 IEEE International Conference on
  Cloud Engineering (IC2E)}, \bibinfo{organization}{IEEE}. pp.
  \bibinfo{pages}{20--30}.
\bibitem[{Bechini et~al.(2022)Bechini, Lunghi, Lavagna
  et~al.}]{bechini2022spacecraft}
\bibinfo{author}{Bechini, M.}, \bibinfo{author}{Lunghi, P.},
  \bibinfo{author}{Lavagna, M.}, et~al., \bibinfo{year}{2022}.
\newblock \bibinfo{title}{Spacecraft pose estimation via monocular image
  processing: Dataset generation and validation}, in: \bibinfo{booktitle}{9th
  European Conference for Aerospace Sciences (EUCASS 2022)}, pp.
  \bibinfo{pages}{1--15}.
\bibitem[{Beedu et~al.(2022)Beedu, Alamri and Essa}]{beedu2022video}
\bibinfo{author}{Beedu, A.}, \bibinfo{author}{Alamri, H.},
  \bibinfo{author}{Essa, I.}, \bibinfo{year}{2022}.
\newblock \bibinfo{title}{Video based object 6d pose estimation using
  transformers}.
\newblock \bibinfo{journal}{arXiv preprint arXiv:2210.13540} .
\bibitem[{Beedu et~al.(2021)Beedu, Ren, Agrawal and Essa}]{beedu2021videopose}
\bibinfo{author}{Beedu, A.}, \bibinfo{author}{Ren, Z.},
  \bibinfo{author}{Agrawal, V.}, \bibinfo{author}{Essa, I.},
  \bibinfo{year}{2021}.
\newblock \bibinfo{title}{Videopose: Estimating 6d object pose from videos}.
\newblock \bibinfo{journal}{arXiv preprint arXiv:2111.10677} .
\bibitem[{Beierle and D’Amico(2019)}]{beierle2019variable}
\bibinfo{author}{Beierle, C.}, \bibinfo{author}{D’Amico, S.},
  \bibinfo{year}{2019}.
\newblock \bibinfo{title}{Variable-magnification optical stimulator for
  training and validation of spaceborne vision-based navigation}.
\newblock \bibinfo{journal}{Journal of Spacecraft and Rockets}
  \bibinfo{volume}{56}, \bibinfo{pages}{1060--1072}.
\bibitem[{Ben-David et~al.(2006)Ben-David, Blitzer, Crammer and
  Pereira}]{ben2006analysis}
\bibinfo{author}{Ben-David, S.}, \bibinfo{author}{Blitzer, J.},
  \bibinfo{author}{Crammer, K.}, \bibinfo{author}{Pereira, F.},
  \bibinfo{year}{2006}.
\newblock \bibinfo{title}{Analysis of representations for domain adaptation}.
\newblock \bibinfo{journal}{Advances in neural information processing systems}
  \bibinfo{volume}{19}.
\bibitem[{Benninghoff et~al.(2017)Benninghoff, Rems, Risse and
  Mietner}]{Benninghoff2017EuropeanPO}
\bibinfo{author}{Benninghoff, H.}, \bibinfo{author}{Rems, F.},
  \bibinfo{author}{Risse, E.A.}, \bibinfo{author}{Mietner, C.},
  \bibinfo{year}{2017}.
\newblock \bibinfo{title}{European proximity operations simulator 2.0 (epos) -
  a robotic-based rendezvous and docking simulator}.
\newblock \bibinfo{journal}{Journal of large-scale research facilities JLSRF}
  \bibinfo{volume}{3}, \bibinfo{pages}{107}.
\bibitem[{Biesbroek et~al.(2021)Biesbroek, Aziz, Wolahan, Cipolla, Richard-Noca
  and Piguet}]{biesbroek2021clearspace}
\bibinfo{author}{Biesbroek, R.}, \bibinfo{author}{Aziz, S.},
  \bibinfo{author}{Wolahan, A.}, \bibinfo{author}{Cipolla, S.f.},
  \bibinfo{author}{Richard-Noca, M.}, \bibinfo{author}{Piguet, L.},
  \bibinfo{year}{2021}.
\newblock \bibinfo{title}{The clearspace-1 mission: Esa and clearspace team up
  to remove debris}, in: \bibinfo{booktitle}{Proc. 8th Eur. Conf. Sp. Debris},
  pp. \bibinfo{pages}{1--3}.
\bibitem[{Black et~al.(2021)Black, Shankar, Fonseka, Deutsch, Dhir and
  Akella}]{black2021real}
\bibinfo{author}{Black, K.}, \bibinfo{author}{Shankar, S.},
  \bibinfo{author}{Fonseka, D.}, \bibinfo{author}{Deutsch, J.},
  \bibinfo{author}{Dhir, A.}, \bibinfo{author}{Akella, M.R.},
  \bibinfo{year}{2021}.
\newblock \bibinfo{title}{Real-time, flight-ready, non-cooperative spacecraft
  pose estimation using monocular imagery}.
\newblock \bibinfo{journal}{arXiv preprint arXiv:2101.09553} .
\bibitem[{Bochkovskiy et~al.(2020)Bochkovskiy, Wang and
  Liao}]{bochkovskiy2020yolov4}
\bibinfo{author}{Bochkovskiy, A.}, \bibinfo{author}{Wang, C.Y.},
  \bibinfo{author}{Liao, H.Y.M.}, \bibinfo{year}{2020}.
\newblock \bibinfo{title}{Yolov4: Optimal speed and accuracy of object
  detection}.
\newblock \bibinfo{journal}{arXiv preprint arXiv:2004.10934} .
\bibitem[{Brochard et~al.(2018)Brochard, Lebreton, Robin, Kanani, Jonniaux,
  Masson, Despr{\'e} and Berjaoui}]{brochard2018scientific}
\bibinfo{author}{Brochard, R.}, \bibinfo{author}{Lebreton, J.},
  \bibinfo{author}{Robin, C.}, \bibinfo{author}{Kanani, K.},
  \bibinfo{author}{Jonniaux, G.}, \bibinfo{author}{Masson, A.},
  \bibinfo{author}{Despr{\'e}, N.}, \bibinfo{author}{Berjaoui, A.},
  \bibinfo{year}{2018}.
\newblock \bibinfo{title}{Scientific image rendering for space scenes with the
  surrender software}.
\newblock \bibinfo{journal}{arXiv preprint arXiv:1810.01423} .
\bibitem[{Bruhn et~al.(2020)Bruhn, Tsog, Kunkel, Flordal and
  Troxel}]{bruhn2020enabling}
\bibinfo{author}{Bruhn, F.C.}, \bibinfo{author}{Tsog, N.},
  \bibinfo{author}{Kunkel, F.}, \bibinfo{author}{Flordal, O.},
  \bibinfo{author}{Troxel, I.}, \bibinfo{year}{2020}.
\newblock \bibinfo{title}{Enabling radiation tolerant heterogeneous gpu-based
  onboard data processing in space}.
\newblock \bibinfo{journal}{CEAS Space Journal} \bibinfo{volume}{12},
  \bibinfo{pages}{551--564}.
\bibitem[{Bukschat and Vetter(2020)}]{bukschat2020efficientpose}
\bibinfo{author}{Bukschat, Y.}, \bibinfo{author}{Vetter, M.},
  \bibinfo{year}{2020}.
\newblock \bibinfo{title}{Efficientpose: An efficient, accurate and scalable
  end-to-end 6d multi object pose estimation approach}.
\newblock \bibinfo{journal}{arXiv preprint arXiv:2011.04307} .
\bibitem[{Cai and Vasconcelos(2018)}]{cai2018cascade}
\bibinfo{author}{Cai, Z.}, \bibinfo{author}{Vasconcelos, N.},
  \bibinfo{year}{2018}.
\newblock \bibinfo{title}{Cascade r-cnn: Delving into high quality object
  detection}, in: \bibinfo{booktitle}{Proceedings of the IEEE conference on
  computer vision and pattern recognition}, pp. \bibinfo{pages}{6154--6162}.
\bibitem[{Cao et~al.(2020)Cao, Zhou, Wu, Ming, Xu and Zhang}]{cao2020research}
\bibinfo{author}{Cao, W.}, \bibinfo{author}{Zhou, C.}, \bibinfo{author}{Wu,
  Y.}, \bibinfo{author}{Ming, Z.}, \bibinfo{author}{Xu, Z.},
  \bibinfo{author}{Zhang, J.}, \bibinfo{year}{2020}.
\newblock \bibinfo{title}{Research progress of zero-shot learning beyond
  computer vision}, in: \bibinfo{booktitle}{Algorithms and Architectures for
  Parallel Processing: 20th International Conference, ICA3PP 2020, New York
  City, NY, USA, October 2--4, 2020, Proceedings, Part II 20},
  \bibinfo{organization}{Springer}. pp. \bibinfo{pages}{538--551}.
\bibitem[{Capuano et~al.(2019)Capuano, Alimo, Ho and Chung}]{capuano2019robust}
\bibinfo{author}{Capuano, V.}, \bibinfo{author}{Alimo, S.R.},
  \bibinfo{author}{Ho, A.Q.}, \bibinfo{author}{Chung, S.J.},
  \bibinfo{year}{2019}.
\newblock \bibinfo{title}{Robust features extraction for on-board
  monocular-based spacecraft pose acquisition}, in: \bibinfo{booktitle}{AIAA
  Scitech 2019 Forum}, p. \bibinfo{pages}{2005}.
\bibitem[{Cassinis et~al.(2019)Cassinis, Fonod and Gill}]{cassinis2019review}
\bibinfo{author}{Cassinis, L.P.}, \bibinfo{author}{Fonod, R.},
  \bibinfo{author}{Gill, E.}, \bibinfo{year}{2019}.
\newblock \bibinfo{title}{Review of the robustness and applicability of
  monocular pose estimation systems for relative navigation with an
  uncooperative spacecraft}.
\newblock \bibinfo{journal}{Progress in Aerospace Sciences}
  \bibinfo{volume}{110}, \bibinfo{pages}{100548}.
\bibitem[{Cassinis et~al.(2021)Cassinis, Menicucci, Gill, Ahrns and
  Fernandez}]{cassinis2021ground}
\bibinfo{author}{Cassinis, L.P.}, \bibinfo{author}{Menicucci, A.},
  \bibinfo{author}{Gill, E.}, \bibinfo{author}{Ahrns, I.},
  \bibinfo{author}{Fernandez, J.G.}, \bibinfo{year}{2021}.
\newblock \bibinfo{title}{On-ground validation of a cnn-based monocular pose
  estimation system for uncooperative spacecraft}, in: \bibinfo{booktitle}{8th
  European Conference on Space Debris}.
\bibitem[{Chai et~al.(2021)Chai, Zeng, Li and Ngai}]{chai2021deep}
\bibinfo{author}{Chai, J.}, \bibinfo{author}{Zeng, H.}, \bibinfo{author}{Li,
  A.}, \bibinfo{author}{Ngai, E.W.}, \bibinfo{year}{2021}.
\newblock \bibinfo{title}{Deep learning in computer vision: A critical review
  of emerging techniques and application scenarios}.
\newblock \bibinfo{journal}{Machine Learning with Applications}
  \bibinfo{volume}{6}, \bibinfo{pages}{100134}.
\bibitem[{Chen et~al.(2019a)Chen, Cao, Parra and Chin}]{chen2019}
\bibinfo{author}{Chen, B.}, \bibinfo{author}{Cao, J.}, \bibinfo{author}{Parra,
  A.}, \bibinfo{author}{Chin, T.J.}, \bibinfo{year}{2019}a.
\newblock \bibinfo{title}{Satellite pose estimation with deep landmark
  regression and nonlinear pose refinement}, in:
  \bibinfo{booktitle}{Proceedings of the IEEE/CVF International Conference on
  Computer Vision (ICCV) Workshops}.
\bibitem[{Chen et~al.(2019b)Chen, Cao, Parra and Chin}]{old_chen2019}
\bibinfo{author}{Chen, B.}, \bibinfo{author}{Cao, J.}, \bibinfo{author}{Parra,
  A.}, \bibinfo{author}{Chin, T.J.}, \bibinfo{year}{2019}b.
\newblock \bibinfo{title}{Satellite pose estimation with deep landmark
  regression and nonlinear pose refinement}, in:
  \bibinfo{booktitle}{Proceedings of the IEEE/CVF International Conference on
  Computer Vision Workshops}, pp. \bibinfo{pages}{0--0}.
\bibitem[{Cheng et~al.(2020)Cheng, Xiao, Wang, Shi, Huang and
  Zhang}]{cheng2020}
\bibinfo{author}{Cheng, B.}, \bibinfo{author}{Xiao, B.}, \bibinfo{author}{Wang,
  J.}, \bibinfo{author}{Shi, H.}, \bibinfo{author}{Huang, T.S.},
  \bibinfo{author}{Zhang, L.}, \bibinfo{year}{2020}.
\newblock \bibinfo{title}{Higherhrnet: Scale-aware representation learning for
  bottom-up human pose estimation}, in: \bibinfo{booktitle}{Proceedings of the
  IEEE/CVF Conference on Computer Vision and Pattern Recognition}, pp.
  \bibinfo{pages}{5386--5395}.
\bibitem[{Ciaparrone et~al.(2020)Ciaparrone, S{\'a}nchez, Tabik, Troiano,
  Tagliaferri and Herrera}]{ciaparrone2020deep}
\bibinfo{author}{Ciaparrone, G.}, \bibinfo{author}{S{\'a}nchez, F.L.},
  \bibinfo{author}{Tabik, S.}, \bibinfo{author}{Troiano, L.},
  \bibinfo{author}{Tagliaferri, R.}, \bibinfo{author}{Herrera, F.},
  \bibinfo{year}{2020}.
\newblock \bibinfo{title}{Deep learning in video multi-object tracking: A
  survey}.
\newblock \bibinfo{journal}{Neurocomputing} \bibinfo{volume}{381},
  \bibinfo{pages}{61--88}.
\bibitem[{Clark et~al.(2017)Clark, Wang, Markham, Trigoni and
  Wen}]{clark2017vidloc}
\bibinfo{author}{Clark, R.}, \bibinfo{author}{Wang, S.},
  \bibinfo{author}{Markham, A.}, \bibinfo{author}{Trigoni, N.},
  \bibinfo{author}{Wen, H.}, \bibinfo{year}{2017}.
\newblock \bibinfo{title}{Vidloc: A deep spatio-temporal model for 6-dof
  video-clip relocalization}, in: \bibinfo{booktitle}{Proceedings of the IEEE
  Conference on Computer Vision and Pattern Recognition}, pp.
  \bibinfo{pages}{6856--6864}.
\bibitem[{Colmenarejo et~al.(2019)Colmenarejo, Graziano, Novelli, Mora, Serra,
  Tomassini, Seweryn, Prisco and Fernandez}]{COLMENAREJO2019206}
\bibinfo{author}{Colmenarejo, P.}, \bibinfo{author}{Graziano, M.},
  \bibinfo{author}{Novelli, G.}, \bibinfo{author}{Mora, D.},
  \bibinfo{author}{Serra, P.}, \bibinfo{author}{Tomassini, A.},
  \bibinfo{author}{Seweryn, K.}, \bibinfo{author}{Prisco, G.},
  \bibinfo{author}{Fernandez, J.G.}, \bibinfo{year}{2019}.
\newblock \bibinfo{title}{On ground validation of debris removal technologies}.
\newblock \bibinfo{journal}{Acta Astronautica} \bibinfo{volume}{158},
  \bibinfo{pages}{206--219}.
\newblock \URLprefix
  \url{https://www.sciencedirect.com/science/article/pii/S0094576517312845},
  \DOIprefix\doi{https://doi.org/10.1016/j.actaastro.2018.01.026}.
\bibitem[{Cosmas and Kenichi(2020)}]{cosmas2020}
\bibinfo{author}{Cosmas, K.}, \bibinfo{author}{Kenichi, A.},
  \bibinfo{year}{2020}.
\newblock \bibinfo{title}{Utilization of fpga for onboard inference of landmark
  localization in cnn-based spacecraft pose estimation}.
\newblock \bibinfo{journal}{Aerospace} \bibinfo{volume}{7},
  \bibinfo{pages}{159}.
\bibitem[{D' et~al.(2014)D', Amico, Benn and Jørgensen}]{dpose2014}
\bibinfo{author}{D', S.}, \bibinfo{author}{Amico, N.}, \bibinfo{author}{Benn,
  M.}, \bibinfo{author}{Jørgensen, J.L.}, \bibinfo{year}{2014}.
\newblock \bibinfo{title}{Pose estimation of an uncooperative spacecraft from
  actual space imagery}.
\newblock \bibinfo{journal}{International Journal of Space Science and
  Engineering} \bibinfo{volume}{2}, \bibinfo{pages}{171}.
\newblock \URLprefix \url{http://www.inderscience.com/link.php?id=60600},
  \DOIprefix\doi{10.1504/IJSPACESE.2014.060600}.
\bibitem[{Deng et~al.(2009)Deng, Dong, Socher, Li, Li and
  Fei-Fei}]{deng2009imagenet}
\bibinfo{author}{Deng, J.}, \bibinfo{author}{Dong, W.},
  \bibinfo{author}{Socher, R.}, \bibinfo{author}{Li, L.J.},
  \bibinfo{author}{Li, K.}, \bibinfo{author}{Fei-Fei, L.},
  \bibinfo{year}{2009}.
\newblock \bibinfo{title}{Imagenet: A large-scale hierarchical image database},
  in: \bibinfo{booktitle}{2009 IEEE conference on computer vision and pattern
  recognition}, \bibinfo{organization}{Ieee}. pp. \bibinfo{pages}{248--255}.
\bibitem[{D’Amico et~al.(2014)D’Amico, Benn and J{\o}rgensen}]{d2014pose}
\bibinfo{author}{D’Amico, S.}, \bibinfo{author}{Benn, M.},
  \bibinfo{author}{J{\o}rgensen, J.L.}, \bibinfo{year}{2014}.
\newblock \bibinfo{title}{Pose estimation of an uncooperative spacecraft from
  actual space imagery}.
\newblock \bibinfo{journal}{International Journal of Space Science and
  Engineering 5} \bibinfo{volume}{2}, \bibinfo{pages}{171--189}.
\bibitem[{(ESA)(2010)}]{tango}
\bibinfo{author}{(ESA), E.S.A.}, \bibinfo{year}{2010}.
\newblock \bibinfo{title}{Prisma’s tango and mango satellites}.
\newblock
  \bibinfo{howpublished}{\url{https://www.esa.int/ESA_Multimedia/Images/2010/10/Prisma_s_Tango_and_Mango_satellites}}.
\newblock \bibinfo{note}{Accessed: 05-April-2023}.
\bibitem[{Fang et~al.(2022)Fang, Yap, Lin, Zhu and Liu}]{fang2022source}
\bibinfo{author}{Fang, Y.}, \bibinfo{author}{Yap, P.T.}, \bibinfo{author}{Lin,
  W.}, \bibinfo{author}{Zhu, H.}, \bibinfo{author}{Liu, M.},
  \bibinfo{year}{2022}.
\newblock \bibinfo{title}{Source-free unsupervised domain adaptation: A
  survey}.
\newblock \bibinfo{journal}{arXiv preprint arXiv:2301.00265} .
\bibitem[{Fischler and Bolles(1981)}]{fischler1981random}
\bibinfo{author}{Fischler, M.A.}, \bibinfo{author}{Bolles, R.C.},
  \bibinfo{year}{1981}.
\newblock \bibinfo{title}{Random sample consensus: a paradigm for model fitting
  with applications to image analysis and automated cartography}.
\newblock \bibinfo{journal}{Communications of the ACM} \bibinfo{volume}{24},
  \bibinfo{pages}{381--395}.
\bibitem[{Furano et~al.(2020)Furano, Meoni, Dunne, Moloney, Ferlet-Cavrois,
  Tavoularis, Byrne, Buckley, Psarakis, Voss et~al.}]{furano2020towards}
\bibinfo{author}{Furano, G.}, \bibinfo{author}{Meoni, G.},
  \bibinfo{author}{Dunne, A.}, \bibinfo{author}{Moloney, D.},
  \bibinfo{author}{Ferlet-Cavrois, V.}, \bibinfo{author}{Tavoularis, A.},
  \bibinfo{author}{Byrne, J.}, \bibinfo{author}{Buckley, L.},
  \bibinfo{author}{Psarakis, M.}, \bibinfo{author}{Voss, K.O.}, et~al.,
  \bibinfo{year}{2020}.
\newblock \bibinfo{title}{Towards the use of artificial intelligence on the
  edge in space systems: Challenges and opportunities}.
\newblock \bibinfo{journal}{IEEE Aerospace and Electronic Systems Magazine}
  \bibinfo{volume}{35}, \bibinfo{pages}{44--56}.
\bibitem[{Ganin et~al.(2016)Ganin, Ustinova, Ajakan, Germain, Larochelle,
  Laviolette, Marchand and Lempitsky}]{ganin2016domain}
\bibinfo{author}{Ganin, Y.}, \bibinfo{author}{Ustinova, E.},
  \bibinfo{author}{Ajakan, H.}, \bibinfo{author}{Germain, P.},
  \bibinfo{author}{Larochelle, H.}, \bibinfo{author}{Laviolette, F.},
  \bibinfo{author}{Marchand, M.}, \bibinfo{author}{Lempitsky, V.},
  \bibinfo{year}{2016}.
\newblock \bibinfo{title}{Domain-adversarial training of neural networks}.
\newblock \bibinfo{journal}{The journal of machine learning research}
  \bibinfo{volume}{17}, \bibinfo{pages}{2096--2030}.
\bibitem[{Garcia et~al.(2021)Garcia, Musallam, Gaudilliere, Ghorbel,
  Al~Ismaeil, Perez and Aouada}]{garcia2021lspnet}
\bibinfo{author}{Garcia, A.}, \bibinfo{author}{Musallam, M.A.},
  \bibinfo{author}{Gaudilliere, V.}, \bibinfo{author}{Ghorbel, E.},
  \bibinfo{author}{Al~Ismaeil, K.}, \bibinfo{author}{Perez, M.},
  \bibinfo{author}{Aouada, D.}, \bibinfo{year}{2021}.
\newblock \bibinfo{title}{Lspnet: A 2d localization-oriented spacecraft pose
  estimation neural network}, in: \bibinfo{booktitle}{Proceedings of the
  IEEE/CVF Conference on Computer Vision and Pattern Recognition}, pp.
  \bibinfo{pages}{2048--2056}.
\bibitem[{Gaudilli{\`e}re et~al.(2023)Gaudilli{\`e}re, Pauly, Rathinam,
  Sanchez, Musallam and Aouada}]{gaudilliere20233d}
\bibinfo{author}{Gaudilli{\`e}re, V.}, \bibinfo{author}{Pauly, L.},
  \bibinfo{author}{Rathinam, A.}, \bibinfo{author}{Sanchez, A.G.},
  \bibinfo{author}{Musallam, M.A.}, \bibinfo{author}{Aouada, D.},
  \bibinfo{year}{2023}.
\newblock \bibinfo{title}{3d-aware object localization using gaussian implicit
  occupancy function}.
\newblock \bibinfo{journal}{arXiv preprint arXiv:2303.02058} .
\bibitem[{Ge et~al.(2021)Ge, Liu, Wang, Li and Sun}]{ge2021yolox}
\bibinfo{author}{Ge, Z.}, \bibinfo{author}{Liu, S.}, \bibinfo{author}{Wang,
  F.}, \bibinfo{author}{Li, Z.}, \bibinfo{author}{Sun, J.},
  \bibinfo{year}{2021}.
\newblock \bibinfo{title}{Yolox: Exceeding yolo series in 2021}.
\newblock \bibinfo{journal}{arXiv preprint arXiv:2107.08430} .
\bibitem[{Gerard(2019)}]{gerard2019}
\bibinfo{author}{Gerard, K.}, \bibinfo{year}{2019}.
\newblock \bibinfo{title}{Segmentation-Driven Satellite Pose Estimation}.
\newblock \bibinfo{type}{Technical Report}. Technical Report. EPFL. Available
  online at: https://indico. esa. int/event~….
\bibitem[{GMV(2018)}]{GMVplatformart}
\bibinfo{author}{GMV}, \bibinfo{year}{2018}.
\newblock \bibinfo{title}{platform-art}.
\newblock
  \bibinfo{howpublished}{\url{https://satsearch.co/services/gmv-platform-art-for-satellite-orbit-simulation}}.
\bibitem[{Gou et~al.(2022)Gou, Pan, Fang, Liu, Lu and Tan}]{gou2022unseen}
\bibinfo{author}{Gou, M.}, \bibinfo{author}{Pan, H.}, \bibinfo{author}{Fang,
  H.S.}, \bibinfo{author}{Liu, Z.}, \bibinfo{author}{Lu, C.},
  \bibinfo{author}{Tan, P.}, \bibinfo{year}{2022}.
\newblock \bibinfo{title}{Unseen object 6d pose estimation: a benchmark and
  baselines}.
\newblock \bibinfo{journal}{arXiv preprint arXiv:2206.11808} .
\bibitem[{Gunning et~al.(2019)Gunning, Stefik, Choi, Miller, Stumpf and
  Yang}]{gunning2019xai}
\bibinfo{author}{Gunning, D.}, \bibinfo{author}{Stefik, M.},
  \bibinfo{author}{Choi, J.}, \bibinfo{author}{Miller, T.},
  \bibinfo{author}{Stumpf, S.}, \bibinfo{author}{Yang, G.Z.},
  \bibinfo{year}{2019}.
\newblock \bibinfo{title}{Xai—explainable artificial intelligence}.
\newblock \bibinfo{journal}{Science robotics} \bibinfo{volume}{4},
  \bibinfo{pages}{eaay7120}.
\bibitem[{Hadidi et~al.(2019)Hadidi, Cao, Xie, Asgari, Krishna and
  Kim}]{hadidi2019characterizing}
\bibinfo{author}{Hadidi, R.}, \bibinfo{author}{Cao, J.}, \bibinfo{author}{Xie,
  Y.}, \bibinfo{author}{Asgari, B.}, \bibinfo{author}{Krishna, T.},
  \bibinfo{author}{Kim, H.}, \bibinfo{year}{2019}.
\newblock \bibinfo{title}{Characterizing the deployment of deep neural networks
  on commercial edge devices}, in: \bibinfo{booktitle}{2019 IEEE International
  Symposium on Workload Characterization (IISWC)},
  \bibinfo{organization}{IEEE}. pp. \bibinfo{pages}{35--48}.
\bibitem[{Hartley and Zisserman(2003)}]{hartley2003multiple}
\bibinfo{author}{Hartley, R.}, \bibinfo{author}{Zisserman, A.},
  \bibinfo{year}{2003}.
\newblock \bibinfo{title}{Multiple view geometry in computer vision}.
\newblock \bibinfo{publisher}{Cambridge university press}.
\bibitem[{He et~al.(2017)He, Gkioxari, Doll{\'a}r and Girshick}]{he2017mask}
\bibinfo{author}{He, K.}, \bibinfo{author}{Gkioxari, G.},
  \bibinfo{author}{Doll{\'a}r, P.}, \bibinfo{author}{Girshick, R.},
  \bibinfo{year}{2017}.
\newblock \bibinfo{title}{Mask r-cnn}, in: \bibinfo{booktitle}{Proceedings of
  the IEEE international conference on computer vision}, pp.
  \bibinfo{pages}{2961--2969}.
\bibitem[{He et~al.(2016)He, Zhang, Ren and Sun}]{he2016deep}
\bibinfo{author}{He, K.}, \bibinfo{author}{Zhang, X.}, \bibinfo{author}{Ren,
  S.}, \bibinfo{author}{Sun, J.}, \bibinfo{year}{2016}.
\newblock \bibinfo{title}{Deep residual learning for image recognition}, in:
  \bibinfo{booktitle}{Proceedings of the IEEE conference on computer vision and
  pattern recognition}, pp. \bibinfo{pages}{770--778}.
\bibitem[{Hinterstoi{\ss}er et~al.(2012)Hinterstoi{\ss}er, Lepetit, Ilic,
  Holzer, Bradski, Konolige and Navab}]{Hinterstoier2012ModelBT}
\bibinfo{author}{Hinterstoi{\ss}er, S.}, \bibinfo{author}{Lepetit, V.},
  \bibinfo{author}{Ilic, S.}, \bibinfo{author}{Holzer, S.},
  \bibinfo{author}{Bradski, G.R.}, \bibinfo{author}{Konolige, K.},
  \bibinfo{author}{Navab, N.}, \bibinfo{year}{2012}.
\newblock \bibinfo{title}{Model based training, detection and pose estimation
  of texture-less 3d objects in heavily cluttered scenes}, in:
  \bibinfo{booktitle}{Asian Conference on Computer Vision}.
\bibitem[{Hochreiter and Schmidhuber(1997)}]{10.1162/neco.1997.9.8.1735}
\bibinfo{author}{Hochreiter, S.}, \bibinfo{author}{Schmidhuber, J.},
  \bibinfo{year}{1997}.
\newblock \bibinfo{title}{Long short-term memory}.
\newblock \bibinfo{journal}{Neural Comput.} \bibinfo{volume}{9},
  \bibinfo{pages}{1735–1780}.
\newblock \URLprefix \url{https://doi.org/10.1162/neco.1997.9.8.1735},
  \DOIprefix\doi{10.1162/neco.1997.9.8.1735}.
\bibitem[{Hogan et~al.(2021)Hogan, Rondao, Aouf and
  Dubois{-}Matra}]{DBLP:journals/corr/abs-2105-13789}
\bibinfo{author}{Hogan, M.}, \bibinfo{author}{Rondao, D.},
  \bibinfo{author}{Aouf, N.}, \bibinfo{author}{Dubois{-}Matra, O.},
  \bibinfo{year}{2021}.
\newblock \bibinfo{title}{Using convolutional neural networks for relative pose
  estimation of a non-cooperative spacecraft with thermal infrared imagery}.
\newblock \bibinfo{journal}{CoRR} \bibinfo{volume}{abs/2105.13789}.
\newblock \URLprefix \url{https://arxiv.org/abs/2105.13789},
  \href{http://arxiv.org/abs/2105.13789}{\tt arXiv:2105.13789}.
\bibitem[{Hou et~al.(2020)Hou, Ahmadyan, Zhang, Wei and
  Grundmann}]{hou2020mobilepose}
\bibinfo{author}{Hou, T.}, \bibinfo{author}{Ahmadyan, A.},
  \bibinfo{author}{Zhang, L.}, \bibinfo{author}{Wei, J.},
  \bibinfo{author}{Grundmann, M.}, \bibinfo{year}{2020}.
\newblock \bibinfo{title}{Mobilepose: Real-time pose estimation for unseen
  objects with weak shape supervision}.
\newblock \bibinfo{journal}{arXiv preprint arXiv:2003.03522} .
\bibitem[{Howard et~al.(2019)Howard, Sandler, Chu, Chen, Chen, Tan, Wang, Zhu,
  Pang, Vasudevan et~al.}]{howard2019searching}
\bibinfo{author}{Howard, A.}, \bibinfo{author}{Sandler, M.},
  \bibinfo{author}{Chu, G.}, \bibinfo{author}{Chen, L.C.},
  \bibinfo{author}{Chen, B.}, \bibinfo{author}{Tan, M.}, \bibinfo{author}{Wang,
  W.}, \bibinfo{author}{Zhu, Y.}, \bibinfo{author}{Pang, R.},
  \bibinfo{author}{Vasudevan, V.}, et~al., \bibinfo{year}{2019}.
\newblock \bibinfo{title}{Searching for mobilenetv3}, in:
  \bibinfo{booktitle}{Proceedings of the IEEE/CVF international conference on
  computer vision}, pp. \bibinfo{pages}{1314--1324}.
\bibitem[{Hu et~al.(2018)Hu, Shen and Sun}]{hu2018squeeze}
\bibinfo{author}{Hu, J.}, \bibinfo{author}{Shen, L.}, \bibinfo{author}{Sun,
  G.}, \bibinfo{year}{2018}.
\newblock \bibinfo{title}{Squeeze-and-excitation networks}, in:
  \bibinfo{booktitle}{Proceedings of the IEEE conference on computer vision and
  pattern recognition}, pp. \bibinfo{pages}{7132--7141}.
\bibitem[{Hu et~al.(2020)Hu, Fua, Wang and Salzmann}]{hu2020single}
\bibinfo{author}{Hu, Y.}, \bibinfo{author}{Fua, P.}, \bibinfo{author}{Wang,
  W.}, \bibinfo{author}{Salzmann, M.}, \bibinfo{year}{2020}.
\newblock \bibinfo{title}{Single-stage 6d object pose estimation}, in:
  \bibinfo{booktitle}{Proceedings of the IEEE/CVF conference on computer vision
  and pattern recognition}, pp. \bibinfo{pages}{2930--2939}.
\bibitem[{Hu et~al.(2019)Hu, Hugonot, Fua and Salzmann}]{hu2019segmentation}
\bibinfo{author}{Hu, Y.}, \bibinfo{author}{Hugonot, J.}, \bibinfo{author}{Fua,
  P.}, \bibinfo{author}{Salzmann, M.}, \bibinfo{year}{2019}.
\newblock \bibinfo{title}{Segmentation-driven 6d object pose estimation}, in:
  \bibinfo{booktitle}{Proceedings of the IEEE/CVF Conference on Computer Vision
  and Pattern Recognition}, pp. \bibinfo{pages}{3385--3394}.
\bibitem[{Hu et~al.(2021)Hu, Speierer, Jakob, Fua and Salzmann}]{hu2021}
\bibinfo{author}{Hu, Y.}, \bibinfo{author}{Speierer, S.},
  \bibinfo{author}{Jakob, W.}, \bibinfo{author}{Fua, P.},
  \bibinfo{author}{Salzmann, M.}, \bibinfo{year}{2021}.
\newblock \bibinfo{title}{Wide-depth-range 6d object pose estimation in space},
  in: \bibinfo{booktitle}{Proceedings of the IEEE/CVF Conference on Computer
  Vision and Pattern Recognition}, pp. \bibinfo{pages}{15870--15879}.
\bibitem[{Huan et~al.(2020)Huan, Liu and Hu}]{huan2020pose}
\bibinfo{author}{Huan, W.}, \bibinfo{author}{Liu, M.}, \bibinfo{author}{Hu,
  Q.}, \bibinfo{year}{2020}.
\newblock \bibinfo{title}{Pose estimation for non-cooperative spacecraft based
  on deep learning}, in: \bibinfo{booktitle}{2020 39th Chinese Control
  Conference (CCC)}, \bibinfo{organization}{IEEE}. pp.
  \bibinfo{pages}{3339--3343}.
\bibitem[{Huang et~al.(2021)Huang, Zhao, Gu and Bo}]{huang2021non}
\bibinfo{author}{Huang, H.}, \bibinfo{author}{Zhao, G.}, \bibinfo{author}{Gu,
  D.}, \bibinfo{author}{Bo, Y.}, \bibinfo{year}{2021}.
\newblock \bibinfo{title}{Non-model-based monocular pose estimation network for
  uncooperative spacecraft using convolutional neural network}.
\newblock \bibinfo{journal}{IEEE Sensors Journal} \bibinfo{volume}{21},
  \bibinfo{pages}{24579--24590}.
\bibitem[{Huo et~al.(2020)Huo, Li and Zhang}]{huo2020}
\bibinfo{author}{Huo, Y.}, \bibinfo{author}{Li, Z.}, \bibinfo{author}{Zhang,
  F.}, \bibinfo{year}{2020}.
\newblock \bibinfo{title}{Fast and accurate spacecraft pose estimation from
  single shot space imagery using box reliability and keypoints existence
  judgments}.
\newblock \bibinfo{journal}{IEEE Access} \bibinfo{volume}{8},
  \bibinfo{pages}{216283--216297}.
\bibitem[{Huynh(2009)}]{huynh2009metrics}
\bibinfo{author}{Huynh, D.Q.}, \bibinfo{year}{2009}.
\newblock \bibinfo{title}{Metrics for 3d rotations: Comparison and analysis}.
\newblock \bibinfo{journal}{Journal of Mathematical Imaging and Vision}
  \bibinfo{volume}{35}, \bibinfo{pages}{155--164}.
\bibitem[{Jackson et~al.(2018)Jackson, Abarghouei, Bonner, Breckon and
  Obara}]{DBLP:journals/corr/abs-1809-05375}
\bibinfo{author}{Jackson, P.T.G.}, \bibinfo{author}{Abarghouei, A.A.},
  \bibinfo{author}{Bonner, S.}, \bibinfo{author}{Breckon, T.P.},
  \bibinfo{author}{Obara, B.}, \bibinfo{year}{2018}.
\newblock \bibinfo{title}{Style augmentation: Data augmentation via style
  randomization}.
\newblock \bibinfo{journal}{CoRR} \bibinfo{volume}{abs/1809.05375}.
\newblock \URLprefix \url{http://arxiv.org/abs/1809.05375},
  \href{http://arxiv.org/abs/1809.05375}{\tt arXiv:1809.05375}.
\bibitem[{Jawaid et~al.(2022)Jawaid, Elms, Latif and Chin}]{seenic}
\bibinfo{author}{Jawaid, M.}, \bibinfo{author}{Elms, E.},
  \bibinfo{author}{Latif, Y.}, \bibinfo{author}{Chin, T.J.},
  \bibinfo{year}{2022}.
\newblock \bibinfo{title}{Towards bridging the space domain gap for satellite
  pose estimation using event sensing}.
\newblock \URLprefix \url{https://arxiv.org/abs/2209.11945},
  \DOIprefix\doi{10.48550/ARXIV.2209.11945}.
\bibitem[{Jiao et~al.(2019)Jiao, Zhang, Liu, Yang, Li, Feng and
  Qu}]{jiao2019survey}
\bibinfo{author}{Jiao, L.}, \bibinfo{author}{Zhang, F.}, \bibinfo{author}{Liu,
  F.}, \bibinfo{author}{Yang, S.}, \bibinfo{author}{Li, L.},
  \bibinfo{author}{Feng, Z.}, \bibinfo{author}{Qu, R.}, \bibinfo{year}{2019}.
\newblock \bibinfo{title}{A survey of deep learning-based object detection}.
\newblock \bibinfo{journal}{IEEE access} \bibinfo{volume}{7},
  \bibinfo{pages}{128837--128868}.
\bibitem[{Jones(2018)}]{jones2018recent}
\bibinfo{author}{Jones, H.}, \bibinfo{year}{2018}.
\newblock \bibinfo{title}{The recent large reduction in space launch cost},
  \bibinfo{organization}{48th International Conference on Environmental
  Systems}.
\bibitem[{Kelsey et~al.(2006)Kelsey, Byrne, Cosgrove, Seereeram and
  Mehra}]{kelsey2006}
\bibinfo{author}{Kelsey, J.}, \bibinfo{author}{Byrne, J.},
  \bibinfo{author}{Cosgrove, M.}, \bibinfo{author}{Seereeram, S.},
  \bibinfo{author}{Mehra, R.}, \bibinfo{year}{2006}.
\newblock \bibinfo{title}{Vision-based relative pose estimation for autonomous
  rendezvous and docking}, in: \bibinfo{booktitle}{2006 IEEE Aerospace
  Conference}, pp. \bibinfo{pages}{20 pp.--}.
\newblock \DOIprefix\doi{10.1109/AERO.2006.1655916}.
\bibitem[{Kendall and Cipolla(2017)}]{kendall2017geometric}
\bibinfo{author}{Kendall, A.}, \bibinfo{author}{Cipolla, R.},
  \bibinfo{year}{2017}.
\newblock \bibinfo{title}{Geometric loss functions for camera pose regression
  with deep learning}, in: \bibinfo{booktitle}{Proceedings of the IEEE
  conference on computer vision and pattern recognition}, pp.
  \bibinfo{pages}{5974--5983}.
\bibitem[{Kendall and Gal(2017)}]{kendall2017uncertainties}
\bibinfo{author}{Kendall, A.}, \bibinfo{author}{Gal, Y.}, \bibinfo{year}{2017}.
\newblock \bibinfo{title}{What uncertainties do we need in bayesian deep
  learning for computer vision?}
\newblock \bibinfo{journal}{Advances in neural information processing systems}
  \bibinfo{volume}{30}.
\bibitem[{Kendall et~al.(2015)Kendall, Grimes and Cipolla}]{Kendall_2015_ICCV}
\bibinfo{author}{Kendall, A.}, \bibinfo{author}{Grimes, M.},
  \bibinfo{author}{Cipolla, R.}, \bibinfo{year}{2015}.
\newblock \bibinfo{title}{Posenet: A convolutional network for real-time 6-dof
  camera relocalization}, in: \bibinfo{booktitle}{Proceedings of the IEEE
  International Conference on Computer Vision (ICCV)}.
\bibitem[{{Kendall} et~al.(2015){Kendall}, {Grimes} and
  {Cipolla}}]{PoseNet2016}
\bibinfo{author}{{Kendall}, A.}, \bibinfo{author}{{Grimes}, M.},
  \bibinfo{author}{{Cipolla}, R.}, \bibinfo{year}{2015}.
\newblock \bibinfo{title}{{PoseNet: A Convolutional Network for Real-Time 6-DOF
  Camera Relocalization}}.
\newblock \bibinfo{journal}{arXiv e-prints} ,
  \bibinfo{pages}{arXiv:1505.07427}\href{http://arxiv.org/abs/1505.07427}{\tt
  arXiv:1505.07427}.
\bibitem[{Kisantal et~al.(2020)Kisantal, Sharma, Park, Izzo, M{\"a}rtens and
  D’Amico}]{kisantal2020}
\bibinfo{author}{Kisantal, M.}, \bibinfo{author}{Sharma, S.},
  \bibinfo{author}{Park, T.H.}, \bibinfo{author}{Izzo, D.},
  \bibinfo{author}{M{\"a}rtens, M.}, \bibinfo{author}{D’Amico, S.},
  \bibinfo{year}{2020}.
\newblock \bibinfo{title}{Satellite pose estimation challenge: Dataset,
  competition design, and results}.
\newblock \bibinfo{journal}{IEEE Transactions on Aerospace and Electronic
  Systems} \bibinfo{volume}{56}, \bibinfo{pages}{4083--4098}.
\bibitem[{Kosmidis et~al.(2020)Kosmidis, Rodriguez, Jover, Alcaide, Lachaize,
  Abella, Notebaert, Cazorla and Steenari}]{kosmidis2020gpu4s}
\bibinfo{author}{Kosmidis, L.}, \bibinfo{author}{Rodriguez, I.},
  \bibinfo{author}{Jover, {\'A}.}, \bibinfo{author}{Alcaide, S.},
  \bibinfo{author}{Lachaize, J.}, \bibinfo{author}{Abella, J.},
  \bibinfo{author}{Notebaert, O.}, \bibinfo{author}{Cazorla, F.J.},
  \bibinfo{author}{Steenari, D.}, \bibinfo{year}{2020}.
\newblock \bibinfo{title}{Gpu4s: Embedded gpus in space-latest project
  updates}.
\newblock \bibinfo{journal}{Microprocessors and Microsystems}
  \bibinfo{volume}{77}, \bibinfo{pages}{103143}.
\bibitem[{Kreisel(2002)}]{kreisel2002orbit}
\bibinfo{author}{Kreisel, J.}, \bibinfo{year}{2002}.
\newblock \bibinfo{title}{On-orbit servicing of satellites (oos): its potential
  market \& impact}, in: \bibinfo{booktitle}{proceedings of 7th ESA Workshop on
  Advanced Space Technologies for Robotics and Automation ‘ASTRA}.
\bibitem[{Krizhevsky et~al.(2012)Krizhevsky, Sutskever and
  Hinton}]{Krizhevsky2012ImageNetCW}
\bibinfo{author}{Krizhevsky, A.}, \bibinfo{author}{Sutskever, I.},
  \bibinfo{author}{Hinton, G.E.}, \bibinfo{year}{2012}.
\newblock \bibinfo{title}{Imagenet classification with deep convolutional
  neural networks}.
\newblock \bibinfo{journal}{Communications of the ACM} \bibinfo{volume}{60},
  \bibinfo{pages}{84 -- 90}.
\bibitem[{Legrand et~al.()Legrand, Detry and De~Vleeschouwer}]{legrandend}
\bibinfo{author}{Legrand, A.}, \bibinfo{author}{Detry, R.},
  \bibinfo{author}{De~Vleeschouwer, C.}, .
\newblock \bibinfo{title}{End-to-end neural estimation of spacecraft pose with
  intermediate detection of keypoints} .
\bibitem[{Lengyel et~al.(2021)Lengyel, Garg, Milford and van
  Gemert}]{lengyel2021zero}
\bibinfo{author}{Lengyel, A.}, \bibinfo{author}{Garg, S.},
  \bibinfo{author}{Milford, M.}, \bibinfo{author}{van Gemert, J.C.},
  \bibinfo{year}{2021}.
\newblock \bibinfo{title}{Zero-shot day-night domain adaptation with a physics
  prior}, in: \bibinfo{booktitle}{Proceedings of the IEEE/CVF International
  Conference on Computer Vision}, pp. \bibinfo{pages}{4399--4409}.
\bibitem[{Leon et~al.(2022)Leon, Lentaris, Soudris, Vellas and
  Bernou}]{leon2022towards}
\bibinfo{author}{Leon, V.}, \bibinfo{author}{Lentaris, G.},
  \bibinfo{author}{Soudris, D.}, \bibinfo{author}{Vellas, S.},
  \bibinfo{author}{Bernou, M.}, \bibinfo{year}{2022}.
\newblock \bibinfo{title}{Towards employing fpga and asip acceleration to
  enable onboard ai/ml in space applications}, in: \bibinfo{booktitle}{2022
  IFIP/IEEE 30th International Conference on Very Large Scale Integration
  (VLSI-SoC)}, \bibinfo{organization}{IEEE}. pp. \bibinfo{pages}{1--4}.
\bibitem[{Leong et~al.(2020a)Leong, Prasad, Lee and Lin}]{leong2020semi}
\bibinfo{author}{Leong, M.C.}, \bibinfo{author}{Prasad, D.K.},
  \bibinfo{author}{Lee, Y.T.}, \bibinfo{author}{Lin, F.},
  \bibinfo{year}{2020}a.
\newblock \bibinfo{title}{Semi-cnn architecture for effective spatio-temporal
  learning in action recognition}.
\newblock \bibinfo{journal}{Applied Sciences} \bibinfo{volume}{10},
  \bibinfo{pages}{557}.
\bibitem[{Leong et~al.(2020b)Leong, Prasad, Lee and Lin}]{leong2020}
\bibinfo{author}{Leong, M.C.}, \bibinfo{author}{Prasad, D.K.},
  \bibinfo{author}{Lee, Y.T.}, \bibinfo{author}{Lin, F.},
  \bibinfo{year}{2020}b.
\newblock \bibinfo{title}{Semi-cnn architecture for effective spatio-temporal
  learning in action recognition}.
\newblock \bibinfo{journal}{Applied Sciences} \bibinfo{volume}{10},
  \bibinfo{pages}{557}.
\bibitem[{Lepetit et~al.(2009)Lepetit, Moreno-Noguer and Fua}]{lepetit2009}
\bibinfo{author}{Lepetit, V.}, \bibinfo{author}{Moreno-Noguer, F.},
  \bibinfo{author}{Fua, P.}, \bibinfo{year}{2009}.
\newblock \bibinfo{title}{Epnp: An accurate o (n) solution to the pnp problem}.
\newblock \bibinfo{journal}{International journal of computer vision}
  \bibinfo{volume}{81}, \bibinfo{pages}{155}.
\bibitem[{Li et~al.(2022)Li, Zhang and Hu}]{kecenli2022}
\bibinfo{author}{Li, K.}, \bibinfo{author}{Zhang, H.}, \bibinfo{author}{Hu,
  C.}, \bibinfo{year}{2022}.
\newblock \bibinfo{title}{Learning-based pose estimation of non-cooperative
  spacecrafts with uncertainty prediction}.
\newblock \bibinfo{journal}{Aerospace} \bibinfo{volume}{9}.
\newblock \URLprefix \url{https://www.mdpi.com/2226-4310/9/10/592},
  \DOIprefix\doi{10.3390/aerospace9100592}.
\bibitem[{Li et~al.(2018)Li, Liu, Chen and Rudin}]{li2018deep}
\bibinfo{author}{Li, O.}, \bibinfo{author}{Liu, H.}, \bibinfo{author}{Chen,
  C.}, \bibinfo{author}{Rudin, C.}, \bibinfo{year}{2018}.
\newblock \bibinfo{title}{Deep learning for case-based reasoning through
  prototypes: A neural network that explains its predictions}, in:
  \bibinfo{booktitle}{Proceedings of the AAAI Conference on Artificial
  Intelligence}.
\bibitem[{Li et~al.(2019)Li, Cheng, Liu, Wang, Shi, Tang, Gao, Zeng, Chai, Luo
  et~al.}]{li2019orbit}
\bibinfo{author}{Li, W.J.}, \bibinfo{author}{Cheng, D.Y.},
  \bibinfo{author}{Liu, X.G.}, \bibinfo{author}{Wang, Y.B.},
  \bibinfo{author}{Shi, W.H.}, \bibinfo{author}{Tang, Z.X.},
  \bibinfo{author}{Gao, F.}, \bibinfo{author}{Zeng, F.M.},
  \bibinfo{author}{Chai, H.Y.}, \bibinfo{author}{Luo, W.B.}, et~al.,
  \bibinfo{year}{2019}.
\newblock \bibinfo{title}{On-orbit service (oos) of spacecraft: A review of
  engineering developments}.
\newblock \bibinfo{journal}{Progress in Aerospace Sciences}
  \bibinfo{volume}{108}, \bibinfo{pages}{32--120}.
\bibitem[{Lin et~al.(2014a)Lin, Maire, Belongie, Hays, Perona, Ramanan,
  Doll{\'{a}}r and Zitnick}]{DBLP:conf/eccv/LinMBHPRDZ14}
\bibinfo{author}{Lin, T.}, \bibinfo{author}{Maire, M.},
  \bibinfo{author}{Belongie, S.J.}, \bibinfo{author}{Hays, J.},
  \bibinfo{author}{Perona, P.}, \bibinfo{author}{Ramanan, D.},
  \bibinfo{author}{Doll{\'{a}}r, P.}, \bibinfo{author}{Zitnick, C.L.},
  \bibinfo{year}{2014}a.
\newblock \bibinfo{title}{Microsoft {COCO:} common objects in context}, in:
  \bibinfo{editor}{Fleet, D.J.}, \bibinfo{editor}{Pajdla, T.},
  \bibinfo{editor}{Schiele, B.}, \bibinfo{editor}{Tuytelaars, T.} (Eds.),
  \bibinfo{booktitle}{Computer Vision - {ECCV} 2014 - 13th European Conference,
  Zurich, Switzerland, September 6-12, 2014, Proceedings, Part {V}},
  \bibinfo{publisher}{Springer}. pp. \bibinfo{pages}{740--755}.
\newblock \URLprefix \url{https://doi.org/10.1007/978-3-319-10602-1\_48},
  \DOIprefix\doi{10.1007/978-3-319-10602-1\_48}.
\bibitem[{Lin et~al.(2017)Lin, Doll{\'a}r, Girshick, He, Hariharan and
  Belongie}]{lin2017feature}
\bibinfo{author}{Lin, T.Y.}, \bibinfo{author}{Doll{\'a}r, P.},
  \bibinfo{author}{Girshick, R.}, \bibinfo{author}{He, K.},
  \bibinfo{author}{Hariharan, B.}, \bibinfo{author}{Belongie, S.},
  \bibinfo{year}{2017}.
\newblock \bibinfo{title}{Feature pyramid networks for object detection}, in:
  \bibinfo{booktitle}{Proceedings of the IEEE conference on computer vision and
  pattern recognition}, pp. \bibinfo{pages}{2117--2125}.
\bibitem[{Lin et~al.(2014b)Lin, Maire, Belongie, Hays, Perona, Ramanan,
  Doll{\'a}r and Zitnick}]{lin2014microsoft}
\bibinfo{author}{Lin, T.Y.}, \bibinfo{author}{Maire, M.},
  \bibinfo{author}{Belongie, S.}, \bibinfo{author}{Hays, J.},
  \bibinfo{author}{Perona, P.}, \bibinfo{author}{Ramanan, D.},
  \bibinfo{author}{Doll{\'a}r, P.}, \bibinfo{author}{Zitnick, C.L.},
  \bibinfo{year}{2014}b.
\newblock \bibinfo{title}{Microsoft coco: Common objects in context}, in:
  \bibinfo{booktitle}{Computer Vision--ECCV 2014: 13th European Conference,
  Zurich, Switzerland, September 6-12, 2014, Proceedings, Part V 13},
  \bibinfo{organization}{Springer}. pp. \bibinfo{pages}{740--755}.
\bibitem[{Liu and Hu(2014)}]{liu2014relative}
\bibinfo{author}{Liu, C.}, \bibinfo{author}{Hu, W.}, \bibinfo{year}{2014}.
\newblock \bibinfo{title}{Relative pose estimation for cylinder-shaped
  spacecrafts using single image}.
\newblock \bibinfo{journal}{IEEE Transactions on Aerospace and Electronic
  Systems} \bibinfo{volume}{50}, \bibinfo{pages}{3036--3056}.
\bibitem[{Liu et~al.(2016)Liu, Anguelov, Erhan, Szegedy, Reed, Fu and
  Berg}]{liu2016ssd}
\bibinfo{author}{Liu, W.}, \bibinfo{author}{Anguelov, D.},
  \bibinfo{author}{Erhan, D.}, \bibinfo{author}{Szegedy, C.},
  \bibinfo{author}{Reed, S.}, \bibinfo{author}{Fu, C.Y.},
  \bibinfo{author}{Berg, A.C.}, \bibinfo{year}{2016}.
\newblock \bibinfo{title}{Ssd: Single shot multibox detector}, in:
  \bibinfo{booktitle}{European conference on computer vision},
  \bibinfo{organization}{Springer}. pp. \bibinfo{pages}{21--37}.
\bibitem[{Llorente et~al.(2013)Llorente, Agenjo, Carrascosa, de~Negueruela,
  Mestreau-Garreau, Cropp and Santovincenzo}]{llorente2013proba}
\bibinfo{author}{Llorente, J.S.}, \bibinfo{author}{Agenjo, A.},
  \bibinfo{author}{Carrascosa, C.}, \bibinfo{author}{de~Negueruela, C.},
  \bibinfo{author}{Mestreau-Garreau, A.}, \bibinfo{author}{Cropp, A.},
  \bibinfo{author}{Santovincenzo, A.}, \bibinfo{year}{2013}.
\newblock \bibinfo{title}{Proba-3: Precise formation flying demonstration
  mission}.
\newblock \bibinfo{journal}{Acta Astronautica} \bibinfo{volume}{82},
  \bibinfo{pages}{38--46}.
\bibitem[{Long et~al.(2020a)Long, Deng, Wang, Zhang, Dang, Gao, Shen, Ren, Han,
  Ding et~al.}]{long2020}
\bibinfo{author}{Long, X.}, \bibinfo{author}{Deng, K.}, \bibinfo{author}{Wang,
  G.}, \bibinfo{author}{Zhang, Y.}, \bibinfo{author}{Dang, Q.},
  \bibinfo{author}{Gao, Y.}, \bibinfo{author}{Shen, H.}, \bibinfo{author}{Ren,
  J.}, \bibinfo{author}{Han, S.}, \bibinfo{author}{Ding, E.}, et~al.,
  \bibinfo{year}{2020}a.
\newblock \bibinfo{title}{Pp-yolo: An effective and efficient implementation of
  object detector}.
\newblock \bibinfo{journal}{arXiv preprint arXiv:2007.12099} .
\bibitem[{Long et~al.(2020b)Long, Deng, Wang, Zhang, Dang, Gao, Shen, Ren, Han,
  Ding et~al.}]{long2020pp}
\bibinfo{author}{Long, X.}, \bibinfo{author}{Deng, K.}, \bibinfo{author}{Wang,
  G.}, \bibinfo{author}{Zhang, Y.}, \bibinfo{author}{Dang, Q.},
  \bibinfo{author}{Gao, Y.}, \bibinfo{author}{Shen, H.}, \bibinfo{author}{Ren,
  J.}, \bibinfo{author}{Han, S.}, \bibinfo{author}{Ding, E.}, et~al.,
  \bibinfo{year}{2020}b.
\newblock \bibinfo{title}{Pp-yolo: An effective and efficient implementation of
  object detector}.
\newblock \bibinfo{journal}{arXiv preprint arXiv:2007.12099} .
\bibitem[{{Lotti} et~al.(2022){Lotti}, {Modenini}, {Tortora}, {Saponara} and
  {Perino}}]{spetpu22}
\bibinfo{author}{{Lotti}, A.}, \bibinfo{author}{{Modenini}, D.},
  \bibinfo{author}{{Tortora}, P.}, \bibinfo{author}{{Saponara}, M.},
  \bibinfo{author}{{Perino}, M.A.}, \bibinfo{year}{2022}.
\newblock \bibinfo{title}{{Deep Learning for Real Time Satellite Pose
  Estimation on Low Power Edge TPU}}.
\newblock \bibinfo{journal}{arXiv e-prints} ,
  \bibinfo{pages}{arXiv:2204.03296}\href{http://arxiv.org/abs/2204.03296}{\tt
  arXiv:2204.03296}.
\bibitem[{Lotti et~al.(2022)Lotti, Modenini, Tortora, Saponara and
  Perino}]{lotti2022deep}
\bibinfo{author}{Lotti, A.}, \bibinfo{author}{Modenini, D.},
  \bibinfo{author}{Tortora, P.}, \bibinfo{author}{Saponara, M.},
  \bibinfo{author}{Perino, M.A.}, \bibinfo{year}{2022}.
\newblock \bibinfo{title}{Deep learning for real time satellite pose estimation
  on low power edge tpu}.
\newblock \bibinfo{journal}{arXiv preprint arXiv:2204.03296} .
\bibitem[{Marchand et~al.(2015)Marchand, Uchiyama and
  Spindler}]{marchand2015pose}
\bibinfo{author}{Marchand, E.}, \bibinfo{author}{Uchiyama, H.},
  \bibinfo{author}{Spindler, F.}, \bibinfo{year}{2015}.
\newblock \bibinfo{title}{Pose estimation for augmented reality: a hands-on
  survey}.
\newblock \bibinfo{journal}{IEEE transactions on visualization and computer
  graphics} \bibinfo{volume}{22}, \bibinfo{pages}{2633--2651}.
\bibitem[{Martin et~al.(2019)Martin, Dunstan and Gestido}]{martin2019planetary}
\bibinfo{author}{Martin, I.}, \bibinfo{author}{Dunstan, M.},
  \bibinfo{author}{Gestido, M.S.}, \bibinfo{year}{2019}.
\newblock \bibinfo{title}{Planetary surface image generation for testing future
  space missions with pangu}, in: \bibinfo{booktitle}{2nd RPI Space Imaging
  Workshop}, \bibinfo{organization}{Sensing, Estimation, and Automation
  Laboratory}.
\bibitem[{Marullo et~al.(2022)Marullo, Tanzi, Piazzolla and
  Vezzetti}]{marullo20226d}
\bibinfo{author}{Marullo, G.}, \bibinfo{author}{Tanzi, L.},
  \bibinfo{author}{Piazzolla, P.}, \bibinfo{author}{Vezzetti, E.},
  \bibinfo{year}{2022}.
\newblock \bibinfo{title}{6d object position estimation from 2d images: a
  literature review}.
\newblock \bibinfo{journal}{Multimedia Tools and Applications} ,
  \bibinfo{pages}{1--39}.
\bibitem[{May(2021)}]{may2021triggers}
\bibinfo{author}{May, C.}, \bibinfo{year}{2021}.
\newblock \bibinfo{title}{Triggers and effects of an active debris removal
  market}.
\newblock \bibinfo{journal}{The Aerospace Corporation, Center for Space Policy
  and Strategy, Tech. Rep} , \bibinfo{pages}{2021--01}.
\bibitem[{Mertan et~al.(2022)Mertan, Duff and Unal}]{mertan2022single}
\bibinfo{author}{Mertan, A.}, \bibinfo{author}{Duff, D.J.},
  \bibinfo{author}{Unal, G.}, \bibinfo{year}{2022}.
\newblock \bibinfo{title}{Single image depth estimation: An overview}.
\newblock \bibinfo{journal}{Digital Signal Processing} ,
  \bibinfo{pages}{103441}.
\bibitem[{Minaee et~al.(2021)Minaee, Boykov, Porikli, Plaza, Kehtarnavaz and
  Terzopoulos}]{minaee2021image}
\bibinfo{author}{Minaee, S.}, \bibinfo{author}{Boykov, Y.Y.},
  \bibinfo{author}{Porikli, F.}, \bibinfo{author}{Plaza, A.J.},
  \bibinfo{author}{Kehtarnavaz, N.}, \bibinfo{author}{Terzopoulos, D.},
  \bibinfo{year}{2021}.
\newblock \bibinfo{title}{Image segmentation using deep learning: A survey}.
\newblock \bibinfo{journal}{IEEE transactions on pattern analysis and machine
  intelligence} .
\bibitem[{Mittelhammer et~al.(2000)Mittelhammer, Judge and
  Miller}]{mittelhammer2000econometric}
\bibinfo{author}{Mittelhammer, R.C.}, \bibinfo{author}{Judge, G.G.},
  \bibinfo{author}{Miller, D.J.}, \bibinfo{year}{2000}.
\newblock \bibinfo{title}{Econometric foundations}.
\newblock \bibinfo{publisher}{Cambridge University Press}.
\bibitem[{Mor{\'e}(1978)}]{more1978levenberg}
\bibinfo{author}{Mor{\'e}, J.J.}, \bibinfo{year}{1978}.
\newblock \bibinfo{title}{The levenberg-marquardt algorithm: implementation and
  theory}, in: \bibinfo{booktitle}{Numerical analysis}.
  \bibinfo{publisher}{Springer}, pp. \bibinfo{pages}{105--116}.
\bibitem[{Mumuni and Mumuni(2022)}]{MUMUNI2022100258}
\bibinfo{author}{Mumuni, A.}, \bibinfo{author}{Mumuni, F.},
  \bibinfo{year}{2022}.
\newblock \bibinfo{title}{Data augmentation: A comprehensive survey of modern
  approaches}.
\newblock \bibinfo{journal}{Array} \bibinfo{volume}{16},
  \bibinfo{pages}{100258}.
\newblock \URLprefix
  \url{https://www.sciencedirect.com/science/article/pii/S2590005622000911},
  \DOIprefix\doi{https://doi.org/10.1016/j.array.2022.100258}.
\bibitem[{Musallam et~al.(2022)Musallam, Gaudilli\`ere, del Castillo,
  Al~Ismaeil and Aouada}]{Musallam_2022_CVPR}
\bibinfo{author}{Musallam, M.A.}, \bibinfo{author}{Gaudilli\`ere, V.},
  \bibinfo{author}{del Castillo, M.O.}, \bibinfo{author}{Al~Ismaeil, K.},
  \bibinfo{author}{Aouada, D.}, \bibinfo{year}{2022}.
\newblock \bibinfo{title}{Leveraging equivariant features for absolute pose
  regression}, in: \bibinfo{booktitle}{Proceedings of the IEEE/CVF Conference
  on Computer Vision and Pattern Recognition (CVPR)}, pp.
  \bibinfo{pages}{6876--6886}.
\bibitem[{OpenCV()}]{opencv}
\bibinfo{author}{OpenCV}, .
\newblock \bibinfo{title}{Perspective-n-point (pnp) pose computation}.
\newblock \URLprefix
  \url{https://docs.opencv.org/4.x/d5/d1f/calib3d_solvePnP.html}.
\bibitem[{Opromolla et~al.(2017)Opromolla, Fasano, Rufino and
  Grassi}]{opromolla2017review}
\bibinfo{author}{Opromolla, R.}, \bibinfo{author}{Fasano, G.},
  \bibinfo{author}{Rufino, G.}, \bibinfo{author}{Grassi, M.},
  \bibinfo{year}{2017}.
\newblock \bibinfo{title}{A review of cooperative and uncooperative spacecraft
  pose determination techniques for close-proximity operations}.
\newblock \bibinfo{journal}{Progress in Aerospace Sciences}
  \bibinfo{volume}{93}, \bibinfo{pages}{53--72}.
\bibitem[{Park et~al.(2020)Park, Mousavian, Xiang and
  Fox}]{park2020latentfusion}
\bibinfo{author}{Park, K.}, \bibinfo{author}{Mousavian, A.},
  \bibinfo{author}{Xiang, Y.}, \bibinfo{author}{Fox, D.}, \bibinfo{year}{2020}.
\newblock \bibinfo{title}{Latentfusion: End-to-end differentiable
  reconstruction and rendering for unseen object pose estimation}, in:
  \bibinfo{booktitle}{Proceedings of the IEEE/CVF conference on computer vision
  and pattern recognition}, pp. \bibinfo{pages}{10710--10719}.
\bibitem[{Park et~al.(2019a)Park, Patten and Vincze}]{park2019pix2pose}
\bibinfo{author}{Park, K.}, \bibinfo{author}{Patten, T.},
  \bibinfo{author}{Vincze, M.}, \bibinfo{year}{2019}a.
\newblock \bibinfo{title}{Pix2pose: Pixel-wise coordinate regression of objects
  for 6d pose estimation}, in: \bibinfo{booktitle}{Proceedings of the IEEE/CVF
  International Conference on Computer Vision}, pp.
  \bibinfo{pages}{7668--7677}.
\bibitem[{Park et~al.(2021)Park, Bosse and D'Amico}]{park2021robotic}
\bibinfo{author}{Park, T.H.}, \bibinfo{author}{Bosse, J.},
  \bibinfo{author}{D'Amico, S.}, \bibinfo{year}{2021}.
\newblock \bibinfo{title}{Robotic testbed for rendezvous and optical
  navigation: Multi-source calibration and machine learning use cases}.
\newblock \bibinfo{journal}{arXiv preprint arXiv:2108.05529} .
\bibitem[{Park and D'Amico(2022a)}]{park2022cnnukf}
\bibinfo{author}{Park, T.H.}, \bibinfo{author}{D'Amico, S.},
  \bibinfo{year}{2022}a.
\newblock \bibinfo{title}{Adaptive neural network-based unscented kalman filter
  for spacecraft pose tracking at rendezvous}.
\newblock \bibinfo{journal}{arXiv preprint arXiv:2206.03796} .
\bibitem[{Park and D'Amico(2022b)}]{park2022robust}
\bibinfo{author}{Park, T.H.}, \bibinfo{author}{D'Amico, S.},
  \bibinfo{year}{2022}b.
\newblock \bibinfo{title}{Robust multi-task learning and online refinement for
  spacecraft pose estimation across domain gap}.
\newblock \bibinfo{journal}{arXiv preprint arXiv:2203.04275} .
\bibitem[{Park and D’Amico(2023)}]{PARK2023}
\bibinfo{author}{Park, T.H.}, \bibinfo{author}{D’Amico, S.},
  \bibinfo{year}{2023}.
\newblock \bibinfo{title}{Robust multi-task learning and online refinement for
  spacecraft pose estimation across domain gap}.
\newblock \bibinfo{journal}{Advances in Space Research} \URLprefix
  \url{https://www.sciencedirect.com/science/article/pii/S0273117723002284},
  \DOIprefix\doi{https://doi.org/10.1016/j.asr.2023.03.036}.
\bibitem[{Park et~al.(2023a)Park, M{\"a}rtens, Jawaid, Wang, Chen, Chin, Izzo
  and D’Amico}]{park2023satellite}
\bibinfo{author}{Park, T.H.}, \bibinfo{author}{M{\"a}rtens, M.},
  \bibinfo{author}{Jawaid, M.}, \bibinfo{author}{Wang, Z.},
  \bibinfo{author}{Chen, B.}, \bibinfo{author}{Chin, T.J.},
  \bibinfo{author}{Izzo, D.}, \bibinfo{author}{D’Amico, S.},
  \bibinfo{year}{2023}a.
\newblock \bibinfo{title}{Satellite pose estimation competition 2021: Results
  and analyses}.
\newblock \bibinfo{journal}{Acta Astronautica} .
\bibitem[{Park et~al.(2023b)Park, Märtens, Jawaid, Wang, Chen, Chin, Izzo and
  D’Amico}]{park2021}
\bibinfo{author}{Park, T.H.}, \bibinfo{author}{Märtens, M.},
  \bibinfo{author}{Jawaid, M.}, \bibinfo{author}{Wang, Z.},
  \bibinfo{author}{Chen, B.}, \bibinfo{author}{Chin, T.J.},
  \bibinfo{author}{Izzo, D.}, \bibinfo{author}{D’Amico, S.},
  \bibinfo{year}{2023}b.
\newblock \bibinfo{title}{Satellite pose estimation competition 2021: Results
  and analyses}.
\newblock \bibinfo{journal}{Acta Astronautica} \bibinfo{volume}{204},
  \bibinfo{pages}{640--665}.
\newblock \URLprefix
  \url{https://www.sciencedirect.com/science/article/pii/S0094576523000048},
  \DOIprefix\doi{https://doi.org/10.1016/j.actaastro.2023.01.002}.
\bibitem[{Park et~al.(2019b)Park, Sharma and D'Amico}]{park2019towards}
\bibinfo{author}{Park, T.H.}, \bibinfo{author}{Sharma, S.},
  \bibinfo{author}{D'Amico, S.}, \bibinfo{year}{2019}b.
\newblock \bibinfo{title}{Towards robust learning-based pose estimation of
  noncooperative spacecraft}.
\newblock \bibinfo{journal}{arXiv preprint arXiv:1909.00392} .
\bibitem[{Park et~al.(2019c)Park, Sharma and D'Amico}]{park2019}
\bibinfo{author}{Park, T.H.}, \bibinfo{author}{Sharma, S.},
  \bibinfo{author}{D'Amico, S.}, \bibinfo{year}{2019}c.
\newblock \bibinfo{title}{Towards robust learning-based pose estimation of
  noncooperative spacecraft}.
\newblock \bibinfo{journal}{arXiv preprint arXiv:1909.00392} .
\bibitem[{Pauly et~al.(2022)Pauly, Jamrozik, Del~Castillo, Borgue, Singh,
  Makhdoomi, Christidi-Loumpasefski, Gaudilliere, Martinez, Rathinam
  et~al.}]{pauly2022lessons}
\bibinfo{author}{Pauly, L.}, \bibinfo{author}{Jamrozik, M.L.},
  \bibinfo{author}{Del~Castillo, M.O.}, \bibinfo{author}{Borgue, O.},
  \bibinfo{author}{Singh, I.P.}, \bibinfo{author}{Makhdoomi, M.R.},
  \bibinfo{author}{Christidi-Loumpasefski, O.O.}, \bibinfo{author}{Gaudilliere,
  V.}, \bibinfo{author}{Martinez, C.}, \bibinfo{author}{Rathinam, A.}, et~al.,
  \bibinfo{year}{2022}.
\newblock \bibinfo{title}{Lessons from a space lab--an image acquisition
  perspective}.
\newblock \bibinfo{journal}{arXiv preprint arXiv:2208.08865} .
\bibitem[{Pearl and Mackenzie(2018)}]{pearl2018book}
\bibinfo{author}{Pearl, J.}, \bibinfo{author}{Mackenzie, D.},
  \bibinfo{year}{2018}.
\newblock \bibinfo{title}{The book of why: the new science of cause and
  effect}.
\newblock \bibinfo{publisher}{Basic books}.
\bibitem[{Peng et~al.(2018)Peng, Tang, Yang, Feris and
  Metaxas}]{peng2018jointly}
\bibinfo{author}{Peng, X.}, \bibinfo{author}{Tang, Z.}, \bibinfo{author}{Yang,
  F.}, \bibinfo{author}{Feris, R.S.}, \bibinfo{author}{Metaxas, D.},
  \bibinfo{year}{2018}.
\newblock \bibinfo{title}{Jointly optimize data augmentation and network
  training: Adversarial data augmentation in human pose estimation}, in:
  \bibinfo{booktitle}{Proceedings of the IEEE conference on computer vision and
  pattern recognition}, pp. \bibinfo{pages}{2226--2234}.
\bibitem[{Phisannupawong et~al.(2020)Phisannupawong, Kamsing, Torteeka,
  Channumsin, Sawangwit, Hematulin, Jarawan, Somjit, Yooyen, Delahaye
  et~al.}]{phisannupawong2020vision}
\bibinfo{author}{Phisannupawong, T.}, \bibinfo{author}{Kamsing, P.},
  \bibinfo{author}{Torteeka, P.}, \bibinfo{author}{Channumsin, S.},
  \bibinfo{author}{Sawangwit, U.}, \bibinfo{author}{Hematulin, W.},
  \bibinfo{author}{Jarawan, T.}, \bibinfo{author}{Somjit, T.},
  \bibinfo{author}{Yooyen, S.}, \bibinfo{author}{Delahaye, D.}, et~al.,
  \bibinfo{year}{2020}.
\newblock \bibinfo{title}{Vision-based spacecraft pose estimation via a deep
  convolutional neural network for noncooperative docking operations}.
\newblock \bibinfo{journal}{Aerospace} \bibinfo{volume}{7},
  \bibinfo{pages}{126}.
\bibitem[{Piazza et~al.(2021)Piazza, Maestrini, Di~Lizia
  et~al.}]{piazza2021deep}
\bibinfo{author}{Piazza, M.}, \bibinfo{author}{Maestrini, M.},
  \bibinfo{author}{Di~Lizia, P.}, et~al., \bibinfo{year}{2021}.
\newblock \bibinfo{title}{Deep learning-based monocular relative pose
  estimation of uncooperative spacecraft}, in: \bibinfo{booktitle}{8th European
  Conference on Space Debris, ESA/ESOC}, \bibinfo{organization}{ESA}. pp.
  \bibinfo{pages}{1--13}.
\bibitem[{Posso et~al.(2022)Posso, Bois and Savaria}]{posso2022mobileurso}
\bibinfo{author}{Posso, J.}, \bibinfo{author}{Bois, G.},
  \bibinfo{author}{Savaria, Y.}, \bibinfo{year}{2022}.
\newblock \bibinfo{title}{Mobile-ursonet: an embeddable neural network for
  onboard spacecraft pose estimation}.
\newblock \bibinfo{journal}{arXiv preprint arXiv:2205.02065} .
\bibitem[{Powell et~al.(2018)Powell, Campola, Sheets, Davidson and
  Welsh}]{powell2018commercial}
\bibinfo{author}{Powell, W.}, \bibinfo{author}{Campola, M.},
  \bibinfo{author}{Sheets, T.}, \bibinfo{author}{Davidson, A.},
  \bibinfo{author}{Welsh, S.}, \bibinfo{year}{2018}.
\newblock \bibinfo{title}{Commercial Off-The-Shelf GPU Qualification for Space
  Applications}.
\newblock \bibinfo{type}{Technical Report}.
\bibitem[{Price and Yoshida(2021)}]{price2021monocular}
\bibinfo{author}{Price, A.}, \bibinfo{author}{Yoshida, K.},
  \bibinfo{year}{2021}.
\newblock \bibinfo{title}{A monocular pose estimation case study: The hayabusa2
  minerva-ii2 deployment}, in: \bibinfo{booktitle}{Proceedings of the IEEE/CVF
  Conference on Computer Vision and Pattern Recognition}, pp.
  \bibinfo{pages}{1992--2001}.
\bibitem[{Proen{\c{c}}a and Gao(2020)}]{proenca2020}
\bibinfo{author}{Proen{\c{c}}a, P.F.}, \bibinfo{author}{Gao, Y.},
  \bibinfo{year}{2020}.
\newblock \bibinfo{title}{Deep learning for spacecraft pose estimation from
  photorealistic rendering}, in: \bibinfo{booktitle}{2020 IEEE International
  Conference on Robotics and Automation (ICRA)}, \bibinfo{organization}{IEEE}.
  pp. \bibinfo{pages}{6007--6013}.
\bibitem[{Quigley et~al.(2009)Quigley, Conley, Gerkey, Faust, Foote, Leibs,
  Wheeler, Ng et~al.}]{quigley2009ros}
\bibinfo{author}{Quigley, M.}, \bibinfo{author}{Conley, K.},
  \bibinfo{author}{Gerkey, B.}, \bibinfo{author}{Faust, J.},
  \bibinfo{author}{Foote, T.}, \bibinfo{author}{Leibs, J.},
  \bibinfo{author}{Wheeler, R.}, \bibinfo{author}{Ng, A.Y.}, et~al.,
  \bibinfo{year}{2009}.
\newblock \bibinfo{title}{Ros: an open-source robot operating system}, in:
  \bibinfo{booktitle}{ICRA workshop on open source software},
  \bibinfo{organization}{Kobe, Japan}. p.~\bibinfo{pages}{5}.
\bibitem[{Rathinam and Gao(2020)}]{rathinam2020}
\bibinfo{author}{Rathinam, A.}, \bibinfo{author}{Gao, Y.},
  \bibinfo{year}{2020}.
\newblock \bibinfo{title}{On-orbit relative navigation near a known target
  using monocular vision and convolutional neural networks for pose
  estimation}, in: \bibinfo{booktitle}{International Symposium on Artificial
  Intelligence, Robotics and Automation in Space (iSAIRAS), Virutal Conference
  (Pasadena, CA:)}, pp. \bibinfo{pages}{1--6}.
\bibitem[{Rathinam et~al.(2022)Rathinam, Gaudilliere, Mohamed~Ali, Ortiz
  Del~Castillo, Pauly and Aouada}]{rathinam_arunkumar_2022_6599762}
\bibinfo{author}{Rathinam, A.}, \bibinfo{author}{Gaudilliere, V.},
  \bibinfo{author}{Mohamed~Ali, M.A.}, \bibinfo{author}{Ortiz Del~Castillo,
  M.}, \bibinfo{author}{Pauly, L.}, \bibinfo{author}{Aouada, D.},
  \bibinfo{year}{2022}.
\newblock \bibinfo{title}{{SPARK 2022 Dataset : Spacecraft Detection and
  Trajectory Estimation}}.
\newblock \URLprefix \url{https://doi.org/10.5281/zenodo.6599762},
  \DOIprefix\doi{10.5281/zenodo.6599762}.
\bibitem[{Rathinam et~al.(2021)Rathinam, Hao and Gao}]{arunkumar2021}
\bibinfo{author}{Rathinam, A.}, \bibinfo{author}{Hao, Z.},
  \bibinfo{author}{Gao, Y.}, \bibinfo{year}{2021}.
\newblock \bibinfo{title}{Autonomous visual navigation for spacecraft on-orbit
  operations}, in: \bibinfo{booktitle}{Space {Robotics} and {Autonomous}
  {Systems}: {Technologies}, advances and applications}.
  \bibinfo{publisher}{Institution of Engineering and Technology}, pp.
  \bibinfo{pages}{125--157}.
\newblock \URLprefix \url{https://doi.org/10.1049/PBCE131E_ch5}.
\bibitem[{Redd(2020)}]{redd2020}
\bibinfo{author}{Redd, N.T.}, \bibinfo{year}{2020}.
\newblock \bibinfo{title}{Bringing satellites back from the dead: Mission
  extension vehicles give defunct spacecraft a new lease on life - [news]}.
\newblock \bibinfo{journal}{IEEE Spectrum} \bibinfo{volume}{57},
  \bibinfo{pages}{6--7}.
\newblock \DOIprefix\doi{10.1109/MSPEC.2020.9150540}.
\bibitem[{Redmon et~al.(2016)Redmon, Divvala, Girshick and
  Farhadi}]{redmon2016you}
\bibinfo{author}{Redmon, J.}, \bibinfo{author}{Divvala, S.},
  \bibinfo{author}{Girshick, R.}, \bibinfo{author}{Farhadi, A.},
  \bibinfo{year}{2016}.
\newblock \bibinfo{title}{You only look once: Unified, real-time object
  detection}, in: \bibinfo{booktitle}{Proceedings of the IEEE conference on
  computer vision and pattern recognition}, pp. \bibinfo{pages}{779--788}.
\bibitem[{Redmon and Farhadi(2017)}]{redmon2017yolo9000}
\bibinfo{author}{Redmon, J.}, \bibinfo{author}{Farhadi, A.},
  \bibinfo{year}{2017}.
\newblock \bibinfo{title}{Yolo9000: better, faster, stronger}, in:
  \bibinfo{booktitle}{Proceedings of the IEEE conference on computer vision and
  pattern recognition}, pp. \bibinfo{pages}{7263--7271}.
\bibitem[{Redmon and Farhadi(2018)}]{redmon2018yolov3}
\bibinfo{author}{Redmon, J.}, \bibinfo{author}{Farhadi, A.},
  \bibinfo{year}{2018}.
\newblock \bibinfo{title}{Yolov3: An incremental improvement}.
\newblock \bibinfo{journal}{arXiv preprint arXiv:1804.02767} .
\bibitem[{Ren et~al.(2015)Ren, He, Girshick and Sun}]{ren2015faster}
\bibinfo{author}{Ren, S.}, \bibinfo{author}{He, K.}, \bibinfo{author}{Girshick,
  R.}, \bibinfo{author}{Sun, J.}, \bibinfo{year}{2015}.
\newblock \bibinfo{title}{Faster r-cnn: Towards real-time object detection with
  region proposal networks}.
\newblock \bibinfo{journal}{Advances in neural information processing systems}
  \bibinfo{volume}{28}.
\bibitem[{Rennie et~al.(2016)Rennie, Shome, Bekris and
  Souza}]{DBLP:journals/ral/RennieSBS16}
\bibinfo{author}{Rennie, C.}, \bibinfo{author}{Shome, R.},
  \bibinfo{author}{Bekris, K.E.}, \bibinfo{author}{Souza, A.F.D.},
  \bibinfo{year}{2016}.
\newblock \bibinfo{title}{A dataset for improved rgbd-based object detection
  and pose estimation for warehouse pick-and-place}.
\newblock \bibinfo{journal}{{IEEE} Robotics Autom. Lett.} \bibinfo{volume}{1},
  \bibinfo{pages}{1179--1185}.
\newblock \URLprefix \url{https://doi.org/10.1109/LRA.2016.2532924},
  \DOIprefix\doi{10.1109/LRA.2016.2532924}.
\bibitem[{Rondao and Aouf(2018)}]{rondao2018multi}
\bibinfo{author}{Rondao, D.}, \bibinfo{author}{Aouf, N.}, \bibinfo{year}{2018}.
\newblock \bibinfo{title}{Multi-view monocular pose estimation for spacecraft
  relative navigation}, in: \bibinfo{booktitle}{2018 AIAA Guidance, Navigation,
  and Control Conference}, p. \bibinfo{pages}{2100}.
\bibitem[{Rondao et~al.(2021)Rondao, Aouf and Richardson}]{rondao2021}
\bibinfo{author}{Rondao, D.}, \bibinfo{author}{Aouf, N.},
  \bibinfo{author}{Richardson, M.A.}, \bibinfo{year}{2021}.
\newblock \bibinfo{title}{Chinet: Deep recurrent convolutional learning for
  multimodal spacecraft pose estimation}.
\newblock \bibinfo{journal}{CoRR} \bibinfo{volume}{abs/2108.10282}.
\newblock \URLprefix \url{https://arxiv.org/abs/2108.10282},
  \href{http://arxiv.org/abs/2108.10282}{\tt arXiv:2108.10282}.
\bibitem[{Rondao et~al.(2022)Rondao, Aouf and Richardson}]{rondao2022chinet}
\bibinfo{author}{Rondao, D.}, \bibinfo{author}{Aouf, N.},
  \bibinfo{author}{Richardson, M.A.}, \bibinfo{year}{2022}.
\newblock \bibinfo{title}{Chinet: Deep recurrent convolutional learning for
  multimodal spacecraft pose estimation}.
\newblock \bibinfo{journal}{IEEE Transactions on Aerospace and Electronic
  Systems} .
\bibitem[{Ronneberger et~al.(2015)Ronneberger, Fischer and
  Brox}]{ronneberger2015u}
\bibinfo{author}{Ronneberger, O.}, \bibinfo{author}{Fischer, P.},
  \bibinfo{author}{Brox, T.}, \bibinfo{year}{2015}.
\newblock \bibinfo{title}{U-net: Convolutional networks for biomedical image
  segmentation}, in: \bibinfo{booktitle}{International Conference on Medical
  image computing and computer-assisted intervention},
  \bibinfo{organization}{Springer}. pp. \bibinfo{pages}{234--241}.
\bibitem[{Ruder(2017)}]{ruder2017overview}
\bibinfo{author}{Ruder, S.}, \bibinfo{year}{2017}.
\newblock \bibinfo{title}{An overview of multi-task learning in deep neural
  networks}.
\newblock \bibinfo{journal}{arXiv preprint arXiv:1706.05098} .
\bibitem[{Russakovsky et~al.(2015)Russakovsky, Deng, Su, Krause, Satheesh, Ma,
  Huang, Karpathy, Khosla, Bernstein, Berg and
  Fei{-}Fei}]{DBLP:journals/ijcv/RussakovskyDSKS15}
\bibinfo{author}{Russakovsky, O.}, \bibinfo{author}{Deng, J.},
  \bibinfo{author}{Su, H.}, \bibinfo{author}{Krause, J.},
  \bibinfo{author}{Satheesh, S.}, \bibinfo{author}{Ma, S.},
  \bibinfo{author}{Huang, Z.}, \bibinfo{author}{Karpathy, A.},
  \bibinfo{author}{Khosla, A.}, \bibinfo{author}{Bernstein, M.S.},
  \bibinfo{author}{Berg, A.C.}, \bibinfo{author}{Fei{-}Fei, L.},
  \bibinfo{year}{2015}.
\newblock \bibinfo{title}{Imagenet large scale visual recognition challenge}.
\newblock \bibinfo{journal}{Int. J. Comput. Vis.} \bibinfo{volume}{115},
  \bibinfo{pages}{211--252}.
\newblock \URLprefix \url{https://doi.org/10.1007/s11263-015-0816-y},
  \DOIprefix\doi{10.1007/s11263-015-0816-y}.
\bibitem[{Sabatini et~al.(2015)Sabatini, Palmerini and
  Gasbarri}]{SABATINI2015184}
\bibinfo{author}{Sabatini, M.}, \bibinfo{author}{Palmerini, G.B.},
  \bibinfo{author}{Gasbarri, P.}, \bibinfo{year}{2015}.
\newblock \bibinfo{title}{A testbed for visual based navigation and control
  during space rendezvous operations}.
\newblock \bibinfo{journal}{Acta Astronautica} \bibinfo{volume}{117},
  \bibinfo{pages}{184--196}.
\newblock \URLprefix
  \url{https://www.sciencedirect.com/science/article/pii/S0094576515003070},
  \DOIprefix\doi{https://doi.org/10.1016/j.actaastro.2015.07.026}.
\bibitem[{Sandler et~al.(2018)Sandler, Howard, Zhu, Zhmoginov and
  Chen}]{sandler2018mobilenetv2}
\bibinfo{author}{Sandler, M.}, \bibinfo{author}{Howard, A.},
  \bibinfo{author}{Zhu, M.}, \bibinfo{author}{Zhmoginov, A.},
  \bibinfo{author}{Chen, L.C.}, \bibinfo{year}{2018}.
\newblock \bibinfo{title}{Mobilenetv2: Inverted residuals and linear
  bottlenecks}, in: \bibinfo{booktitle}{Proceedings of the IEEE conference on
  computer vision and pattern recognition}, pp. \bibinfo{pages}{4510--4520}.
\bibitem[{Shafer and Vovk(2008)}]{shafer2008tutorial}
\bibinfo{author}{Shafer, G.}, \bibinfo{author}{Vovk, V.}, \bibinfo{year}{2008}.
\newblock \bibinfo{title}{A tutorial on conformal prediction.}
\newblock \bibinfo{journal}{Journal of Machine Learning Research}
  \bibinfo{volume}{9}.
\bibitem[{Shannon(1948)}]{shannon1948mathematical}
\bibinfo{author}{Shannon, C.E.}, \bibinfo{year}{1948}.
\newblock \bibinfo{title}{A mathematical theory of communication}.
\newblock \bibinfo{journal}{The Bell system technical journal}
  \bibinfo{volume}{27}, \bibinfo{pages}{379--423}.
\bibitem[{Sharma et~al.(2018a)Sharma, Beierle and D'Amico}]{sharma2018pose}
\bibinfo{author}{Sharma, S.}, \bibinfo{author}{Beierle, C.},
  \bibinfo{author}{D'Amico, S.}, \bibinfo{year}{2018}a.
\newblock \bibinfo{title}{Pose estimation for non-cooperative spacecraft
  rendezvous using convolutional neural networks}, in: \bibinfo{booktitle}{2018
  IEEE Aerospace Conference}, \bibinfo{organization}{IEEE}. pp.
  \bibinfo{pages}{1--12}.
\bibitem[{Sharma and D'Amico(2019)}]{sharma2019pose}
\bibinfo{author}{Sharma, S.}, \bibinfo{author}{D'Amico, S.},
  \bibinfo{year}{2019}.
\newblock \bibinfo{title}{Pose estimation for non-cooperative rendezvous using
  neural networks}.
\newblock \bibinfo{journal}{arXiv preprint arXiv:1906.09868} .
\bibitem[{Sharma et~al.(2018b)Sharma, Ventura and D’Amico}]{sharma2018robust}
\bibinfo{author}{Sharma, S.}, \bibinfo{author}{Ventura, J.},
  \bibinfo{author}{D’Amico, S.}, \bibinfo{year}{2018}b.
\newblock \bibinfo{title}{Robust model-based monocular pose initialization for
  noncooperative spacecraft rendezvous}.
\newblock \bibinfo{journal}{Journal of Spacecraft and Rockets}
  \bibinfo{volume}{55}, \bibinfo{pages}{1414--1429}.
\bibitem[{Shi et~al.(2016)Shi, Ulrich and Ruel}]{shi2016spacecraft}
\bibinfo{author}{Shi, J.}, \bibinfo{author}{Ulrich, S.}, \bibinfo{author}{Ruel,
  S.}, \bibinfo{year}{2016}.
\newblock \bibinfo{title}{Spacecraft pose estimation using a monocular camera},
  in: \bibinfo{booktitle}{67th International Astronautical Congress},
  \bibinfo{organization}{Guadalajara}.
\bibitem[{Shreiner et~al.(2009)Shreiner, Group et~al.}]{shreiner2009opengl}
\bibinfo{author}{Shreiner, D.}, \bibinfo{author}{Group, B.T.K.O.A.W.}, et~al.,
  \bibinfo{year}{2009}.
\newblock \bibinfo{title}{OpenGL programming guide: the official guide to
  learning OpenGL, versions 3.0 and 3.1}.
\newblock \bibinfo{publisher}{Pearson Education}.
\bibitem[{Shui et~al.(2019)Shui, Abbasi, Robitaille, Wang and
  Gagn{\'e}}]{shui2019principled}
\bibinfo{author}{Shui, C.}, \bibinfo{author}{Abbasi, M.},
  \bibinfo{author}{Robitaille, L.{\'E}.}, \bibinfo{author}{Wang, B.},
  \bibinfo{author}{Gagn{\'e}, C.}, \bibinfo{year}{2019}.
\newblock \bibinfo{title}{A principled approach for learning task similarity in
  multitask learning}.
\newblock \bibinfo{journal}{arXiv preprint arXiv:1903.09109} .
\bibitem[{Song et~al.(2022a)Song, Rondao and Aouf}]{song2022deep}
\bibinfo{author}{Song, J.}, \bibinfo{author}{Rondao, D.},
  \bibinfo{author}{Aouf, N.}, \bibinfo{year}{2022}a.
\newblock \bibinfo{title}{Deep learning-based spacecraft relative navigation
  methods: A survey}.
\newblock \bibinfo{journal}{Acta Astronautica} \bibinfo{volume}{191},
  \bibinfo{pages}{22--40}.
\bibitem[{Song et~al.(2022b)Song, Wang, Mondal and
  Sahoo}]{song2022comprehensive}
\bibinfo{author}{Song, Y.}, \bibinfo{author}{Wang, T.},
  \bibinfo{author}{Mondal, S.K.}, \bibinfo{author}{Sahoo, J.P.},
  \bibinfo{year}{2022}b.
\newblock \bibinfo{title}{A comprehensive survey of few-shot learning:
  Evolution, applications, challenges, and opportunities}.
\newblock \bibinfo{journal}{arXiv preprint arXiv:2205.06743} .
\bibitem[{Strutz(2011)}]{strutz2011data}
\bibinfo{author}{Strutz, T.}, \bibinfo{year}{2011}.
\newblock \bibinfo{title}{Data fitting and uncertainty: A practical
  introduction to weighted least squares and beyond}.
\newblock \bibinfo{publisher}{Springer}.
\bibitem[{Sun et~al.(2019a)Sun, Xiao, Liu and Wang}]{sun2019deep}
\bibinfo{author}{Sun, K.}, \bibinfo{author}{Xiao, B.}, \bibinfo{author}{Liu,
  D.}, \bibinfo{author}{Wang, J.}, \bibinfo{year}{2019}a.
\newblock \bibinfo{title}{Deep high-resolution representation learning for
  human pose estimation}, in: \bibinfo{booktitle}{Proceedings of the IEEE/CVF
  conference on computer vision and pattern recognition}, pp.
  \bibinfo{pages}{5693--5703}.
\bibitem[{Sun et~al.(2019b)Sun, Zhao, Jiang, Cheng, Xiao, Liu, Mu, Wang, Liu
  and Wang}]{sun2019}
\bibinfo{author}{Sun, K.}, \bibinfo{author}{Zhao, Y.}, \bibinfo{author}{Jiang,
  B.}, \bibinfo{author}{Cheng, T.}, \bibinfo{author}{Xiao, B.},
  \bibinfo{author}{Liu, D.}, \bibinfo{author}{Mu, Y.}, \bibinfo{author}{Wang,
  X.}, \bibinfo{author}{Liu, W.}, \bibinfo{author}{Wang, J.},
  \bibinfo{year}{2019}b.
\newblock \bibinfo{title}{High-resolution representations for labeling pixels
  and regions}.
\newblock \bibinfo{journal}{arXiv preprint arXiv:1904.04514} .
\bibitem[{Sweden()}]{PRISMA_OHB}
\bibinfo{author}{Sweden, O.}, .
\newblock \bibinfo{title}{Prisma}.
\newblock
  \bibinfo{howpublished}{\url{https://www.ohb-sweden.se/space-missions/prisma}}.
\newblock \bibinfo{note}{Accessed: April 5, 2023}.
\bibitem[{Szegedy et~al.(2015)Szegedy, Liu, Jia, Sermanet, Reed, Anguelov,
  Erhan, Vanhoucke and Rabinovich}]{szegedy2015going}
\bibinfo{author}{Szegedy, C.}, \bibinfo{author}{Liu, W.}, \bibinfo{author}{Jia,
  Y.}, \bibinfo{author}{Sermanet, P.}, \bibinfo{author}{Reed, S.},
  \bibinfo{author}{Anguelov, D.}, \bibinfo{author}{Erhan, D.},
  \bibinfo{author}{Vanhoucke, V.}, \bibinfo{author}{Rabinovich, A.},
  \bibinfo{year}{2015}.
\newblock \bibinfo{title}{Going deeper with convolutions}, in:
  \bibinfo{booktitle}{Proceedings of the IEEE conference on computer vision and
  pattern recognition}, pp. \bibinfo{pages}{1--9}.
\bibitem[{Szeliski(2022)}]{szeliski2022computer}
\bibinfo{author}{Szeliski, R.}, \bibinfo{year}{2022}.
\newblock \bibinfo{title}{Computer vision: algorithms and applications}.
\newblock \bibinfo{publisher}{Springer Nature}.
\bibitem[{Tan and Le(2019)}]{tan2019efficientnet}
\bibinfo{author}{Tan, M.}, \bibinfo{author}{Le, Q.}, \bibinfo{year}{2019}.
\newblock \bibinfo{title}{Efficientnet: Rethinking model scaling for
  convolutional neural networks}, in: \bibinfo{booktitle}{International
  conference on machine learning}, \bibinfo{organization}{PMLR}. pp.
  \bibinfo{pages}{6105--6114}.
\bibitem[{Tan et~al.(2020)Tan, Pang and Le}]{tan2020efficientdet}
\bibinfo{author}{Tan, M.}, \bibinfo{author}{Pang, R.}, \bibinfo{author}{Le,
  Q.V.}, \bibinfo{year}{2020}.
\newblock \bibinfo{title}{Efficientdet: Scalable and efficient object
  detection}, in: \bibinfo{booktitle}{Proceedings of the IEEE/CVF conference on
  computer vision and pattern recognition}, pp. \bibinfo{pages}{10781--10790}.
\bibitem[{Tensorflow()}]{efficientnetlite}
\bibinfo{author}{Tensorflow}, .
\newblock \bibinfo{title}{Tpu/models/official/efficientnet/lite at master ·
  tensorflow/tpu}.
\newblock \URLprefix
  \url{https://github.com/tensorflow/tpu/tree/master/models/official/efficientnet/lite}.
\bibitem[{Tibshirani et~al.(2019)Tibshirani, Foygel~Barber, Candes and
  Ramdas}]{NEURIPS2019_8fb21ee7}
\bibinfo{author}{Tibshirani, R.J.}, \bibinfo{author}{Foygel~Barber, R.},
  \bibinfo{author}{Candes, E.}, \bibinfo{author}{Ramdas, A.},
  \bibinfo{year}{2019}.
\newblock \bibinfo{title}{Conformal prediction under covariate shift}, in:
  \bibinfo{editor}{Wallach, H.}, \bibinfo{editor}{Larochelle, H.},
  \bibinfo{editor}{Beygelzimer, A.}, \bibinfo{editor}{d\textquotesingle
  Alch\'{e}-Buc, F.}, \bibinfo{editor}{Fox, E.}, \bibinfo{editor}{Garnett, R.}
  (Eds.), \bibinfo{booktitle}{Advances in Neural Information Processing
  Systems}, \bibinfo{publisher}{Curran Associates, Inc.}
\newblock \URLprefix
  \url{https://proceedings.neurips.cc/paper/2019/file/8fb21ee7a2207526da55a679f0332de2-Paper.pdf}.
\bibitem[{Tobin et~al.(2017a)Tobin, Fong, Ray, Schneider, Zaremba and
  Abbeel}]{Tobin2017DomainRF}
\bibinfo{author}{Tobin, J.}, \bibinfo{author}{Fong, R.}, \bibinfo{author}{Ray,
  A.}, \bibinfo{author}{Schneider, J.}, \bibinfo{author}{Zaremba, W.},
  \bibinfo{author}{Abbeel, P.}, \bibinfo{year}{2017}a.
\newblock \bibinfo{title}{Domain randomization for transferring deep neural
  networks from simulation to the real world}.
\newblock \bibinfo{journal}{2017 IEEE/RSJ International Conference on
  Intelligent Robots and Systems (IROS)} , \bibinfo{pages}{23--30}.
\bibitem[{Tobin et~al.(2017b)Tobin, Fong, Ray, Schneider, Zaremba and
  Abbeel}]{tobin2017domain}
\bibinfo{author}{Tobin, J.}, \bibinfo{author}{Fong, R.}, \bibinfo{author}{Ray,
  A.}, \bibinfo{author}{Schneider, J.}, \bibinfo{author}{Zaremba, W.},
  \bibinfo{author}{Abbeel, P.}, \bibinfo{year}{2017}b.
\newblock \bibinfo{title}{Domain randomization for transferring deep neural
  networks from simulation to the real world}, in: \bibinfo{booktitle}{2017
  IEEE/RSJ international conference on intelligent robots and systems (IROS)},
  \bibinfo{organization}{IEEE}. pp. \bibinfo{pages}{23--30}.
\bibitem[{Toft et~al.(2020)Toft, Maddern, Torii, Hammarstrand, Stenborg,
  Safari, Okutomi, Pollefeys, Sivic, Pajdla et~al.}]{toft2020long}
\bibinfo{author}{Toft, C.}, \bibinfo{author}{Maddern, W.},
  \bibinfo{author}{Torii, A.}, \bibinfo{author}{Hammarstrand, L.},
  \bibinfo{author}{Stenborg, E.}, \bibinfo{author}{Safari, D.},
  \bibinfo{author}{Okutomi, M.}, \bibinfo{author}{Pollefeys, M.},
  \bibinfo{author}{Sivic, J.}, \bibinfo{author}{Pajdla, T.}, et~al.,
  \bibinfo{year}{2020}.
\newblock \bibinfo{title}{Long-term visual localization revisited}.
\newblock \bibinfo{journal}{IEEE Transactions on Pattern Analysis and Machine
  Intelligence} .
\bibitem[{Voulodimos et~al.(2018)Voulodimos, Doulamis, Doulamis and
  Protopapadakis}]{voulodimos2018deep}
\bibinfo{author}{Voulodimos, A.}, \bibinfo{author}{Doulamis, N.},
  \bibinfo{author}{Doulamis, A.}, \bibinfo{author}{Protopapadakis, E.},
  \bibinfo{year}{2018}.
\newblock \bibinfo{title}{Deep learning for computer vision: A brief review}.
\newblock \bibinfo{journal}{Computational intelligence and neuroscience}
  \bibinfo{volume}{2018}.
\bibitem[{Wang and Yeung(2020)}]{wang2020survey}
\bibinfo{author}{Wang, H.}, \bibinfo{author}{Yeung, D.Y.},
  \bibinfo{year}{2020}.
\newblock \bibinfo{title}{A survey on bayesian deep learning}.
\newblock \bibinfo{journal}{ACM computing surveys (csur)} \bibinfo{volume}{53},
  \bibinfo{pages}{1--37}.
\bibitem[{Wang et~al.(2022a)Wang, Lan, Liu, Ouyang, Qin, Lu, Chen, Zeng and
  Yu}]{wang2022generalizing}
\bibinfo{author}{Wang, J.}, \bibinfo{author}{Lan, C.}, \bibinfo{author}{Liu,
  C.}, \bibinfo{author}{Ouyang, Y.}, \bibinfo{author}{Qin, T.},
  \bibinfo{author}{Lu, W.}, \bibinfo{author}{Chen, Y.}, \bibinfo{author}{Zeng,
  W.}, \bibinfo{author}{Yu, P.}, \bibinfo{year}{2022}a.
\newblock \bibinfo{title}{Generalizing to unseen domains: A survey on domain
  generalization}.
\newblock \bibinfo{journal}{IEEE Transactions on Knowledge and Data
  Engineering} .
\bibitem[{Wang et~al.(2020)Wang, Sun, Cheng, Jiang, Deng, Zhao, Liu, Mu, Tan,
  Wang et~al.}]{wang2020deep}
\bibinfo{author}{Wang, J.}, \bibinfo{author}{Sun, K.}, \bibinfo{author}{Cheng,
  T.}, \bibinfo{author}{Jiang, B.}, \bibinfo{author}{Deng, C.},
  \bibinfo{author}{Zhao, Y.}, \bibinfo{author}{Liu, D.}, \bibinfo{author}{Mu,
  Y.}, \bibinfo{author}{Tan, M.}, \bibinfo{author}{Wang, X.}, et~al.,
  \bibinfo{year}{2020}.
\newblock \bibinfo{title}{Deep high-resolution representation learning for
  visual recognition}.
\newblock \bibinfo{journal}{IEEE transactions on pattern analysis and machine
  intelligence} \bibinfo{volume}{43}, \bibinfo{pages}{3349--3364}.
\bibitem[{Wang and Deng(2018)}]{wang2018deep}
\bibinfo{author}{Wang, M.}, \bibinfo{author}{Deng, W.}, \bibinfo{year}{2018}.
\newblock \bibinfo{title}{Deep visual domain adaptation: A survey}.
\newblock \bibinfo{journal}{Neurocomputing} \bibinfo{volume}{312},
  \bibinfo{pages}{135--153}.
\bibitem[{Wang et~al.(2022b)Wang, Ma, Zhao and Tian}]{wang2022comprehensive}
\bibinfo{author}{Wang, Q.}, \bibinfo{author}{Ma, Y.}, \bibinfo{author}{Zhao,
  K.}, \bibinfo{author}{Tian, Y.}, \bibinfo{year}{2022}b.
\newblock \bibinfo{title}{A comprehensive survey of loss functions in machine
  learning}.
\newblock \bibinfo{journal}{Annals of Data Science} \bibinfo{volume}{9},
  \bibinfo{pages}{187--212}.
\bibitem[{{Wang} et~al.(2022){Wang}, {Wang}, {Jiao}, {Yang}, {Su}, {Zhai},
  {Chen} and {Zhang}}]{CA-SpaceNet2022}
\bibinfo{author}{{Wang}, S.}, \bibinfo{author}{{Wang}, S.},
  \bibinfo{author}{{Jiao}, B.}, \bibinfo{author}{{Yang}, D.},
  \bibinfo{author}{{Su}, L.}, \bibinfo{author}{{Zhai}, P.},
  \bibinfo{author}{{Chen}, C.}, \bibinfo{author}{{Zhang}, L.},
  \bibinfo{year}{2022}.
\newblock \bibinfo{title}{{CA-SpaceNet: Counterfactual Analysis for 6D Pose
  Estimation in Space}}.
\newblock \bibinfo{journal}{arXiv e-prints} ,
  \bibinfo{pages}{arXiv:2207.07869}\href{http://arxiv.org/abs/2207.07869}{\tt
  arXiv:2207.07869}.
\bibitem[{Wang et~al.(2019)Wang, Yang, Wang, Wang and Li}]{wang2019development}
\bibinfo{author}{Wang, W.}, \bibinfo{author}{Yang, Y.}, \bibinfo{author}{Wang,
  X.}, \bibinfo{author}{Wang, W.}, \bibinfo{author}{Li, J.},
  \bibinfo{year}{2019}.
\newblock \bibinfo{title}{Development of convolutional neural network and its
  application in image classification: a survey}.
\newblock \bibinfo{journal}{Optical Engineering} \bibinfo{volume}{58},
  \bibinfo{pages}{040901--040901}.
\bibitem[{Wang et~al.(2022)Wang, Shen, Hu, Yuan, Crowley and
  Vaufreydaz}]{wang2022self}
\bibinfo{author}{Wang, Y.}, \bibinfo{author}{Shen, X.}, \bibinfo{author}{Hu,
  S.X.}, \bibinfo{author}{Yuan, Y.}, \bibinfo{author}{Crowley, J.L.},
  \bibinfo{author}{Vaufreydaz, D.}, \bibinfo{year}{2022}.
\newblock \bibinfo{title}{Self-supervised transformers for unsupervised object
  discovery using normalized cut}, in: \bibinfo{booktitle}{Proceedings of the
  IEEE/CVF Conference on Computer Vision and Pattern Recognition}, pp.
  \bibinfo{pages}{14543--14553}.
\bibitem[{Weiler and Cesa(2019)}]{NEURIPS2019_45d6637b}
\bibinfo{author}{Weiler, M.}, \bibinfo{author}{Cesa, G.}, \bibinfo{year}{2019}.
\newblock \bibinfo{title}{General e(2)-equivariant steerable cnns}, in:
  \bibinfo{editor}{Wallach, H.}, \bibinfo{editor}{Larochelle, H.},
  \bibinfo{editor}{Beygelzimer, A.}, \bibinfo{editor}{d\textquotesingle
  Alch\'{e}-Buc, F.}, \bibinfo{editor}{Fox, E.}, \bibinfo{editor}{Garnett, R.}
  (Eds.), \bibinfo{booktitle}{Advances in Neural Information Processing
  Systems}.
\newblock \URLprefix
  \url{https://proceedings.neurips.cc/paper_files/paper/2019/file/45d6637b718d0f24a237069fe41b0db4-Paper.pdf}.
\bibitem[{Wijayatunga et~al.(2023)Wijayatunga, Armellin, Holt, Pirovano and
  Lidtke}]{wijayatunga_design_2023}
\bibinfo{author}{Wijayatunga, M.C.}, \bibinfo{author}{Armellin, R.},
  \bibinfo{author}{Holt, H.}, \bibinfo{author}{Pirovano, L.},
  \bibinfo{author}{Lidtke, A.A.}, \bibinfo{year}{2023}.
\newblock \bibinfo{title}{Design and guidance of a multi-active debris removal
  mission}.
\newblock \bibinfo{journal}{Astrodynamics} \URLprefix
  \url{https://link.springer.com/10.1007/s42064-023-0159-3},
  \DOIprefix\doi{10.1007/s42064-023-0159-3}.
\bibitem[{Wistuba et~al.(2019)Wistuba, Rawat and Pedapati}]{wistuba2019survey}
\bibinfo{author}{Wistuba, M.}, \bibinfo{author}{Rawat, A.},
  \bibinfo{author}{Pedapati, T.}, \bibinfo{year}{2019}.
\newblock \bibinfo{title}{A survey on neural architecture search}.
\newblock \bibinfo{journal}{arXiv preprint arXiv:1905.01392} .
\bibitem[{Witze(2023)}]{witze_2023}
\bibinfo{author}{Witze, A.}, \bibinfo{year}{2023}.
\newblock \bibinfo{title}{2022 was a record year for space launches}.
\newblock \URLprefix \url{https://www.nature.com/articles/d41586-023-00048-7}.
\bibitem[{Xiang et~al.(2018)Xiang, Schmidt, Narayanan and
  Fox}]{DBLP:conf/rss/XiangSNF18}
\bibinfo{author}{Xiang, Y.}, \bibinfo{author}{Schmidt, T.},
  \bibinfo{author}{Narayanan, V.}, \bibinfo{author}{Fox, D.},
  \bibinfo{year}{2018}.
\newblock \bibinfo{title}{Posecnn: {A} convolutional neural network for 6d
  object pose estimation in cluttered scenes}, in:
  \bibinfo{editor}{Kress{-}Gazit, H.}, \bibinfo{editor}{Srinivasa, S.S.},
  \bibinfo{editor}{Howard, T.}, \bibinfo{editor}{Atanasov, N.} (Eds.),
  \bibinfo{booktitle}{Robotics: Science and Systems XIV, Carnegie Mellon
  University, Pittsburgh, Pennsylvania, USA, June 26-30, 2018}.
\newblock \URLprefix \url{http://www.roboticsproceedings.org/rss14/p19.html},
  \DOIprefix\doi{10.15607/RSS.2018.XIV.019}.
\bibitem[{Xilinx(2022)}]{Xilinxfpga}
\bibinfo{author}{Xilinx}, \bibinfo{year}{2022}.
\newblock \bibinfo{title}{Product guide: Dpuczdx8g for zynq ultrascale+
  mpsocs}.
\newblock \URLprefix
  \url{https://www.xilinx.com/content/dam/xilinx/support/documents/ip_documentation/dpu/v4_0/pg338-dpu.pdf}.
\bibitem[{Xilinx(15 Jun 2022)}]{vitisaiuserguide}
\bibinfo{author}{Xilinx, A.}, \bibinfo{year}{15 Jun 2022}.
\newblock \bibinfo{title}{Vitis ai user guide}.
\newblock
  \bibinfo{howpublished}{\url{https://www.xilinx.com/content/dam/xilinx/support/documents/sw_manuals/vitis_ai/2_5/ug1414-vitis-ai.pdf}}.
\newblock \bibinfo{note}{[Online; accessed 30-Jan-2023]}.
\bibitem[{Xiong et~al.(2021)Xiong, Liu, Gupta, Akin, Bender, Wang, Kindermans,
  Tan, Singh and Chen}]{xiong2021mobiledets}
\bibinfo{author}{Xiong, Y.}, \bibinfo{author}{Liu, H.}, \bibinfo{author}{Gupta,
  S.}, \bibinfo{author}{Akin, B.}, \bibinfo{author}{Bender, G.},
  \bibinfo{author}{Wang, Y.}, \bibinfo{author}{Kindermans, P.J.},
  \bibinfo{author}{Tan, M.}, \bibinfo{author}{Singh, V.},
  \bibinfo{author}{Chen, B.}, \bibinfo{year}{2021}.
\newblock \bibinfo{title}{Mobiledets: Searching for object detection
  architectures for mobile accelerators}, in: \bibinfo{booktitle}{Proceedings
  of the IEEE/CVF Conference on Computer Vision and Pattern Recognition}, pp.
  \bibinfo{pages}{3825--3834}.
\bibitem[{Xu et~al.(2019)Xu, Uszkoreit, Du, Fan, Zhao and
  Zhu}]{xu2019explainable}
\bibinfo{author}{Xu, F.}, \bibinfo{author}{Uszkoreit, H.}, \bibinfo{author}{Du,
  Y.}, \bibinfo{author}{Fan, W.}, \bibinfo{author}{Zhao, D.},
  \bibinfo{author}{Zhu, J.}, \bibinfo{year}{2019}.
\newblock \bibinfo{title}{Explainable ai: A brief survey on history, research
  areas, approaches and challenges}, in: \bibinfo{booktitle}{Natural Language
  Processing and Chinese Computing: 8th CCF International Conference, NLPCC
  2019, Dunhuang, China, October 9--14, 2019, Proceedings, Part II 8},
  \bibinfo{organization}{Springer}. pp. \bibinfo{pages}{563--574}.
\bibitem[{Yuheng and Hao(2017)}]{yuheng2017image}
\bibinfo{author}{Yuheng, S.}, \bibinfo{author}{Hao, Y.}, \bibinfo{year}{2017}.
\newblock \bibinfo{title}{Image segmentation algorithms overview}.
\newblock \bibinfo{journal}{arXiv preprint arXiv:1707.02051} .
\bibitem[{Zaidi et~al.(2022)Zaidi, Ansari, Aslam, Kanwal, Asghar and
  Lee}]{zaidi2022survey}
\bibinfo{author}{Zaidi, S.S.A.}, \bibinfo{author}{Ansari, M.S.},
  \bibinfo{author}{Aslam, A.}, \bibinfo{author}{Kanwal, N.},
  \bibinfo{author}{Asghar, M.}, \bibinfo{author}{Lee, B.},
  \bibinfo{year}{2022}.
\newblock \bibinfo{title}{A survey of modern deep learning based object
  detection models}.
\newblock \bibinfo{journal}{Digital Signal Processing} ,
  \bibinfo{pages}{103514}.
\bibitem[{Zou et~al.(2019)Zou, Shi, Guo and Ye}]{zou2019object}
\bibinfo{author}{Zou, Z.}, \bibinfo{author}{Shi, Z.}, \bibinfo{author}{Guo,
  Y.}, \bibinfo{author}{Ye, J.}, \bibinfo{year}{2019}.
\newblock \bibinfo{title}{Object detection in 20 years: A survey}.
\newblock \bibinfo{journal}{arXiv preprint arXiv:1905.05055} .

\end{thebibliography}


\end{document}